%% file: main.tex
\documentclass{article} % For LaTeX2e

\usepackage{letltxmacro}
\usepackage{amsmath}
\LetLtxMacro{\originaleqref}{\eqref}

\usepackage{iclr2024_conference,times}
\input{math_commands.tex}

\usepackage{hyperref}
\usepackage{url}

\usepackage[utf8]{inputenc} % allow utf-8 input
\usepackage[T1]{fontenc}    % use 8-bit T1 fonts
\usepackage{hyperref}       % hyperlinks
\usepackage{url}            % simple URL typesetting
\usepackage{booktabs}       % professional-quality tables
\usepackage{amsfonts}       % blackboard math symbols
\usepackage{nicefrac}       % compact symbols for 1/2, etc.
\usepackage{microtype}      % microtypography
\usepackage{xcolor}         % colors

\usepackage{algorithm}
\usepackage[noend]{algpseudocode}
\usepackage{amsmath}
\usepackage{amssymb}
\usepackage{amsthm}
\usepackage{lipsum}
\usepackage{makecell}
\usepackage{mathtools}
\usepackage{multicol}
\usepackage{multirow}
\usepackage{thmtools}
\usepackage{thm-restate}

\usepackage{changes}

\renewcommand{\eqref}{~\originaleqref}

\def\beq{\begin{equation} }\def\eeq{\end{equation} }\def\1{\mathbf{1}}

\newcommand{\mcomment}[1]{{\quad\quad\rhd\text{ #1}}}
\numberwithin{equation}{section}

\newtheorem{lemma}{Lemma}
\newtheorem{theorem}{Theorem}

\newtheorem{corollary}[theorem]{Corollary}
\newtheorem{remark}{Remark}

\ifx\assumption\undefined
\newtheorem{assumption}{Assumption}
\fi

\makeatletter
\def\smallunderbrace#1{\mathop{\vtop{\m@th\ialign{##\crcr
   $\hfil\displaystyle{#1}\hfil$\crcr
   \noalign{\kern3\p@\nointerlineskip}%
   \tiny\upbracefill\crcr\noalign{\kern3\p@}}}}\limits}
\makeatother

\newcommand{\cO}{\mathcal{O}}

\newcommand{\EE}{\mathbb{E}}
\newcommand{\RR}{\mathbb{R}}
\newcommand{\CC}{\mathbb{C}}

\newcommand{\bT}{\mathbf{T}}
\newcommand{\bTp}{\tilde{\mathbf{T}}}
\newcommand{\bX}{\mathbf{X}}
\newcommand{\bY}{\mathbf{Y}}

\newcommand{\bz}{\mathbf{z}}
\newcommand{\bA}{\mathbf{A}}
\newcommand{\bB}{\mathbf{B}}
\newcommand{\bC}{\mathbf{C}}

\newcommand{\bM}{\mathbf{M}}
\newcommand{\bS}{\mathbf{S}}
\newcommand{\bs}{\mathbf{s}}

\newcommand{\bPi}{\mathbf{\Pi}}
\newcommand{\bV}{\mathbf{V}}
\newcommand{\be}{\mathbf{e}}
\newcommand{\bx}{\mathbf{x}}
\newcommand{\xb}{\mathbf{x}}
\newcommand{\Sb}{\mathbf{S}}

\newcommand{\bH}{\mathbf{H}}
\newcommand{\bHp}{\tilde{\mathbf{H}}}
\newcommand{\bU}{\mathbf{U}}

\newcommand{\bLambda}{\mathbf{\Lambda}}
\newcommand{\bO}{\mathbf{O}}
\newcommand{\bI}{\mathbf{I}}
\newcommand{\bb}{\mathbf{b}}
\newcommand{\bw}{\mathbf{w}}
\newcommand{\bwp}{\tilde{\mathbf{w}}}
\newcommand{\bwpv}{\tilde{\mathbf{w}}^{(v)}}
\newcommand{\bwpb}{\tilde{\mathbf{w}}^{(b)}}
\newcommand{\bn}{\mathbf{n}}
\newcommand{\bnp}{\tilde{\mathbf{n}}}
\newcommand{\bv}{\mathbf{v}}
\newcommand{\bg}{\mathbf{g}}

\newcommand{\bDelta}{\bm{\Delta}}
\newcommand{\bzero}{\mathbf{0}}

\def\diag{\mathrm{diag}}
\newcommand{\norm}[1]{\left\|#1\right\|}
\newcommand{\modulus}[1]{\left|#1\right|}

\newcommand{\conjugate}[1]{\overline{#1}}
\newcommand{\dotprod}[1]{\left\langle #1\right\rangle}

\def\tr{\mathrm{tr}}
\newcommand{\numstage}{n}
\newcommand{\stagelength}{K}

\title{Accelerated Convergence of Stochastic Heavy Ball Method under Anisotropic Gradient Noise}

\author{%
  Rui Pan$^{1}$\thanks{Equal contribution.},
  \quad
  Yuxing Liu$^{2}$\footnotemark[1],
  \quad Xiaoyu Wang$^{1}$,
  \quad
  Tong Zhang$^{3}$
  \\
  \\
  $^{1}$The Hong Kong University of Science and Technology
  \\$^{2}$Fudan University \\ $^{3}$University of Illinois Urbana-Champaign \\
  \small\texttt{rpan@connect.ust.hk}, \texttt{yuxingliu20@fudan.edu.cn}, 
  \texttt{maxywang@ust.hk}, \\
  \texttt{tozhang@illinois.edu}
}

\date{}

\iclrfinalcopy % Uncomment for camera-ready version, but NOT for submission.
\begin{document}

\maketitle

\begin{abstract}
Heavy-ball momentum with decaying learning rates is widely used with SGD for optimizing deep learning models. In contrast to its empirical popularity, the understanding of its theoretical property is still quite limited, especially under the standard anisotropic gradient noise condition for quadratic regression problems. Although it is widely conjectured that heavy-ball momentum method can provide accelerated convergence and should work well in large batch settings, there is no rigorous theoretical analysis. In this paper, we fill this theoretical gap by establishing a non-asymptotic convergence bound for stochastic heavy-ball methods with step decay scheduler on quadratic objectives, under the anisotropic gradient noise condition. As a direct implication, we show that heavy-ball momentum can provide $\tilde{\mathcal{O}}(\sqrt{\kappa})$ accelerated convergence of the bias term of SGD while still achieving near-optimal convergence rate with respect to the stochastic variance term. The combined effect implies an overall convergence rate within log factors from the statistical minimax rate. This means SGD with heavy-ball momentum is useful in the large-batch settings such as distributed machine learning or federated learning, where a smaller number of iterations can significantly reduce the number of communication rounds, leading to acceleration in practice.
\end{abstract}
% This assumption is shown to be much more realistic than the common noisy ball assumptions, under which no positive result has been proved for HB momentum so far.

\section{Introduction}
\label{sec:intro}

Optimization techniques that can efficiently train large foundation models~\citep{devlin2018bert, brown2020gpt3, touvron2023llama, touvron2023llama2, ouyang2022instructgpt} are rapidly gaining importance. Mathematically, most of those optimization problems can be formulated as minimizing a finite sum
\begin{align*}
  \min_{\bw} f(\bw) \triangleq \frac{1}{N} \sum_{i=1}^N f_i(\bw),
\end{align*}

where numerical methods are normally applied to find the minimum of the above form. Among all those methods, stochastic gradient descent (SGD)~\citep{robbins1951sgd} and its variants can be regarded as one of the most widely used algorithms.

For instance, heavy-ball (HB) methods~\citep{polyak19641heavyball}, commonly referred as heavy-ball momentum, are one of those popular variants. Empirically, it was extremely helpful for accelerating the training of convolutional neural networks~\citep{szegedy2014googlenet, simonyan2015vgg16, he2015resnet, huang2017densenet, sandler2018mobilenetv2}. Theoretically, it has been shown to provide optimal acceleration for gradient descent (GD) on quadratic objectives~\citep{nemirovski1995information}.

Nonetheless, when it comes to SGD in theory, things become much different. Despite its huge success in practice, most theoretical results of stochastic heavy ball (SHB) were negative, showing that the convergence rates of heavy-ball methods are no better than vanilla SGD~\citep{devolder2013first, yuan2016influence, loizou2017linearly,kidambi2018moment-insufficiency, jain2018accelerating, li2022moment-last-iterate}. The existence of these gaps between GD and SGD, between practice and theory, is rather intriguing, which may make one wonder: \textit{Can stochastic heavy ball provide $\tilde{\Theta}(\sqrt{\kappa})$ accelerated convergence when the noise is small, such as under large-batch settings?}

To answer this question, the first step is to find the missing pieces in those negative results. One key observation is that all those negative results assumed constant learning rates, while in practice, decaying learning rates are usually used instead. Those decaying learning rates, often referred as learning rate schedules, were demonstrated to be critical for improving the performance of a trained model in real-world tasks~\citep{LoshchilovH17cosinedecay,howard2018lineardecay}. Furthermore, if one only considers the vanilla SGD algorithm, the theoretical property of most schedules have already been well inspected~\citep{shamir2013stochastic, jain2019making, ge2019step, harvey2019tight, pan2021eigencurve, wu2022last}. Briefly speaking, one can view learning rate schedules as a variance reduction technique, which helps alleviate the instability and deviation caused by stochastic gradient noise.

Since it has been pointed out by~\citep{polyak1987introduction} that variance reduction is the key to improving stochastic heavy ball's convergence rate, it is then natural to ask: \textit{Are there proper learning rate schedules that can help us achieve accelerated convergence for SHB under large-batch settings?}

Our paper gives a positive answer to this question. As a first step, we restrict ourselves to quadratic objectives. Although these problem instances are considered one of the simplest settings in optimization, they provide important insights for understanding a model's behavior when the parameter is close to a local optimum. Furthermore, past literature on Neural Tangent Kernel (NTK)~\citep{arora2019exact, jacot2018neural} suggests that the gradient dynamics of sufficiently wide neural networks resemble NTKs and can have their objectives approximated by quadratic objectives given specific loss functions.

Motivated by the empirical anisotropic behavior of SGD noises near minima of modern neural networks~\citep{sagun2017empirical, chaudhari2018stochastic, zhu2018anisotropic} and theoretical formalization of this noise property in least square regression~\citep{jain2018parallelizing, jain2018accelerating, pan2021eigencurve}, we conduct our analysis based on the assumption of anisotropic gradient noise, which is formally defined later as Assumption~\ref{ass:anisotropic_noise} in Section~\ref{sec:theory}. Notice that the very same condition has already been adopted or suggested by many past literatures~\citep{dieuleveut2017harder,jastrzkebski2017three,zhang2018energy, zhu2018anisotropic, pan2021eigencurve}.

\subsection{Our Contributions}
\label{sec:contrib}
\begin{enumerate}
  \item We introduce novel theoretical techniques for analyzing stochastic heavy ball with multistage schedules, providing several key properties for the involved $2\times2$ update matrix $\bT_i$. Specifically, we show that $\norm{\bT_{T-1}\bT_{T-2}...\bT_{0}}$ can be upper bounded by $\norm{\bT_{T-1}^T}$ under certain conditions. This allows SHB with changing learning rates to exhibit similar theoretical properties as vanilla SGD: for each eigenvalue, SHB first exponentially decreases the loss with large learning rates, then retains the reduced loss with small learning rates.
  \item As a direct result of this technical innovation, we present a non-asymptotic last iterate convergence rate for stochastic heavy ball with step decay learning rate schedule on quadratic objectives, under the standard anisotropic gradient noise assumption. To the best of our knowledge, this is the first non-asymptotic result for SHB on quadratics that clearly expresses the relationship among iteration number $T$, condition number $\kappa$ and convergence rate with step-decay schedules.
  \item Our results show that stochastic heavy ball can achieve near-optimal accelerated convergence under large-batch settings, while still retaining near-optimal convergence rate $\tilde{O}(d \sigma^2 / T)$ in variance (up to log factors away from the statistical minimax rate). % This bridges the gap between SHB in theory and its common practices in real-world applications. The role of large batch sizes is featured in our analysis, which allows the theoretically unfriendly method of SHB to provide true acceleration.
\end{enumerate}

\section{Related Work}
\label{sec:related_work}

\paragraph{Large batch training:} Large-batch training is a realistic setting of its own practical interest. In several recent efforts of accelerating large model training, it has been observed that large batch sizes are beneficial for accelerating the training process~\citep{you2017large, you2018imagenet, you2019large,24hbert2021,pan2022extremebert,wettig2022mask15}. On top of that, in distributed machine learning~\citep{verbraeken2019distributed-ml} and federated learning~\citep{kairouz2021federated-learning}, one can normally support an outrageous size of large batches by adding machines/devices to the cluster/network, but unable to afford a large number of iterations due to the heavy cost of communication~\citep{zheng2019communication,qian2021error}. This makes acceleration techniques even more tempting under those settings.

\paragraph{SGD + learning rate schedules:} In contrast, the research in SGD with learning rate schedules focused on more general settings without assuming constraints on the batch size. In~\citep{ge2019step}, the convergence rate of step decay was proved to be nearly optimal on strongly convex linear square regression problems. \citep{pan2021eigencurve} further pushed these limits to optimal for some special problem instances and offered a tighter upper bound, along with a lower bound for step decay. Concurrently, \citep{wu2022last} extended the analysis of \citep{ge2019step} to a dimension-free version under overparamterized settings, with tighter lower and upper bounds provided for step decay schedules. In~\citep{loizou2021polyak-stepsize}, the convergence rate of Polyak step size on strongly convex objectives was investigated. 
Nevertheless, all the bounds in above works require SGD to have at least $\tilde{\Omega}(\kappa \log c)$ iterations to reduce the excess risk by any factor of $c$. There are also works with looser bounds but focus on more general objectives. Since we restrict ourselves to quadratics, we just list some of them here for reference:~\citep{ghadimi2013optimal, hazan2014beyond, xu2016accelerate, yuan2019stagewise, vaswani2019painless, kulunchakov2019generic, davis2023stochastic, wolf2021stochastic}.

\paragraph{SGD + HB + constant learning rates:} Opposed to the positive results of near optimality for SGD, most results of stochastic HB with constant learning rates were negative, showing that its convergence rate cannot be improved unless extra techniques like iterate averaging are applied. In~\citep{loizou2017linearly, loizou2020momentum}, a linear convergence rate of SGD momentum on quadratic objectives for L2 convergence $\EE[\norm{\bw_T - \bw_*}^2]$ and loss $\EE[f(\bw_T) - f(\bw_*)]$ was established, which requires at least $\tilde{\Omega}(\kappa \log c)$ iterations. A better bound for L1 convergence $\norm{\EE[\bw_T - \bw_*]}^2$ and $\bB$ norm $\norm{\EE[\bw_T - \bw_*]}_\bB^2$ was also proposed, but whether they are relevant to loss convergence is unclear. Here $\bB$ is a positive definite matrix related to the problem instance and samples. In~\citep{kidambi2018moment-insufficiency}, momentum was proved to be no better than vanilla SGD on worst-case linear regression problems. In~\citep{jain2018accelerating}, both SGD and momentum are shown to require at least $\Omega(\kappa)$ single-sample stochastic first-order oracle calls to reduce excess risk by any factor of $c$, thus extra assumptions must be made to the noise. A modified momentum method using iterate averaging was then proposed on least square regression problems and achieves $\tilde{\cO}(\sqrt{\kappa \tilde{\kappa}})$ iteration complexity with an extra noise assumption. Here $\tilde{\kappa} \le \kappa$ is the statistical condition number. In~\citep{gitman2019understanding}, a last iterate rate of SGD momentum on quadratic objectives was presented, but the convergence rate is asymptotic. Non-asymptotic linear distributional convergence was shown in~\citep{can2019accelerated}, where SHB with constant learning rates achieves accelerated linear rates $\Omega(\exp(-T/\sqrt{\kappa}))$ in terms of Wasserstein Distances between distributions. However, this does not imply linear convergence in excess risks, where the variance is still a non-convergent constant term. In~\citep{mai2020convergence}, a class of weakly convex objectives were studied and a convergence rate of $\cO(\kappa/\sqrt{T})$ was established for gradient L2 norm. In~\citep{wang2021modular}, HB on GD is analyzed and shown to yield non-trivial speedup on quadratic objectives and two overparameterized models. However, the analysis was done in GD instead of SGD. 
\textcolor{black}{In \citep{bollapragada2022fast}, SHB was shown to have a linear convergence rate $1-1/\sqrt{\kappa}$ with standard constant stepsize and large enough batch size on finite-sum quadratic problems. Their analysis, however, was based on an extra assumption on the sample method. \citep{tang2023acceleration} proved SHB converges to a neighborhood of the global minimum faster than SGD on quadratic target functions using constant stepsize. }
%However, their assumptions on noise allowed the noise factor to be very large and they didn't explicitly show how much can HB accelerates GD.}
In~\citep{yuan2021decentlam}, a modified decentralized SGD momentum algorithm was proposed for large-batch deep training. Although it achieves $\tilde{\cO}(1/T)$ convergence rate on a $L$-smooth and $\mu$-strongly convex objectives, it still requires at least $\tilde{\Omega}(\kappa)$ number of iterations to converge, which is no better than SGD. \citet{wang2023marginal} also provided cases where SHB fails to surpass SGD in small and medium batch size settings, suggesting that momentum cannot help reduce variance. There are also other variants of momentum such as Nesterov momentum~\citep{nesterov2003introductory, liu2018accelerating,aybat2019universally}, or modified heavy ball, but since we only consider the common version of heavy ball momentum here, we omit them in our context.

\paragraph{SGD + HB + learning rate schedules:}
As for SHB with learning rate schedules, only a limited amount of research has been conducted so far. In~\citep{liu2020improved}, the convergence property of SHB with multistage learning rate schedule on $L$-smooth objectives was investigated. However, the inverse relationship between the stage length and learning rate size was implicitly assumed, thus its convergence rate is actually $\cO(1/\log_{\alpha} T)$ for some constant $\alpha > 1$. In~\citep{jin2022msgd-smooth}, a convergence rate was derived for general smooth objectives. But the relationship between the convergence rate and $T$ is still unclear, and the results were comparing SGD and SHB by their upper bounds.
In~\citep{li2022moment-last-iterate}, a worst-case lower bound of $\Omega(\ln T / \sqrt{T})$ was found for SHB with certain choices of step sizes and momentum factors on Lipschitz and convex functions. A FTRL-based SGD momentum method was then proposed to improve SHB and achieve $\cO(1/\sqrt{T})$ convergence rate for unconstrained convex objectives. Furthermore, in~\citep{wang2021nonconvex-sgdm-step}, a $\cO(1/\sqrt{T})$ bound was derived on general smooth non-convex objectives, whose analysis supports a more general class of non-monotonic and cyclic learning rate schedules. All these results only proved that SHB is no worse than SGD, or were comparing two methods by their upper bounds instead of lower bound against upper bound. Only until recently has SHB been shown to be superior over SGD in some settings. 
In~\citep{zeng2023adaptive}, a modified adaptive heavy-ball momentum method was applied to solve linear systems and achieved better performance than a direct application of SGD.
In~\citep{sebbouh2021almost}, SHB was shown to have a convergence rate arbitrarily close to $o(1/\sqrt{T})$ on smooth convex objectives. However, the analysis stopped at this asymptotic bound and did not provide any practical implications of this result.

In contrast to all the aforementioned works, we provide positive results in theory to back up SHB's superior empirical performance, showing that SHB can yield accelerated convergence on quadratic objectives by equipping with large batch sizes and step decay learning rate schedules.

\section{Main Theory}
\label{sec:theory}

\subsection{Problem Setup}
\label{sec:problem_setup}
In this paper, we analyze quadratic objectives with the following form,
\begin{align}
	\label{eq:problem_setup}
	 \min_{\bw} f(\bw) \triangleq
	   \EE_\xi\left[ f(\bw, \xi)\right],
	 \mbox{ where } f(\bw, \xi) = \frac{1}{2} \bw^\top \bH(\xi) \bw - \bb(\xi)^\top \bw,
\end{align}

where $\xi$ denotes the data sample. By setting gradient to $\mathbf{0}$, the optimum of $f(\bw)$ is obtained at
\begin{align}
\label{eq:optima}
	\bw_* = \bH^{-1} \bb,
	\mbox{ where }
	\bH = \EE_\xi\left[\bH(\xi)\right], \quad
	\bb = \EE_\xi\left[\bb(\xi)\right].
\end{align}

In addition, we denote the smallest/largest eigenvalue and condition number of the Hessian $\bH$ to be
\begin{equation}
    \label{eq:hess_eigenvalue_notation}
	\mu \triangleq \lambda_{\min}(\bH), \quad
	L \triangleq \lambda_{\max}(\bH), \quad
	\kappa \triangleq L/\mu,
\end{equation}
where eigenvalues from largest to smallest are denoted as
\begin{align*}
  L = \lambda_1 \ge \lambda_2 \ge \dots \ge \lambda_d = \mu > 0.
\end{align*}

We consider the standard stochastic approximation framework~\citep{kushner2012stochastic} and denote the gradient noise to be
\begin{equation}
    \label{eq:nt_def}
	\bn_t
	\triangleq
    \nabla f(\bw_t) - \nabla_\bw f(\bw_t, \xi).
	%=
	%\left(\bH \bw_t - \bb\right)
	%- \left(\bH(\xi) \bw_t - \bb(\xi)\right).
\end{equation}

Throughout the paper, the following assumptions are adopted.

\begin{assumption}
\label{ass:independent_noise}
  (Independent gradient noise)
  \begin{equation}\begin{aligned}
    \label{eq:independent_noise}
    \{ \bn_t \} \mbox{ are pairwise independent.}
  \end{aligned}\end{equation}
\end{assumption}

\begin{assumption}
\label{ass:unbiased_noise}
  (Unbiased gradient noise)
  \begin{equation}\begin{aligned}
    \label{eq:unbiased_noise}
    \EE\left[\bn_t\right] = \bzero.
  \end{aligned}\end{equation}
\end{assumption}

\begin{assumption}
\label{ass:anisotropic_noise}
  (Anisotropic gradient noise)
  \begin{equation}\begin{aligned}
    \label{eq:anisotropic_noise}
    \EE\left[\bn_t \bn_t^\top\right] \preceq \sigma^2 \bH.
  \end{aligned}\end{equation}
\end{assumption}

% Assumption~\ref{ass:independent_noise} and \ref{ass:unbiased_noise} are common in gradient noise setting(\textcolor{blue}{need some examples}). 
The anisotropic gradient noise assumption has been adopted by several past literatures~\citep{dieuleveut2017harder,pan2021eigencurve}, along with evidence supported in~\citep{zhu2018anisotropic,sagun2017empirical,zhang2018energy,jastrzkebski2017three,wu2022alignment}, which suggest that gradient noise covariance is normally close to the Hessian in neural networks training. 
% \textcolor{blue}{\citep{li2020hessian,wu2022alignment} further argue that this relationship may come from the closeness between noise covariance and the Fisher Matrix.}
% There is a line of works~\citep{jastrzkebski2017three} following this observation and trying to make theoretical analysis.

Let $\mathcal{B}_t$ be the minibatch of samples at iteration $t$. For simplicity, we only consider the setting where all minibatches share the same batch size
\begin{equation}
\label{def:batch_size}
\begin{aligned}
  |\mathcal{B}_t| \equiv M,
  \mbox{ for } \forall t=0,1,\dots,T-1.
\end{aligned}
\end{equation}

It follows that the number of samples is $N = M T$.

\begin{remark}
    One may also employ the common assumptions on strongly convex least square regressions as \citep{bach2013non,jain2018accelerating,ge2019step}:
    \begin{equation}
        \begin{aligned}
        &\min_w f(\bw), \quad\text{where } f(\bw) \overset{\text{def}}{=} \frac{1}{2}\mathbb{E}_{(\xb,y)\thicksim\mathcal{D}}\left[(y-\dotprod{\xb,\bw})^2 \right], \text{ and} \\
        &\text{(1) } y = \bw_*^T\xb+\epsilon, \text{ where } \mathbb{E}_{(\xb,y)\thicksim\mathcal{D}}\left[\epsilon^2\xb\xb^\top\right]\preceq \Tilde{\sigma}^2\bH, \\
        &\text{(2) } \mathbb{E}\left[\norm{\xb}^2\xb\xb^\top\right] \preceq R^2\bH
        \end{aligned}
    \end{equation}
    which can also be translated into our settings under the compact set constraint $\bw \in \Lambda$, as suggested in~\citep{jain2018accelerating}. 
    %with proofs provided in Appendix \ref{appendix:assumption_discussion}
    % For simplicity (?) we use the quadratics setting in our main paper and delay the discussion for the assumptions to Appendix \ref{appendix:assumption_discussion}. 
\end{remark}

\subsection{Suboptimality of SGD}
\label{sec:sgd_suboptimal}

We begin with the vanilla version of SGD,
\begin{equation}
\label{eq:sgd_update}
\begin{aligned}
    &\bw_{t+1} = \bw_{t} - \frac{\eta_t }{|\mathcal{B}_t|}
    \sum_{\xi \in \mathcal{B}_t} \nabla_{\bw} f(\bw_t, \xi),
\end{aligned}
\end{equation}

whose theoretical property is well understood on quadratic objectives~\citep{bach2013non,jain2018parallelizing,ge2019step, pan2021eigencurve}. Here $\eta_t$ means the learning rate at iteration $t$. It is known that SGD requires at least $\Omega(\kappa)$ iterations under the setting of batch size $M=1$~\citep{jain2018accelerating}, 
nevertheless, whether this lower bound still holds for large batch settings is not rigorously claimed yet. Here we provide Theorem~\ref{thm:sgd_lower_bound} to make things clearer.

\begin{restatable}[]{theorem}{theoremsgdlowerbound}
  \label{thm:sgd_lower_bound}
  There exist quadratic objectives $f(\bw)$ and initialization $\bw_0$, no matter how large the batch size is or what learning rate scheduler is used, as long as $\eta_t \le 2/L$ for $\forall t = 0, 1, \dots, T-1$, running SGD for $T$ iterations will result in
  \begin{align*}
     \EE\left[f(\bw_{T}) - f(\bw_*)\right]
     \ge
     \frac{f(\bw_0) - f(\bw_*)}{2} \cdot \exp\left(-\frac{8T}{\kappa}\right)
  \end{align*}
\end{restatable}

The proof is available in Appendix~\ref{appendix:proof_sgd_suboptimal}. The existence of those counterexamples suggests that in the worst case, SGD requires at least $T \ge \kappa/8 \cdot \ln (c/2) = \Omega(\kappa \log c)$ iterations to reduce the excess risk by a factor of $c \ge 2$, while in practice, $\kappa$ can be quite large near the converged point~\citep{sagun2017eigenvalues, arjevani2020analytic, yao2020pyhessian}.

% This number of iterations will incur huge training costs for large models, rendering the fast rollout of AI products impossible.

\subsection{Acceleration with Stochastic Heavy Ball}
\label{sec:sgdm}

To overcome this limitation, heavy-ball momentum~\citep{polyak19641heavyball} is normally  adopted by engineers to speed up SGD, equipped with various types of learning rate schedulers
\begin{equation}
\label{eq:sgdm_update}
\begin{aligned}
    &\bv_{t+1}
    = \beta \bv_t
    + \frac{\eta_t }{|\mathcal{B}_t|}
    \sum_{\xi \in \mathcal{B}_t} \nabla_{\bw} f(\bw_t, \xi)
    \\
    &\bw_{t+1} = \bw_{t} - \bv_{t+1}.
\end{aligned}
\end{equation}

Despite its huge success in practice, the theoretical understanding of this method is still limited, especially for quadratic objectives. Furthermore, although it was widely recognized that stochastic heavy ball should provide acceleration in large batch settings, positive theoretical results so far are still insufficient to clearly account for that. We attempt to fill this gap.

In this section, we will show that SHB equipped with proper learning rate schedules can indeed speed up large batch training. The whole analysis is done in a general multistage learning rate scheduler framework, as shown in Algorithm~\ref{alg:multistage_sgdm}. Specifically, in this framework, learning rates are divided into $\numstage$ stages, with each stages' learning rates and number of iterations being $\eta'_\ell$ and $k_\ell \triangleq \stagelength$ respectively, i.e.
\begin{equation}
\label{eq:def_multistage}
\begin{aligned}
  &t^{(\text{start})}_\ell
  \triangleq
 \stagelength (\ell - 1),
  \quad
  t^{(\text{end})}_\ell
  \triangleq
  \stagelength \ell - 1
  \\
  &\eta_t \equiv \eta'_\ell,
  % \quad \beta_t = \beta'_\ell,
  \quad \mbox{ for }
  \forall t=t^{(\text{start})}_\ell, t^{(\text{start})}_\ell + 1, \dots, t^{(\text{end})}_\ell.
\end{aligned}
\end{equation}

\begin{algorithm}
\caption{Multistage Stochastic Heavy Ball with minibatch}
\label{alg:multistage_sgdm}
\textbf{Input:} Number of stages $\numstage$, learning rates $\{ \eta'_\ell \}_{\ell=1}^{\numstage}$, momentum $\beta$, stage lengths $\stagelength$, minibatch size $M$, initialization $\bw_0 \in \mathbb{R}^d$ and $\bv_0 = \bzero$.
\begin{algorithmic}[1]
  \State $t \gets 0$
  \Comment Iteration counter
  \For{$\ell = 1, 2, \dots, \numstage$}
    \State $\eta_t \gets \eta'_\ell$
    %\State $\beta_t \gets \beta'_\ell$
    \For {$i = 1, 2, \dots, \stagelength$}
      \State Sample a minibatch $\mathcal{B}$ uniformly from the training data
      \State $\bg_t \gets \frac{1}{M} \sum_{\xi \in \mathcal{B}} \nabla_\bw f(\bw, \xi)$
      \Comment Mean gradient over a minibatch
      \State $\bv_{t+1} \gets \beta \bv_t + \eta_t \bg_t$
      \label{alg:line:sgdm_update_begin}
      \State $\bw_{t+1} \gets \bw_t - \bv_{t+1}$
      \label{alg:line:sgdm_update_end}
      \State $t \gets t + 1$
    \EndFor
  \EndFor
  \State \Return $\bw_t$
  \Comment Last iterate
\end{algorithmic}
\end{algorithm}

Given the above step decay scheduler, the following theorem states the convergence rate for SHB on quadratic objectives. To the best of our knowledge, this is the first non-asymptotic result that explicitly expresses the relationship between $T$ and the convergence rate of mutlistage SHB on quadratic objectives.

\begin{restatable}[]{theorem}{theoremmain}
\label{thm:main}
Given a quadratic objective $f(\bw)$ and a step decay learning rate scheduler with $\beta=\left(1-1/\sqrt{\kappa}\right)^2$ with $\kappa \ge 4$, and $\numstage\equiv T/\stagelength$ with settings that 
    \begin{enumerate}
        \item decay factor $C$
        \begin{align}\label{eq:req_C}
            1 < C \le T\sqrt{\kappa}.
        \end{align}
        \item stepsize $\eta_\ell'$ 
        \begin{align}\label{eq:req_var_eta_1}
            \eta_\ell' = \frac{1}{L}\cdot\frac{1}{C^{\ell-1}}
        \end{align}
        
        \item stage length $\stagelength$ 
        \begin{align}\label{eq:req_var_k_l}
            \stagelength = \frac{T}{\log_C\left(T\sqrt{\kappa}\right)}
        \end{align}
        \item total iteration number $T$
        \begin{align}\label{eq:req_var_T}
            \frac{T}{\ln\left(2^{14}T^8\right)\cdot \ln\left( 2^6T^4\right)\cdot\log_C(T^2)} \geq 2C\sqrt{\kappa} ,
        \end{align}
    \end{enumerate}
   then such scheduler exists, and the output of Algorithm \ref{alg:multistage_sgdm} satisfies
    \small\begin{align*}
        \mathbb{E}[f(\bw_{T})-f(\bw_*)] 
        \le&
        \mathbb{E}\left[ f(\bw_0)-f(\bw_*) \right]\cdot \exp\left( 15\ln 2+2\ln T+2\ln\kappa -\frac{2T}{\sqrt{\kappa}\log_C\left(T\sqrt{\kappa}\right)} \right) \\
        &+
        \frac{4096C^2d\sigma^2}{MT} \ln^2\left( 2^6T^4\right)\cdot \log_C^2\left(T\sqrt{\kappa}\right).
    \end{align*}
\end{restatable}
Or equivalently, the result can be simplified to the following corollary.
\begin{corollary}\label{cor:simplified_main}
    Given a quadratic objective $f(\bw)$ and a step decay learning rate scheduler and momentum defined in Theorem \ref{thm:main}, with $T\geq\Tilde{\Omega}\left( \sqrt{\kappa}\right)$ and $\kappa \ge 4$, the output of Algorithm~\ref{alg:multistage_sgdm} satisfies 
    \begin{align*}
        \EE\left[f(\bw_T)-f(\bw_*)\right] \le \mathbb{E}\left[ f(\bw_0)-f(\bw_*) \right]\cdot \exp\left(-\tilde{\Omega}\left(\frac{T}{\sqrt{\kappa}}\right)\right)
        +
        \tilde{\cO}\left( \frac{d\sigma^2}{MT}\right),
    \end{align*}
    where $\Tilde{\cO}(\cdot)$ and $\Tilde{\Omega}(\cdot)$ are used to hide the log factors.
\end{corollary}

Notice that the bias term $[f(\bw_0) - f(\bw_*)] \cdot \exp(-\tilde{\Omega}(T/ \sqrt{\kappa}))$ is exponentially decreasing after $T = \tilde{\cO}(\sqrt{\kappa})$ iterations, while the variance term can be bounded by $\tilde{\cO}(1/T)$. This implies that under the large batch setting, if the batch size is large enough to counteract the extra constant in the variance term, accelerated convergence will be possible as compared to the iteration number of $\Tilde{O}\left(\kappa\right)$ required by SGD. It is worth noting that this $\tilde{\Theta}(\sqrt{\kappa})$ acceleration is only log factors away from the optimal acceleration~\citep{nemirovski1995information} of Heavy Ball~\citep{polyak19641heavyball} and Nesterov Accelerated Gradient~\citep{nesterov1983method} in deterministic case.

% 2 points: anisotropic noise? and matrix technique
The proof outline can be split into two major steps. The first step is bias-variance decomposition, which decomposes the expected excess risk~$\EE[f(\bw_T)]-f(\bw_*)$ into two terms: bias and variance, where bias measures the deterministic convergence error and variance measures the effect of the gradient noise. This step adapts the well-known bias-variance decomposition technique of SGD~\citep{bach2013non,jain2018parallelizing,ge2019step,pan2021eigencurve} to SHB. Inside the adapted decomposition, a critical ``contraction'' term $\norm{\bT_{T-1}\bT_{T-2}...\bT_{0}}$ is introduced in both bias and variance, where each matrix $\bT_t\in\RR^{2\times 2}$ depends on step size $\eta_t$ and differs only by a diagonal matrix $\bDelta_t \triangleq \bT_t - \bT_0 = \diag(\delta_t, 0)$.

% If a constant step size is employed, say, $\eta_t\equiv \eta$ for $t=0,...,T-1$, it follows that $\bT_t\equiv\bT$ for $t=0,...,T-1$ and~$\norm{\bT_{T-1}\bT_{T-2}...\bT_{0}} = \norm{\bT^{T}}$. It is well-known that for a matrix $\bT$, $\lim_{n\to \infty}\norm{\bT^n}^{1/n}=\rho(\bT)$, where $\rho(A)$ is the spectral radius of $A$. For a finite constant $T$, we can also bound $\norm{\bT^{T}}$ tightly using its spectral radius and thus obtain a tight bound for bias and variance. In~\citep{can2019accelerated,bollapragada2022fast}, the authors have also presented convergence of SHB on quadratic objectives, both with constant step sizes. We note that their results can be obtained straightforwardly by combining the adapted bias-variance decomposition and this bound for matrix powers.

The second major step is to bound the matrix product tightly. Notice that this term has a form of $\prod_{t=0}^{T-1} (1 - \eta_t \lambda_j)$ for SGD and is much easier to analyze. For the general form of $\norm{\bT_{T-1}\bT_{T-2}...\bT_{0}} = \norm{(\bT_0 + \bDelta_{T-1}) (\bT_0 + \bDelta_{T-2})...\bT_{0}}$, the major difficulty arises from the \textit{non-commutative} matrix products of different $\bT_t$'s and $\bDelta_t$'s. To overcome this obstacle, a novel technique is proposed in our paper, which is based on the special structure of $\bT_t$. The key observation is that product with form $(\bT_{s_1} \bDelta_{s'_1} \bT_{s_2} \bDelta_{s'_2} \dots \bT_{s_n} \bDelta_{s'_n}) \in \RR^{2\times 2}$ retains two important properties: 1) The first column is always nonnegative and second column is always nonpositive; 2) The absolute value of each entry is a monotonical increasing function of $\delta_1, \dots, \delta_{T-1}$. Hence the sum of the exponential number of terms in the binomial-like expansion also retains those two properties, which leads to a bound $\norm{\bT_{T-1}\bT_{T-2}...\bT_{0}} \le \norm{\bT_{T-1}^T}$ tight under certain conditions. This key technique, as rigorously stated in Lemma~\ref{lem:matrix_main} in Appendix, combined with subtle analysis of $\bT_t$ and learning rate schedule techniques in~\citep{ge2019step,pan2021eigencurve}, gives birth to Theorem~\ref{thm:main}.
The full detail of the proof is provided Appendix~\ref{appendix:proof_sgdm}.

\section{Experiments}
\label{sec:exp}

To verify our theoretical findings, two sets of experiments are conducted. The first one is ridge regression, which has a quadratic loss objective and is closer our theoretical settings. The second one is image classification on CIFAR-10~\citep{krizhevsky2009cifar10} with ResNet18~\citep{he2015resnet}, DenseNet121~\citep{huang2017densenet} and MobilenetV2~\citep{sandler2018mobilenetv2}, which is more of a practical interest regarding our theory's potential applications.

\subsection{Ridge Regression}
\label{sec:exp_ridge_regression}

In ridge regression, we consider the following setting
\begin{equation}
\label{eq:ridge_setting}
\begin{aligned}
  f(\bw) = \frac{1}{n} \norm{\bX \bw - \bY}_2^2 + \alpha \norm{\bw}_2^2,
\end{aligned}
\end{equation}

whose optimum has an analytic form
\begin{align*}
  \bw_*
  =
  \left(\bX^\top \bX + n \alpha \bI\right)^{-1} \bX^\top \bY.
\end{align*}
Therefore the optimum loss $f(\bw_*)$ can be directly computed. We use \texttt{a4a}\footnote{The dataset is accessible in \url{https://www.csie.ntu.edu.tw/\~cjlin/libsvmtools/datasets/binary.html\#a4a/}.} dataset~\citep{chang2011libsvm, Dua2017UCI} to realize this setting, which contains $n=4,781$ samples and $d=123$ features.

In all of our experiments, we set the number of epochs to $100$, so the total amount of data is $N = 478,100$. Besides, we set different batch sizes $M \in \{2048, 512, 128\}$, and initialize $\bw_0$ from a uniform distribution $(-1, 1)^d$. The partial batch at the end of each epoch is not truncated, which means the total number of iterations $T = \ceil{N / M}.$

Regarding hyperparameter choices for each scheduler \& method, we do grid searches according to Table~\ref{tab:ridge_hyperparam} in Appendix~\ref{appendix:exp_details} and report the best loss for each random seed. For all schedulers, we set $\eta_0 \in \{ 10^0, 10^{-1}, 10^{-2}, 10^{-3} \}$. As for the choice of momentum factor $\beta$, we set $\beta = 0.9$ for stochastic heavy ball methods.

\begin{table}[h!]
  \caption{Training loss statistics of ridge regression in \texttt{a4a} dataset over 5 runs.}
  \label{tab:ridge_exp_main}
  \footnotesize
  \begin{center}
    \begin{tabular}{cccccc}
      \toprule

      \makecell{Methods/Schedules}
      & \multicolumn{4}{c}{$\left(f(\bw) - f(\bw_*)\right) \times 10^{-2}$}

      \\
      \midrule
      & Batch size $M=512$
      & $M=128$
      & $M=32$
      & $M=8$

      \\
      \midrule

      \makecell{SGD + constant $\eta_t$}
      & 2.10$\pm$0.46 
      & 1.17$\pm$0.81
      & 1.27$\pm$0.27
      & 0.94$\pm$0.83
      \\
      \midrule

      \makecell{SGD + step decay}
      & 2.44$\pm$0.45
      & 0.64$\pm$0.04
      & 0.11$\pm$0.01
      & \textbf{0.04$\pm$0.04}     

      \\
      \midrule

      \makecell{SHB + constant $\eta_t$}
      & 0.86$\pm$0.55 
      & 0.55$\pm$0.26
      & 1.03$\pm$0.35
      & 0.97$\pm$0.58

      \\
      \midrule

      \makecell{SHB + step decay}
      & \textbf{0.13$\pm$0.03} 
      & \textbf{0.01$\pm$0.00}
      & \textbf{0.03$\pm$0.02}
      & 0.06$\pm$0.05
      \\
      \bottomrule
    \end{tabular}
  \end{center}
\end{table}

As shown in Table~\ref{tab:ridge_exp_main}, one can observe that SHB are generally much better than SGD under large batch settings, and the step decay schedule always helps. The role of learning rate schedule and heavy-ball momentum is especially evident under the setting of $M=512$, where SHB is able to greatly reduce the loss with a much smaller bias, but still has a large loss due to the existence of variance. This variance term is then further handled by step decay schedule and leads to a fast convergence. As the batch size decreases, the variance term becomes dominant, which explains the closing gap between SGD and SHB.

\subsection{Image Classification on CIFAR-10}
\label{sec:exp_cifar10}

In image classification, our key focus is still verifying the superiority of SHB over SGD, so no heavy tuning was done for $\beta$. We follow the common practice of $\beta = 0.9$ for our algorithm in Theorem~\ref{thm:main}. To simulate the practical settings of distributed learning and federated learning, we restrict the number of iterations to be a few thousands~\citep{kairouz2021federated-learning}, which roughly translated into $\#Epoch = 10$ for batch size $M=128$ and $\#Epoch = 100$ for batch size $M=2048$. On top of that, for batch size $M=2048$, we replicate $16$ nodes with micro batch size $128$ on each node, hence the performance on distributed learning can be further simulated.

% In addition, notice that the implemented momentum method\footnote{Please refer to the docstring of SGD's implementation in Pytorch (\url{https://pytorch.org/docs/stable/_modules/torch/optim/sgd.html\#SGD}).} in PyTorch~\citep{pytorch2019} differs from the standard form of heavy-ball momentum defined in Eqn.~\eqref{eq:sgdm_update}, we modify the code accordingly to align with the theory.

In this experiment, CIFAR-10~\citep{krizhevsky2009cifar10} dataset is adopted, which contains $50,000$ training samples and $10,000$ test samples. We use $5,000$ randomly chosen samples in the training set to form a validation set, then conduct grid searches by training on the remaining $45,000$ samples and selecting the hyperparameter with the best validation accuracy. The selected hyperparameter is then used for training the whole $50,000$ samples and testing on the test set. The final test results are thereby summarized in Table~\ref{tab:cifar-10_main}. For grid searches, we choose learning rate $\eta_0 \in \{ 1, 0.1, 0.01, 0.001\}$, with decay rate $\gamma \in \{1/2, 1/5, 1/10\}$ and number of intervals $\numstage \in \{ 3, 4, 5, 6\}$.

\begin{table}[h!]
  \caption{CIFAR-10: training losses and test accuracy of different methods over 5 trials.}
  \label{tab:cifar-10_main}
  \scriptsize
  \begin{center}
    \begin{tabular}{cccccccc}
      \toprule
      Setting
      & Method
      & \multicolumn{2}{c}{Resnet18}
      & \multicolumn{2}{c}{DenseNet121}
      & \multicolumn{2}{c}{MobilenetV2}
      %& \multicolumn{2}{c}{VGG16}
      %& \multicolumn{2}{c}{GoogLeNet}
      \\
      &
      & \makecell{Crossent.\\Loss} & \makecell{Acc(\%)}
      & \makecell{Crossent.\\Loss} & \makecell{Acc(\%)}
      & \makecell{Crossent.\\Loss} & \makecell{Acc(\%)}
      \\
      \midrule

      \multirow{2}{*}{\makecell{$M=128$\\$(\#Epoch=10)$}}
      & SGD
      & 0.46$\pm$0.01 & 81.19$\pm$0.93
      & 0.22$\pm$0.01 & 88.58$\pm$0.23
      & 0.45$\pm$0.00 & 82.90$\pm$0.37
      % & \textbf{0.31$\pm$0.00} & 84.94$\pm$0.82
      % & \textbf{0.15$\pm$0.00} & 88.80$\pm$0.26
      \\
      & SHB
      & \textbf{0.38$\pm$0.08} & \textbf{85.16$\pm$2.30}
      & \textbf{0.18$\pm$0.00} & \textbf{88.63$\pm$0.27}
      & \textbf{0.35$\pm$0.01} & \textbf{86.23$\pm$0.23}
      % & \textbf{0.31$\pm$0.00} & \textbf{86.72$\pm$0.20}
      % & 0.18$\pm$0.00 & \textbf{89.30$\pm$0.19}

      \\
      \midrule

      \multirow{2}{*}{\makecell{$M=128\times16$\\$(\#Epoch=100)$}}
      & SGD
      & 0.33$\pm$0.01 & 83.82$\pm$0.42
      & 0.01$\pm$0.00 & 89.28$\pm$0.23
      & 0.32$\pm$0.02 & 84.37$\pm$0.77
      % & 0.02$\pm$0.00 & 88.65$\pm$0.17
      % & 0.01$\pm$0.00 & 88.11$\pm$0.20
      \\
      & SHB
      & \textbf{0.01$\pm$0.00} & \textbf{89.78$\pm$0.23}
      & \textbf{0.00$\pm$0.00} & \textbf{92.46$\pm$0.15}
      & \textbf{0.07$\pm$0.01} & \textbf{89.57$\pm$0.18}
      % & 0.01$\pm$0.00 & 91.14$\pm$0.17
      % & 0.00$\pm$0.00 & 92.79$\pm$0.24
      \\
      \bottomrule
    \end{tabular}
  \end{center}
\end{table}

\newcommand{\figurescale}{0.33}
\begin{figure}[h!]
    \centering
    \includegraphics[scale=\figurescale]{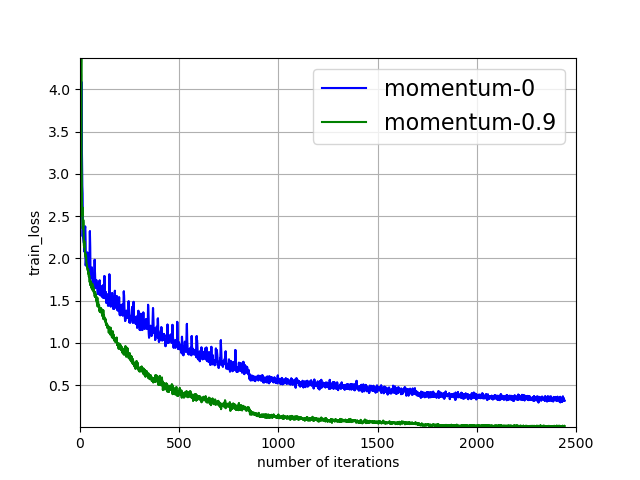}
    \includegraphics[scale=\figurescale]{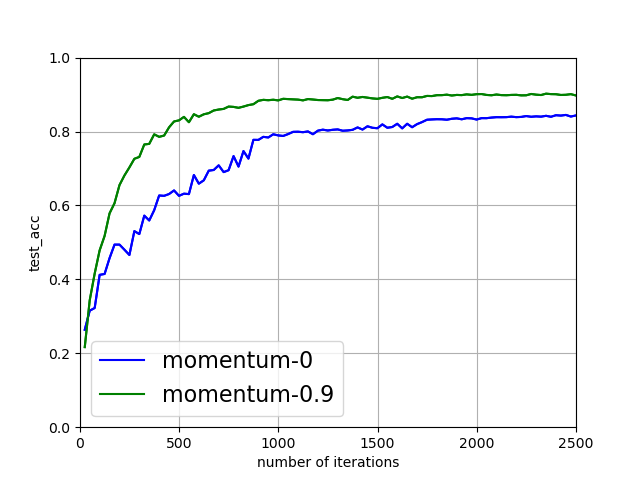}
    \includegraphics[scale=\figurescale]{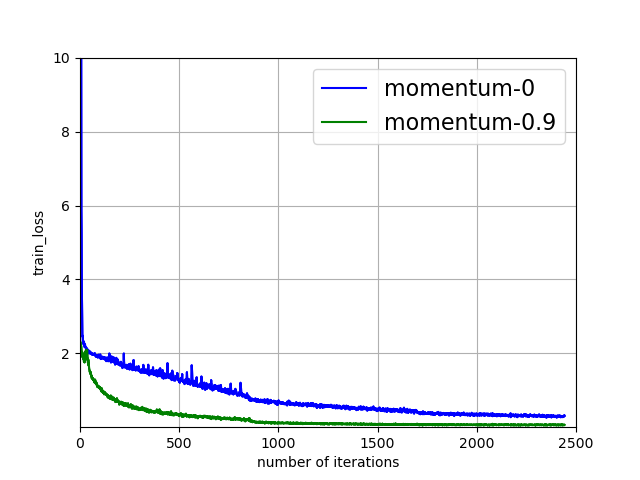}
    \includegraphics[scale=\figurescale]{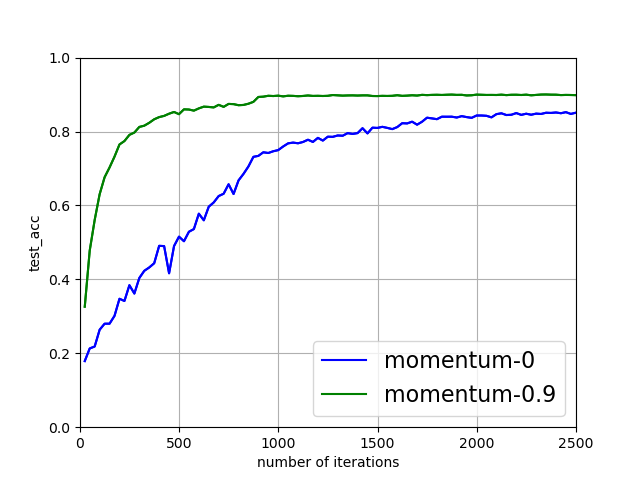}
    \includegraphics[scale=\figurescale]{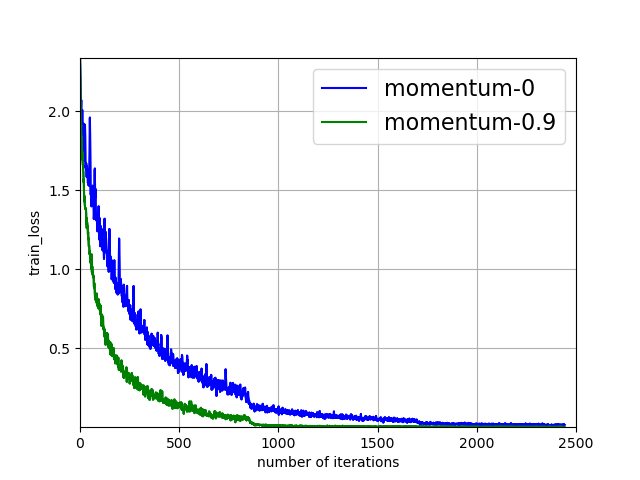}
    \includegraphics[scale=\figurescale]{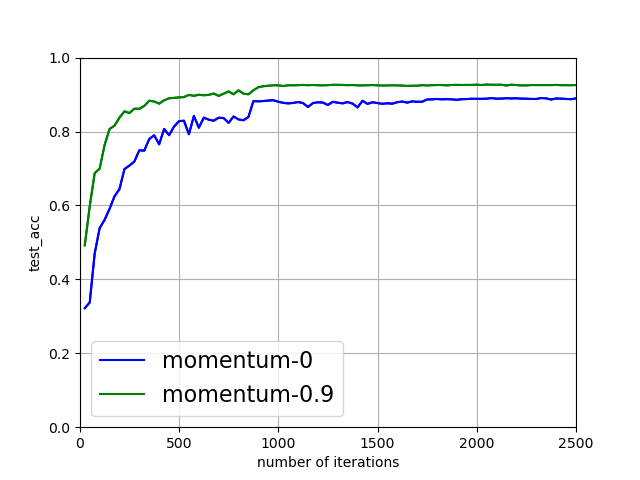}

    \caption{CIFAR-10 training statistics of batch size $M=128 \times 16$ and $\#Epoch=100$ on Resnet18, DenseNet121 and MobilenetV2 (from top to bottom). \textbf{Left}: Training loss; \textbf{Right}: Test accuracy.}
    \label{fig:cifar-10_main}
\end{figure}

One can observe in Table~\ref{tab:cifar-10_main} and Figure~\ref{fig:cifar-10_main} that under the large batch setting, SHB provides huge acceleration over SGD and achieves a significant performance improvement. This offers empirical evidence for our theory and suggests its practical value: \textit{Heavy Ball Momentum can provide true acceleration for SGD under large-batch settings.}

\section{Conclusion}
\label{sec:conclusion}

In this paper, we present a non-asymptotic convergence rate for Stochastic Heavy Ball with step decay learning rate schedules on quadratic objectives. The proposed result demonstrates SHB's superiority over SGD under large-batch settings. To the best of our knowledge, this is the first time that the convergence rate of SHB is explicitly expressed in terms of iteration number $T$ given decaying learning rates on quadratic objectives. Theoretically, our analysis provides techniques general enough to analyze any multi-stage schedulers with SHB on quadratics. Empirically, we demonstrate the practical benefits of heavy-ball momentum for accelerating large-batch training, which matches our theoretical prediction and explains heavy-ball momentum's effectiveness in practice to a certain degree.

\newpage
\section*{Acknowledgments}
Rui Pan acknowledges support from the Hong Kong PhD Fellowship Scheme (HKPFS).

\bibliographystyle{plainnat}
\bibliography{iclr2024_conference}

\appendix
\allowdisplaybreaks

% \subsection{Convex Functions}
% \begin{lemma}[\citep{nesterov2018lectures}]\label{lem:nesterov_convex_smooth}
%     Suppose function $f(\cdot)$ is $L$-smooth and convex, for all $x,y$, it holds that
%     \begin{align}\label{eq:nesterov_convex_smooth}
%         \frac{1}{L}\norm{\nabla f(x) - \nabla f(y)} \le f(x) - f(y) - \dotprod{\nabla f(y), x-y}.
%     \end{align}
% \end{lemma}

% TODO

\section{Experimental Details}
\label{appendix:exp_details}

The detailed learning rate schedule forms and grid search hyperparameters for ridge regression in Section~\ref{sec:exp_ridge_regression} are listed in Table~\ref{tab:ridge_hyperparam}.

\begin{table}[h!]
    \caption{Grid-search hyperparameter choices for ridge regression.}
    \label{tab:ridge_hyperparam}
  \small
  \centering
    \begin{tabular}{ccc}
      \toprule 

      Scheduler
      & Form
      & \makecell{Hyperparameter choices}
      \\
      \midrule
      Constant
      & $\eta_t = \eta_0$
      & -
      \\

      \midrule
      Step decay
      & \makecell{
        $\eta_t = \eta_0 \cdot \gamma^\ell$,
        \\
        if $t \in \left[\ell, \ell+1\right) \cdot \frac{T}{\numstage}$
      }
      & \makecell{
        $\numstage \in \{2, 3, 4, 5\}$
        \\
        $\gamma \in \{ \frac{1}{2}, \frac{1}{4}, \frac{1}{8} \}$
      }
      \\
      \bottomrule
    \end{tabular}
\end{table}

\section{Proof for Section~\ref{sec:sgd_suboptimal}: Suboptimality of SGD}
\label{appendix:proof_sgd_suboptimal}

\theoremsgdlowerbound*
\begin{proof}
  Consider the case of
  $\kappa \ge 4$,
  $\bw_0 = c_0 \cdot \bI$,
  $d \ge \kappa + 1$,
  $\bH(\xi) \equiv \bH = \diag(L, \mu, \mu, \dots, \mu)$ and
  $\bb(\xi) \equiv \bzero$, then according to SGD's update formula in Eqn.~\eqref{eq:sgd_update}
  \begin{align*}
    \bw_{t+1} = \bw_{t} - \frac{\eta_t }{|\mathcal{B}_t|}
    \sum_{\xi \in \mathcal{B}_t} \nabla_{\bw} f(\bw_t, \xi)
    =
    (\bI - \eta_t \bH) \bw_{t},
  \end{align*}
  we have the update of $j$-th ($j \ge 2$) entry being
  \begin{align*}
     w_{t+1,j}
     =&
     (1 - \eta_t \lambda_j) w_{t,j}
     =
     (1 - \eta_t \mu) w_{t,j}
     \ge
     \left(1 - \frac{2\mu}{L}\right) w_{t,j}
     =
     \left(1 - \frac{2}{\kappa}\right) w_{t,j}
     \ge
     \exp\left(-\frac{4}{\kappa}\right) w_{t,j},
  \end{align*}
  where the first inequality comes from $\eta_t \le 2/L$ and the second inequality is entailed by $1 - x \ge \exp(-2x)$ for $x = 2/\kappa \in [0, 1/2]$, since for $\forall x \in [0, 1/2]$
  \begin{align*}
    &g(x) \triangleq \ln(1 - x) - \ln(\exp(-2x)),
    \quad
    g(0) = 0,
    \quad
    \frac{\partial g(x)}{\partial x}
    = -\frac{1}{1-x} + 2
    \ge 0
    \\
    \Rightarrow \quad&
    g(x) \ge g(0) = 0
    \quad
    \Rightarrow
    \ln(1-x) \ge \ln(\exp(-2x))
    \quad
    \Rightarrow
    1-x \ge \exp(-2x).
  \end{align*}
  It follows,
  \begin{align*}
    f(\bw_T) - f(\bw_*)
    =& f(\bw_T)
    = \frac{1}{2} \bw_T^\top \bH \bw_T
    =
    \frac{1}{2}
    \left(
      L w_{T,1}^2 + \mu \sum_{j=2}^d w_{T,j}^2
    \right)
    \\
    \ge&
    \frac{\mu}{2} \sum_{j=2}^d w_{T,j}^2
    =
    \frac{\mu}{2} \sum_{j=2}^d w_{0,j}^2
    \prod_{t=0}^{T-1} (1 - \eta_t \lambda_j)^2
    \\
    \ge&
    \frac{\mu}{2} \sum_{j=2}^d w_{0,j}^2
    \exp\left(-\frac{8T}{\kappa}\right)
    \\
    =&
    \frac{\mu}{2} (d-1)
    \cdot c_0^2
    \cdot \exp\left(-\frac{8T}{\kappa}\right)
    =
    \mu (d-1)
    \cdot
    \frac{f(\bw_0) - f(\bw_*)}{L + \mu (d-1)}
    \cdot \exp\left(-\frac{8T}{\kappa}\right)
    \\
    =&
    \frac{f(\bw_0) - f(\bw_*)}{\frac{L}{\mu (d-1)} + 1}
    \cdot \exp\left(-\frac{8T}{\kappa}\right)
    \\
    \ge&
    \frac{f(\bw_0) - f(\bw_*)}{2}
    \cdot \exp\left(-\frac{8T}{\kappa}\right).
  \end{align*}
\end{proof}

\section{Proof for Section~\ref{sec:sgdm}: Acceleration with Stochastic Heavy Ball}
\label{appendix:proof_sgdm}

We provide the broad stroke of our proofs here. The ultimate goal is bounding $\EE[f(\bw_T) - f(\bw_*)]$ in terms of $T, \kappa$ and other fixed model parameters. To achieve this goal, we conduct the whole analysis in a layer-by-layer fashion. In each layer, we translate the terms in the previous layer into the terms of the current layer, specifically
\begin{itemize}
  \item Appendix~\ref{appendix:bias_var_decomp}: $\EE[f(\bw) - f(\bw_*)] \Rightarrow \norm{\bT_{t,j}^k}$ for some special $2\times2$ matrix $\bT_{t,j}$ via bias-variance decomposition.
  \item Appendix~\ref{appendix:matrix_product_to_spectral_radius}:
  $\norm{\bT_{t+1,j}\bT_{t+2,j}...\bT_{t+k,j}} \Rightarrow \rho(\bT_{t+k,j})^k$ via analyzing the property of $\norm{\bT_{t,j}^k}$ and $\norm{\bT_{t+1,j}\bT_{t+2,j}...\bT_{t+k,j}}$.
  \item Appendix~\ref{appendix:spetral_raidus_to_parameters}:
  $\rho(\bT_{t,j})^k \Rightarrow \{ \eta_t, \beta, \lambda_j\}$ via analyzing the property of $\rho(\bT_{t,j})$.
  % \item Appendix~\ref{appendix:stagewise_contraction_and_var}
  % $\{ \eta_t, \beta_t, \lambda_j\} \Rightarrow \{ \eta'_\ell, \beta'_\ell, \lambda_j, \stagelength\}$ via focusing on general multistage schedulers with only weak constraints on hyperparameters.
  \item Appendix~\ref{appendix:speify_to_step_decay}
  $\{ \eta_t, \beta, \lambda_j\} \Rightarrow \{ \eta'_\ell, \beta, \lambda_j\} \Rightarrow \{ T, \kappa, \dots \}$ via specializing the scheduler with multistage scheduler step decay.
\end{itemize}

\subsection{Bias Variance Decomposition: $\EE[f(\bw) - f(\bw_*)] \Rightarrow \norm{\bT_{t,j}^k}$}
\label{appendix:bias_var_decomp}

The section presents the lemmas that decompose the loss into bias and variance, expressing them in terms of the norm of product of a series of special $2\times2$ matrices $\norm{\prod_t \bT_{t,j}}$'s, specifically $\norm{\bT_{T-1,j} \bT_{T-2,j} \dots \bT_{\tau+1,j}}$ for some $\tau=0,1,\dots,T-1$. For simplicity, we only consider the case of batch size $M=1$. The case of larger batches is equivalent to replacing noise term $\sigma^2$ with $\sigma^2 / M$.

\begin{lemma}
\label{lem:loss-is-H-norm}
\textbf{(Quadratic excess risk is $\bH$-norm)}
Given a quadratic objective $f(\bw)$, we have
\begin{equation}
\label{eq:loss-is-H-norm}
\begin{aligned}
  f(\bw) - f(\bw_*)
  = \frac{1}{2} \norm{\bw - \bw_*}_\bH^2
  \triangleq \frac{1}{2} (\bw - \bw_*)^\top \bH (\bw - \bw_*)
\end{aligned}
\end{equation}
\end{lemma}
\begin{proof}
It holds that
\begin{align*}
  f(\bw) - f(\bw_*)
  \overset{\eqref{eq:problem_setup}}{=}&
  \EE_\xi\left[
    \frac{1}{2} \bw^\top \bH(\xi) \bw - \bb(\xi)^\top \bw
  \right]
  -
  \EE_\xi\left[
    \frac{1}{2} \bw_*^\top \bH(\xi) \bw_* - \bb(\xi)^\top \bw_*
  \right]
  \\
  =&
  \left[
    \frac{1}{2} \bw^\top \EE[\bH(\xi)] \bw
    - \EE[\bb(\xi)]^\top \bw
  \right]
  -
  \left[
    \frac{1}{2} \bw_*^\top \EE[\bH(\xi)] \bw_*
    - \EE[\bb(\xi)]^\top \bw_*
  \right]
  \\
  \overset{\eqref{eq:optima}}{=}&
  \left[
    \frac{1}{2} \bw^\top \bH \bw - \bb^\top \bw
  \right]
  -
  \left[
    \frac{1}{2} \bw_*^\top \bH \bw_* - \bb^\top \bw_*
  \right]
  \\
  \overset{\eqref{eq:optima}}{=}&
  \left[
    \frac{1}{2} \bw^\top \bH \bw
    - (\bH \bw_*)^\top \bw
  \right]
  -
  \left[
    \frac{1}{2} \bw_*^\top \bH \bw_*
    - (\bH \bw_*)^\top \bw_*
  \right]
  \\
  =&
  \frac{1}{2} \left[
    \bw^\top \bH \bw
    - \bw^\top \bH \bw_*
    - \bw_*^\top \bH \bw
    + \bw_*^\top \bH \bw_*
  \right]
  \\
  =&
  \frac{1}{2} (\bw - \bw_*)^\top \bH (\bw - \bw_*).
\end{align*}
Here the fifth equality comes from $\bH = \bH^\top$ being a symmetric matrix, and $\bw_*^\top \bH \bw = (\bw_*^\top \bH \bw)^\top = \bw^\top \bH \bw_*$ being a scalar.
\end{proof}

\begin{lemma}
\label{lem:bias-var-decomp}
\textbf{(Bias-variance decomposition)}
Given a quadratic function $f(\bw)$, if batch size $M=1$, after running $T$ iterations of Algorithm~\ref{alg:multistage_sgdm}, we have the last iterate $\bw_T$ satisfying
\begin{equation}
\label{eq:bias-var-decomp}
\begin{aligned}
  &\EE\left[\norm{\bw_T - \bw_*}_\bH^2\right]
  \\
  =&
  \EE\left[
    \begin{bmatrix}
      \bw_0 - \bw_* \\
      \bw_{-1} - \bw_*
    \end{bmatrix}^\top
    \bM_0^\top \bM_1^\top \dots \bM_{T-1}^\top
    \begin{bmatrix}
      \bH & \bO \\
      \bO & \bO
    \end{bmatrix}
    \bM_{T-1} \bM_{T-2} \dots \bM_0
    \begin{bmatrix}
      \bw_0 - \bw_* \\
      \bw_{-1} - \bw_*
    \end{bmatrix}
  \right]
  \\
  &+
  \sum_{\tau=0}^{T-1}
    \EE\left[
      \eta_\tau^2
      \begin{bmatrix}
        \bn_\tau \\
        \bzero
      \end{bmatrix}^\top
      \bM_{\tau+1}^\top \bM_{\tau+2}^\top \dots \bM_{T-1}^\top
      \begin{bmatrix}
        \bH & \bO \\
        \bO & \bO
      \end{bmatrix}
      \bM_{T-1} \bM_{T-2} \dots \bM_{\tau+1}
      \begin{bmatrix}
        \bn_\tau \\
        \bzero
      \end{bmatrix}
    \right],
\end{aligned}
\end{equation}
where $\bw_{-1}=\bw_0+\bv_0$ and
\begin{equation}
\label{eq:def_M_t}
\begin{aligned}
  \bM_t \triangleq
  \begin{bmatrix}
    (1 + \beta) \bI - \eta_t \bH & -\beta \bI \\
    \bI & \bO
  \end{bmatrix}.
\end{aligned}
\end{equation}
\end{lemma}
\begin{proof}
  Denote
  \begin{equation}
  \label{eq:bwp_bnp_bHp_def}
  \begin{aligned}
    \bwp_t \triangleq
      \begin{bmatrix}
        \bw_t - \bw_* \\ \bw_{t-1} - \bw_*
      \end{bmatrix}
      \in \RR^{2d},
    \quad
    \bnp_t \triangleq
      \begin{bmatrix}
        \bn_t \\ \bzero
      \end{bmatrix}
      \in \RR^{2d},
    \quad
    \bHp \triangleq
      \begin{bmatrix}
        \bH & \bO \\
        \bO & \bO
      \end{bmatrix}
      \in \RR^{2d\times2d}
  \end{aligned}
  \end{equation}
  as the extended parameter difference, extended gradient noise and extended Hessian matrix, then according to momentum's update formula in Eqn.~\eqref{eq:sgdm_update}, or line~\ref{alg:line:sgdm_update_begin}-\ref{alg:line:sgdm_update_end} in Algorithm~\ref{alg:multistage_sgdm} with batch size $M=1$, we have
  \begin{align*}
    \bv_{t+1} =& \beta \bv_t + \eta_t \nabla_\bw f(\bw_t, \xi)
    \\
    \bw_{t+1} =& \bw_t - \bv_{t+1}
    \\
    \Rightarrow \quad
    \bw_{t+1} - \bw_*
    =&
    \bw_t - \bw_* - \bv_{t+1}
    \\
    =&
    \bw_t - \bw_* - \left(\beta \bv_t + \eta_t \nabla_\bw f(\bw_t, \xi)\right)
    \\
    =&
    \bw_t - \bw_* - \beta \left(\bw_{t-1} - \bw_t\right) - \eta_t \nabla_\bw f(\bw_t, \xi)
    \\
    \overset{\eqref{eq:nt_def}}{=}&
    \bw_t - \bw_* - \beta \left(\bw_{t-1} - \bw_t\right) - \eta_t \left(\nabla_\bw f(\bw_t) - \bn_t\right)
    \\
    \overset{\eqref{eq:optima}}{=}&
    \bw_t - \bw_* - \beta \left(\bw_{t-1} - \bw_t\right) - \eta_t \left(\bH \left(\bw_t - \bw_*\right)  - \bn_t\right)
    \\
    =&
    \bw_t - \bw_*
    - \beta \left[
      \left(\bw_{t-1} - \bw_*\right)
      - \left(\bw_t - \bw_*\right)
    \right]
    - \eta_t \left(\bH \left(\bw_t - \bw_*\right)  - \bn_t\right)
    \\
    =&
    \left[
      \left(1 + \beta\right) \bI - \eta_t \bH
    \right]
    \left(
      \bw_t - \bw_*
    \right)
    - \beta \bI
    \left(
      \bw_{t-1} - \bw_*
    \right)
    + \eta_t \bn_t
    \\
    \Rightarrow\quad
    \bwp_{t+1}
    =&
    \begin{bmatrix}
      \bw_{t+1} - \bw_* \\
      \bw_t - \bw_*
    \end{bmatrix}
    =
    \begin{bmatrix}
      (1 + \beta) \bI - \eta_t \bH & -\beta \bI \\
      \bI & \bO
    \end{bmatrix}
    \begin{bmatrix}
      \bw_t - \bw_* \\
      \bw_{t-1} - \bw_*
    \end{bmatrix}
    + \eta_t
      \begin{bmatrix}
        \bn_t \\
        \bzero
      \end{bmatrix}
    \\
    &=
    \bM_t \bwp_t + \eta_t \bnp_t.
  \end{align*}

  It follows
  \begin{align*}
    \bwp_{t+1}
    =& \bM_t \bwp_t + \eta_t \bnp_t
    \\
    =& \bM_t \bM_{t-1} \bwp_{t-1}
      + \eta_{t-1} \bM_t \bnp_{t-1}
      + \eta_t \bnp_t
    \\
    =& \bM_t \bM_{t-1} \bM_{t-2} \bwp_{t-2}
      + \eta_{t-2} \bM_t \bM_{t-1} \bnp_{t-2}
      + \eta_{t-1} \bM_t \bnp_{t-1}
      + \eta_t \bnp_t
    \\
    =& \dots
    \\
    =&
    \bM_t \bM_{t-1} \dots \bM_0 \bwp_0
    + \sum_{\tau=0}^t \left(
        \eta_\tau
        \bM_t \bM_{t-1} \dots \bM_{\tau+1}
        \bnp_\tau
      \right).
  \end{align*}
  where $\bwp_0 = \begin{bmatrix}
          \bw_0-\bw_* \\ \bw_{-1}-\bw_*
      \end{bmatrix}$
  and $\bw_{-1}=\bw_0+\bv_0$ which is associated with iteration $t=0$.

  We can decompose the above process into two parts
  \begin{equation}
  \label{eq:bias_var_update}
  \begin{aligned}
    \bwpb_{t+1} =& \bM_t \bwpb_t
    \quad \quad \quad \quad \mbox{ with }
    \bwpb_0 = \bwp_0
    \\
    \bwpv_{t+1} =& \bM_t \bwpv_t + \eta_t \bnp_t
    \quad \mbox{ with } \bwpv_0 = \bzero,
  \end{aligned}
  \end{equation}
  since
  \begin{equation}
  \label{eq:bias_var_expand}
  \begin{aligned}
    \bwpb_{t+1}
    =&
    \bM_t \bM_{t-1} \dots \bM_0 \bwpb_0
    =
    \bM_t \bM_{t-1} \dots \bM_0 \bwp_0
    \\
    \bwpv_{t+1}
    =&
    \sum_{\tau=0}^t \left(
      \eta_\tau
      \bM_t \bM_{t-1} \dots \bM_{\tau+1}
      \bnp_\tau
    \right)
    \\
    \Rightarrow \quad
    \bwp_{t+1} =& \bwpb_{t+1} + \bwpv_{t+1}.
  \end{aligned}
  \end{equation}

  Furthermore, we have
  \begin{equation}
  \label{eq:w_extended_v_expect}
  \begin{aligned}
    \EE\left[
      \left(\bwpb_{t+1}\right)^\top \bHp \bwpv_{t+1}
    \right]
    =&
    \left(\bwpb_{t+1} \right)^\top \bHp
    \EE\left[\bwpv_{t+1}\right]
    \\
    \overset{\eqref{eq:bias_var_expand}}{=}&
    \left(\bwpb_{t+1}\right)^\top \bHp
    \EE\left[
      \sum_{\tau=0}^t \left(
        \eta_\tau
        \bM_t \bM_{t-1} \dots \bM_{\tau+1}
        \bnp_\tau
      \right)  
    \right]
    \\
    =&
    \left(\bwpb_{t+1}\right)^\top \bHp
    \sum_{\tau=0}^t \left(
      \eta_\tau
      \bM_t \bM_{t-1} \dots \bM_{\tau+1}
      \EE\left[\bnp_\tau\right]
    \right)
    \\
    \overset{\eqref{eq:unbiased_noise}}{=}&
    \bzero
  \end{aligned}
  \end{equation}
  and
  \begin{equation}
  \label{eq:w_extended_v_cov}
  \begin{aligned}
    &\EE\left[
      \left(\bwpv_{t+1}\right)^\top \bHp \bwpv_{t+1}
    \right]
    \\
    =&
    \EE\left[
      \left(
        \sum_{\tau=0}^t \left(
          \eta_\tau
          \bM_t \bM_{t-1} \dots \bM_{\tau+1}
          \bnp_\tau
        \right)
      \right)^\top
      \bHp
      \left(
        \sum_{\tau'=0}^t \left(
          \eta_{\tau'}
          \bM_t \bM_{t-1} \dots \bM_{\tau'+1}
          \bnp_{\tau'}
        \right)
      \right)
    \right]
    \\
    =&
    \sum_{\tau=0, \tau'=0}^t \EE\left[ 
      \eta_\tau \eta_{\tau'}
      \bnp_\tau^\top
      \bM_{\tau+1}^\top \bM_{\tau+2}^\top \dots \bM_t^\top
      \bHp
      \bM_t \bM_{t-1} \dots \bM_{\tau'+1}
      \bnp_{\tau'}
    \right]
    \\
    =&
    \sum_{\tau=0}^t \EE\left[ 
      \eta_\tau^2
      \bnp_\tau^\top
      \bM_{\tau+1}^\top \bM_{\tau+2}^\top \dots \bM_t^\top
      \bHp
      \bM_t \bM_{t-1} \dots \bM_{\tau+1}
      \bnp_{\tau}
    \right]
    \\
    &+
    \sum_{\tau=0, \tau'=0, \tau \neq \tau'}^t
    \EE\left[ 
      \eta_\tau \eta_{\tau'}
      \bnp_\tau^\top
      \bM_{\tau+1}^\top \bM_{\tau+2}^\top \dots \bM_t^\top
      \bHp
      \bM_t \bM_{t-1} \dots \bM_{\tau'+1}
      \bnp_{\tau'}
    \right]
    \\
    =&
    \sum_{\tau=0}^t \EE\left[ 
      \eta_\tau^2
      \bnp_\tau^\top
      \bM_{\tau+1}^\top \bM_{\tau+2}^\top \dots \bM_t^\top
      \bHp
      \bM_t \bM_{t-1} \dots \bM_{\tau+1}
      \bnp_{\tau}
    \right]
    + 0,
  \end{aligned}
  \end{equation}
  where the last equality is because
  \begin{align*}
    \bnp_\tau
    \overset{\eqref{eq:bwp_bnp_bHp_def}}{=}
    \begin{bmatrix} \bn_\tau \\ 0 \end{bmatrix}
    \mbox{ and }
    \bnp_{\tau'}
    \overset{\eqref{eq:bwp_bnp_bHp_def}}{=}
    \begin{bmatrix} \bn_{\tau'} \\ 0 \end{bmatrix}
  \end{align*}
  are pairwise independent and have mean $\bzero$ given Assumption~\ref{ass:independent_noise} and~\ref{ass:unbiased_noise}.

  Then we obtain
  \begin{align*}
    &\EE \norm{\bw_T - \bw_*}_\bH^2
    \\
    \overset{\eqref{eq:loss-is-H-norm}}{=}&
      \left(\bw_T - \bw_*\right) \bH \left(\bw_T - \bw_*\right)
    =
      \begin{bmatrix}
        \bw_T - \bw_* \\
        \bw_{T-1} - \bw_*
      \end{bmatrix}^\top
      \begin{bmatrix}
        \bH & \bO \\
        \bO & \bO
      \end{bmatrix}
      \begin{bmatrix}
        \bw_T - \bw_* \\
        \bw_{T-1} - \bw_*
      \end{bmatrix}
    \\
    \overset{\eqref{eq:bwp_bnp_bHp_def}}{=}&
      \EE\left[
        \bwp_T^\top \bHp \bwp_T
      \right]
    \overset{\eqref{eq:bias_var_expand}}{=}
      \EE\left[
        \left(
          \bwpb_T + \bwpv_T
        \right)^\top
        \bHp
        \left(
          \bwpb_T + \bwpv_T
        \right)
      \right]
    \\
    =&
    \EE\left[
      \left(\bwpb_T\right)^\top
      \bHp
      \left(\bwpb_T\right)
      +
      \left(\bwpv_T\right)^\top
      \bHp
      \left(\bwpv_T\right)
      +
      2 \left(\bwpb_T\right)^\top
      \bHp
      \left(\bwpv_T\right)
    \right]
    \\
    \overset{\eqref{eq:w_extended_v_expect}}{=}&
    \EE\left[
      \left(\bwpb_T\right)^\top
      \bHp
      \left(\bwpb_T\right)
    \right]
    +
    \EE\left[
      \left(\bwpv_T\right)^\top
      \bHp
      \left(\bwpv_T\right)
    \right]
    \\
    \overset{\eqref{eq:bias_var_expand},\eqref{eq:w_extended_v_cov}}{=}&
    \EE\left[
      \bwp_0
      \bM_0^\top \bM_1^\top \dots \bM_{T-1}^\top
      \bHp
      \bM_{T-1} \bM_{T-2} \dots \bM_0
      \bwp_0
    \right]
    \\
    &+
    \sum_{\tau=0}^{T-1}
      \EE\left[
        \eta_\tau^2
        \bnp_{\tau}^\top
        \bM_{\tau+1}^\top \bM_{\tau+2}^\top \dots \bM_{T-1}^\top
        \bHp
        \bM_{T-1} \bM_{T-2} \dots \bM_{\tau+1}
        \bnp_{\tau}
      \right]
    \end{align*}
  Here the fourth equality is because $(\bwpb_{t+1})^\top \bHp (\bwpv_{t+1})$ is a scalar and $\bHp$ is symmetric. Replacing the extended terms $\bwp$, $\bnp$, $\bHp$ with their definitions in Eqn.~\eqref{eq:bwp_bnp_bHp_def}, we get the desired form in Eqn.~\eqref{eq:bias-var-decomp}.
\end{proof}

\begin{lemma}
\label{lem:M_t_decomp}
\textbf{(Decomposing $\bM_t$ into a block diagonal matrix)}
Given a matrix $\bM_t \in \RR^{2d\times2d}$ defined in Eqn.~\eqref{eq:def_M_t}, we have
\begin{equation}
\label{eq:M_t_decomp}
\begin{aligned}
  \bM_t
  =
  \bV \bPi
  \begin{bmatrix}
    \bT_{t,1} & & & \\
    & \bT_{t,2} & & \\
    & & \ddots & \\
    & & & \bT_{t,d}
  \end{bmatrix}
  \bPi^\top \bV^\top,
\end{aligned}
\end{equation}
where
\begin{equation}
\label{eq:T_tj_def}
\begin{aligned}
  \bT_{t,j} \triangleq
  \begin{bmatrix}
    1 + \beta - \eta_t \lambda_j & -\beta \\
    1 & 0
  \end{bmatrix} \in \RR^{2\times2}
\end{aligned}
\end{equation}
and orthogonal matrices
\begin{equation}
\label{eq:Pi_V_def}
\begin{aligned}
  \bPi \triangleq
  \begin{bmatrix}
    \be_1 & \bzero & \be_2 & \bzero & \dots & \be_d & \bzero \\
    \bzero & \be_1 & \bzero & \be_2 & \dots & \bzero & \be_d
  \end{bmatrix} \in \RR^{2d\times2d},
  \bV \triangleq
  \begin{bmatrix}
    \bU & \bO \\
    \bO & \bU
  \end{bmatrix} \in \RR^{2d\times2d},
\end{aligned}
\end{equation}
given the eigendecomposition of $\bH$ being
\begin{equation}
\label{eq:H_eigendecomp}
\begin{aligned}
  \bH = \bU \bLambda \bU^\top \in \RR^{d\times d}, \quad\left(\bU^\top \bU = \bI\right)
\end{aligned}
\end{equation}
and standard unit vectors/standard basis being
\begin{equation}
\label{eq:e_i_def}
\begin{aligned}
  \be_i =
  \begin{bmatrix}\strut\smash{
    0 \dots 0 \underbrace{1}_{i-th} 0 \dots 0
  }\end{bmatrix}^\top \in \RR^{d\times1}
\end{aligned}
\end{equation}
\end{lemma}
\begin{proof}
\begin{align*}
  &\bPi^\top \bV^\top \bM_t \bV \bPi
  \\
  \overset{\eqref{eq:def_M_t}}{=}&
  \bPi^\top \bV^\top
  \begin{bmatrix}
    (1 + \beta) \bI - \eta_t \bH & -\beta \bI \\
    \bI & \bO
  \end{bmatrix}
  \bV \bPi
  \\
  \overset{\eqref{eq:Pi_V_def}}{=}&
  \bPi^\top
  \begin{bmatrix}
    \bU^\top & \bO \\
    \bO & \bU^\top
  \end{bmatrix}
  \begin{bmatrix}
    (1 + \beta) \bI - \eta_t \bH & -\beta \bI \\
    \bI & \bO
  \end{bmatrix}
  \begin{bmatrix}
    \bU & \bO \\
    \bO & \bU
  \end{bmatrix}
  \bPi
  \\
  \overset{\eqref{eq:H_eigendecomp}}{=}&
  \bPi^\top
  \begin{bmatrix}
    (1 + \beta) \bI - \eta_t \bLambda & -\beta \bI \\
    \bI & \bO
  \end{bmatrix}
  \bPi
  \\
  \overset{\eqref{eq:Pi_V_def}}{=}&
  \begin{bmatrix}
    \be_1^\top & \bzero^\top \\
    \bzero^\top & \be_1^\top \\
    \vdots & \\
    \be_d^\top & \bzero^\top \\
    \bzero^\top & \be_d^\top
  \end{bmatrix}
  \begin{bmatrix}
    (1 + \beta) \bI - \eta_t \bLambda & -\beta \bI \\
    \bI & \bO
  \end{bmatrix}
  \begin{bmatrix}
    \be_1 & \bzero & \be_2 & \bzero & \dots & \be_d & \bzero \\
    \bzero & \be_1 & \bzero & \be_2 & \dots & \bzero & \be_d
  \end{bmatrix}
  \\
  \overset{\eqref{eq:e_i_def}}{=}&
  \begin{bmatrix}
    \be_1^\top & \bzero^\top \\
    \bzero^\top & \be_1^\top \\
    \vdots & \\
    \be_d^\top & \bzero^\top \\
    \bzero^\top & \be_d^\top
  \end{bmatrix}
  \begin{bmatrix}
    (1 + \beta - \eta_t \lambda_1) \be_1
    & -\beta \be_1
    & \dots
    & (1 + \beta - \eta_t \lambda_d) \be_d
    & -\beta \be_d
    \\
    \be_1
    & \bzero
    & \dots
    & \be_d
    & \bzero
  \end{bmatrix}
  \\
  =&
  \begin{bmatrix}
    \bS_{1,1} & \bS_{1,2} & \dots & \bS_{1,d} \\
    \bS_{2,1} & \bS_{2,2} & \dots & \bS_{2,d} \\
    \vdots & \vdots & \ddots & \vdots \\
    \bS_{d,1} & \bS_{d,2} & \dots & \bS_{d,d}
  \end{bmatrix},
  \mbox{ where }
  \bS_{i,j} =
  \begin{bmatrix}
    (1 + \beta - \eta_t \lambda_j) \be_i^\top \be_j
    & -\beta \be_i^\top \be_j
    \\
    \be_i^\top \be_j
    & 0
  \end{bmatrix}
  \\
  \overset{\eqref{eq:T_tj_def}}{=}&
  \begin{bmatrix}
    \bT_{t,1} & & & \\
    & \bT_{t,2} & & \\
    & & \ddots & \\
    & & & \bT_{t,d}
  \end{bmatrix}
\end{align*}
Since
\begin{equation}
\label{eq:Pi_V_are_orthogonal}
\begin{aligned}
  \bPi^\top \bPi
  \overset{\eqref{eq:Pi_V_def}}{=}&
  \begin{bmatrix}
    \be_1^\top & \bzero^\top \\
    \bzero^\top & \be_1^\top \\
    \vdots & \\
    \be_d^\top & \bzero^\top \\
    \bzero^\top & \be_d^\top
  \end{bmatrix}
  \begin{bmatrix}
    \be_1 & \bzero & \be_2 & \bzero & \dots & \be_d & \bzero \\
    \bzero & \be_1 & \bzero & \be_2 & \dots & \bzero & \be_d
  \end{bmatrix}
  \overset{\eqref{eq:e_i_def}}{=} \bI_{2d\times2d}
  \\
  \bV^\top \bV
  \overset{\eqref{eq:Pi_V_def}}{=}&
  \begin{bmatrix}
    \bU^\top & \bO \\
    \bO & \bU^\top
  \end{bmatrix}
  \begin{bmatrix}
    \bU & \bO \\
    \bO & \bU
  \end{bmatrix}
  \overset{\eqref{eq:H_eigendecomp}}{=}
  \bI_{2d\times2d}
\end{aligned}
\end{equation}
are both orthogonal matrices, we thereby have
\begin{align*}
  \bM_t
  =
  \bV \bPi
  \begin{bmatrix}
    \bT_{t,1} & & & \\
    & \bT_{t,2} & & \\
    & & \ddots & \\
    & & & \bT_{t,d}
  \end{bmatrix}
  \bPi^\top \bV^\top.
\end{align*}
\end{proof}

\begin{lemma}
\label{lem:bound_var_with_T_norm}
\textbf{(Bound Variance with $\norm{\bT_{t,j}^k}$)}
Assuming batch size $M=1$, we have
\begin{equation}
\label{eq:bound_var_with_T_norm}
\begin{aligned}
  &\sum_{\tau=0}^{T-1}
    \EE\left[
      \eta_\tau^2
      \begin{bmatrix}
        \bn_\tau \\
        \bzero
      \end{bmatrix}^\top
      \bM_{\tau+1}^\top \bM_{\tau+2}^\top \dots \bM_{T-1}^\top
      \begin{bmatrix}
        \bH & \bO \\
        \bO & \bO
      \end{bmatrix}
      \bM_{T-1} \bM_{T-2} \dots \bM_{\tau+1}
      \begin{bmatrix}
        \bn_\tau \\
        \bzero
      \end{bmatrix}
    \right]
  \\
  \le&
  \sigma^2 \sum_{j=1}^d \lambda_j^2
    \sum_{\tau=0}^{T-1} \eta_\tau^2
      \norm{\bT_{T-1,j} \bT_{T-2,j} \dots \bT_{\tau+1,j}}^2,
\end{aligned}
\end{equation}
where $\bM_t$ is defined in Eqn.~\eqref{eq:def_M_t} and $\bT_{t,j} \in \RR^{2\times2}$ is defined in~\eqref{eq:T_tj_def}.
\end{lemma}
\begin{proof}
  Similarly, we define $\bnp$ and $\bHp$ as in Eqn.~\eqref{eq:bwp_bnp_bHp_def}. Notice that $\bHp$ is a positive semi-definite matrix since if the eigenvalue decomposition of Hessian $\bH = \bU \bLambda \bU^\top$, we have
  \begin{align*}
    \bHp =
    \begin{bmatrix}
      \bH & \bO \\
      \bO & \bO
    \end{bmatrix}
    =
    \begin{bmatrix}
      \bU & \bO \\
      \bO & \bU
    \end{bmatrix}
    \begin{bmatrix}
      \bLambda & \bO \\
      \bO & \bO
    \end{bmatrix}
    \begin{bmatrix}
      \bU & \bO \\
      \bO & \bU
    \end{bmatrix}^\top.
  \end{align*}

  Therefore, $\bHp^{1/2}$ is well-defined. Denote
  \begin{equation}
  \label{eq:def_A_tau}
  \begin{aligned}
    \bA_\tau \triangleq
    \bHp^{1/2} \bM_{T-1} \bM_{T-2} \dots \bM_{\tau+1},
  \end{aligned}
  \end{equation}
  then
  \begin{equation}
  \label{eq:A_tau_transpose}
  \begin{aligned}
    \bA_\tau^\top =
    \bM_{\tau+1}^\top \bM_{\tau+2}^\top \dots \bM_{T-1}^\top \left(\bHp^{1/2}\right)^\top
    =
    \bM_{\tau+1}^\top \bM_{\tau+2}^\top \dots \bM_{T-1}^\top \bHp^{1/2},
  \end{aligned}
  \end{equation}
  where the second equality is because $\bHp^{1/2}$ is symmetric.

  It follows
  \begin{align*}
    &\sum_{\tau=0}^{T-1}
      \EE\left[
        \eta_\tau^2
        \begin{bmatrix}
          \bn_\tau \\
          \bzero
        \end{bmatrix}^\top
        \bM_{\tau+1}^\top \bM_{\tau+2}^\top \dots     \bM_{T-1}^\top
        \begin{bmatrix}
          \bH & \bO \\
          \bO & \bO
        \end{bmatrix}
        \bM_{T-1} \bM_{T-2} \dots \bM_{\tau+1}
        \begin{bmatrix}
          \bn_\tau \\
          \bzero
        \end{bmatrix}
      \right]
    \\
    \overset{\eqref{eq:bwp_bnp_bHp_def}}{=}&
    \sum_{\tau=0}^{T-1}
      \EE\left[
        \eta_\tau^2
        \bnp_\tau^\top
        \bM_{\tau+1}^\top \bM_{\tau+2}^\top \dots     \bM_{T-1}^\top
        \bHp
        \bM_{T-1} \bM_{T-2} \dots \bM_{\tau+1}
        \bnp_\tau
      \right]
    \\
    \overset{\eqref{eq:def_A_tau}\eqref{eq:A_tau_transpose}}{=}&
    \sum_{\tau=0}^{T-1}
      \EE\left[
        \eta_\tau^2
        \bnp_\tau^\top
        \bA_\tau^\top
        \bA_\tau
        \bnp_\tau
      \right]
    =
    \sum_{\tau=0}^{T-1} \eta_\tau^2
      \EE\left[\tr\left(
        \bnp_\tau^\top
        \bA_\tau^\top
        \bA_\tau
        \bnp_\tau
      \right)\right]
    \mcomment{Trace of a scalar is itself}
    \\
    =&
    \sum_{\tau=0}^{T-1} \eta_\tau^2
      \EE\left[\tr\left(
        \bA_\tau
        \bnp_\tau
        \bnp_\tau^\top
        \bA_\tau^\top
      \right)\right]
    \mcomment{Cyclic property of trace}
    \\
    =&
    \sum_{\tau=0}^{T-1} \eta_\tau^2
      \tr\left(\EE\left[
        \bA_\tau
        \bnp_\tau
        \bnp_\tau^\top
        \bA_\tau^\top
      \right]\right)
    \mcomment{Linearity of expectation}
    \\
    \overset{\eqref{eq:independent_matrix_prod}}{=}&
    \sum_{\tau=0}^{T-1} \eta_\tau^2
      \tr\left(
        \bA_\tau
        \EE\left[\bnp_\tau \bnp_\tau^\top \right]
        \bA_\tau^\top
      \right)
    \overset{\eqref{eq:bwp_bnp_bHp_def}}{=}
    \sum_{\tau=0}^{T-1} \eta_\tau^2
      \tr\left(
        \bA_\tau
        \EE
          \begin{bmatrix}
            \bn_\tau \bn_\tau^\top & \bO \\
            \bO & \bO
          \end{bmatrix}
        \bA_\tau^\top
      \right)
    \\
    \le&
    \sum_{\tau=0}^{T-1} \eta_\tau^2
      \tr\left(
        \bA_\tau
        \begin{bmatrix}
          \sigma^2 \bH & \bO \\
          \bO & \bO
        \end{bmatrix}
        \bA_\tau^\top
      \right)
    =
    \sigma^2 \sum_{\tau=0}^{T-1} \eta_\tau^2
      \tr\left(
        \bA_\tau
        \begin{bmatrix}
          \bH & \bO \\
          \bO & \bO
        \end{bmatrix}
        \bA_\tau^\top
      \right)
    \\
    =&
    \sigma^2 \sum_{\tau=0}^{T-1} \eta_\tau^2
      \tr\left(
        \bA_\tau^\top
        \bA_\tau
        \begin{bmatrix}
          \bH & \bO \\
          \bO & \bO
        \end{bmatrix}
      \right)
    \mcomment{Cyclic property of trace}
    \\
    \overset{\eqref{eq:def_A_tau}\eqref{eq:A_tau_transpose}}{=}&
    \sigma^2 \sum_{\tau=0}^{T-1} \eta_\tau^2
      \tr\left(
        \bM_{\tau+1}^\top \bM_{\tau+2}^\top \dots \bM_{T-1}^\top \bHp \bM_{T-1} \bM_{T-2} \dots \bM_{\tau+1}
        \begin{bmatrix}
          \bH & \bO \\
          \bO & \bO
        \end{bmatrix}
      \right)
    \\
    \overset{\eqref{eq:bwp_bnp_bHp_def}}{=}&
    \sigma^2 \sum_{\tau=0}^{T-1} \eta_\tau^2
      \tr\left(
        \bM_{\tau+1}^\top \bM_{\tau+2}^\top \dots \bM_{T-1}^\top
        \bHp
        \bM_{T-1} \bM_{T-2} \dots \bM_{\tau+1}
        \bHp
      \right)
  \end{align*}
  where the inequality is because
  \begin{align*}
    \EE \begin{bmatrix}
      \bn_\tau \bn_\tau^\top & \bO \\
      \bO & \bO
    \end{bmatrix}
    \preceq
    \begin{bmatrix}
      \sigma^2 \bH & \bO \\
      \bO & \bO
    \end{bmatrix}
  \end{align*}

  given $\EE[\bn_\tau \bn_\tau^\top] \preceq \sigma^2 \bH$ in Assumption~\ref{ass:anisotropic_noise}, along with basic properties of Loewner order in Lemma~\ref{lem:loewner_replace_partial_in_2x2},~\ref{lem:loewner_outside_product} and~\ref{lem:loewner_trace}.
  
  Let
  \begin{equation}
  \label{eq:bTp_def}
  \begin{aligned}
    \bTp_t \triangleq
    \begin{bmatrix}
      \bT_{t,1} & & & \\
      & \bT_{t,2} & & \\
      & & \ddots & \\
      & & & \bT_{t,d}
    \end{bmatrix},
  \end{aligned}
  \end{equation}

  we have the variance term being
  \begin{align*}
    &\sum_{\tau=0}^{T-1}
      \EE\left[
        \eta_\tau^2
        \begin{bmatrix}
          \bn_\tau \\
          \bzero
        \end{bmatrix}^\top
        \bM_{\tau+1}^\top \bM_{\tau+2}^\top \dots     \bM_{T-1}^\top
        \begin{bmatrix}
          \bH & \bO \\
          \bO & \bO
        \end{bmatrix}
        \bM_{T-1} \bM_{T-2} \dots \bM_{\tau+1}
        \begin{bmatrix}
          \bn_\tau \\
          \bzero
        \end{bmatrix}
      \right]
    \\
    \le&
    \sigma^2 \sum_{\tau=0}^{T-1} \eta_\tau^2
      \tr\left(
        \bM_{\tau+1}^\top \bM_{\tau+2}^\top \dots \bM_{T-1}^\top
        \bHp
        \bM_{T-1} \bM_{T-2} \dots \bM_{\tau+1}
        \bHp
      \right)
    \\
    \overset{\eqref{eq:M_t_decomp}}{=}&
    \sigma^2 \sum_{\tau=0}^{T-1} \eta_\tau^2
      \tr\left(
        \bV \bPi
        \bTp_{\tau+1}^\top \bTp_{\tau+2}^\top \dots \bTp_{T-1}^\top
        \bPi^\top \bV^\top
        \bHp
        \bV \bPi
        \bTp_{T-1} \bTp_{T-2} \dots \bTp_{\tau+1}
        \bPi^\top \bV^\top
        \bHp
      \right)
    \\
    =&
    \sigma^2 \sum_{\tau=0}^{T-1} \eta_\tau^2
      \tr\left(
        \bTp_{\tau+1}^\top \bTp_{\tau+2}^\top \dots \bTp_{T-1}^\top
        \left(
          \bPi^\top \bV^\top
          \bHp
          \bV \bPi
        \right)
        \bTp_{T-1} \bTp_{T-2} \dots \bTp_{\tau+1}
        \left(
          \bPi^\top \bV^\top
          \bHp
          \bV \bPi
        \right)
      \right).
  \end{align*}

  Here the last equality comes from the cyclic property of trace. Given the definition of $\bPi, \bV$ and $\bHp$ in Eqn.~\eqref{eq:Pi_V_def} and Eqn.~\eqref{eq:bwp_bnp_bHp_def}, we have
  \begin{equation}
  \label{eq:Pi'_V'_H_V_Pi}
  \begin{aligned}
    \bPi^\top \bV^\top
    \bHp
    \bV \bPi
    =&
    \begin{bmatrix}
      \be_1^\top & \bzero^\top \\
      \bzero^\top & \be_1^\top \\
      \vdots & \\
      \be_d^\top & \bzero^\top \\
      \bzero^\top & \be_d^\top
    \end{bmatrix}
    \begin{bmatrix}
      \bU^\top & \bO \\
      \bO & \bU^\top
    \end{bmatrix}
    \begin{bmatrix}
      \bH & \bO \\
      \bO & \bO
    \end{bmatrix}
    \begin{bmatrix}
      \bU & \bO \\
      \bO & \bU
    \end{bmatrix}
    \begin{bmatrix}
      \be_1 & \bzero & \be_2 & \bzero & \dots & \be_d & \bzero \\
      \bzero & \be_1 & \bzero & \be_2 & \dots & \bzero & \be_d
    \end{bmatrix}
    \\
    \overset{\eqref{eq:H_eigendecomp}}{=}&
    \begin{bmatrix}
      \be_1^\top & \bzero^\top \\
      \bzero^\top & \be_1^\top \\
      \vdots & \\
      \be_d^\top & \bzero^\top \\
      \bzero^\top & \be_d^\top
    \end{bmatrix}
    \begin{bmatrix}
      \bLambda & \bO \\
      \bO & \bO
    \end{bmatrix}
    \begin{bmatrix}
      \be_1 & \bzero & \be_2 & \bzero & \dots & \be_d & \bzero \\
      \bzero & \be_1 & \bzero & \be_2 & \dots & \bzero & \be_d
    \end{bmatrix}
    \\
    =&
    \begin{bmatrix}
      \lambda_1 & & & & & & \\
      & 0 & & & & & \\
      & & \lambda_2 & & & & \\
      & & & 0 & & & \\
      & & & & \ddots & & \\
      & & & & & \lambda_d & \\
      & & & & & & 0
    \end{bmatrix}
    =
    \bLambda \otimes
    \begin{bmatrix}
      1 & 0 \\
      0 & 0
    \end{bmatrix},
  \end{aligned}
  \end{equation}
  Here $\otimes$ is the Kronecker product. Then the variance term is simplified to
  \begin{align*}
    &\sum_{\tau=0}^{T-1}
      \EE\left[
        \eta_\tau^2
        \begin{bmatrix}
          \bn_\tau \\
          \bzero
        \end{bmatrix}^\top
        \bM_{\tau+1}^\top \bM_{\tau+2}^\top \dots     \bM_{T-1}^\top
        \begin{bmatrix}
          \bH & \bO \\
          \bO & \bO
        \end{bmatrix}
        \bM_{T-1} \bM_{T-2} \dots \bM_{\tau+1}
        \begin{bmatrix}
          \bn_\tau \\
          \bzero
        \end{bmatrix}
      \right]
    \\
    \le&
    \sigma^2 \sum_{\tau=0}^{T-1} \eta_\tau^2
      \tr\left(
        \bTp_{\tau+1}^\top \bTp_{\tau+2}^\top \dots \bTp_{T-1}^\top
        \left(
          \bLambda \otimes
          \begin{bmatrix}
            1 & 0 \\
            0 & 0
          \end{bmatrix}
        \right)
        \bTp_{T-1} \bTp_{T-2} \dots \bTp_{\tau+1}
        \left(
          \bLambda \otimes
          \begin{bmatrix}
            1 & 0 \\
            0 & 0
          \end{bmatrix}
        \right)
      \right)
    \\
    \overset{\eqref{eq:bTp_def}}{=}&
    \sigma^2 \sum_{\tau=0}^{T-1} \eta_\tau^2
      \sum_{j=1}^d \tr\left(
        \bT_{\tau+1,j}^\top \bT_{\tau+2,j}^\top \dots \bT_{T-1,j}^\top
        \begin{bmatrix}
            \lambda_j & 0 \\
            0 & 0
        \end{bmatrix}
        \bT_{T-1,j} \bT_{T-2,j} \dots \bT_{\tau+1,j}
        \begin{bmatrix}
            \lambda_j & 0 \\
            0 & 0
        \end{bmatrix}
      \right)  
    \\
    &\mcomment{All are block diagonal matrices}
    \\
    =&
    \sigma^2 \sum_{\tau=0}^{T-1} \eta_\tau^2
      \sum_{j=1}^d \lambda_j^2 \tr\left(
        \bT_{\tau+1,j}^\top \bT_{\tau+2,j}^\top \dots \bT_{T-1,j}^\top
        \begin{bmatrix}
          1 \\ 0
        \end{bmatrix}
        \begin{bmatrix}
          1 \\ 0
        \end{bmatrix}^\top
        \bT_{T-1,j} \bT_{T-2,j} \dots \bT_{\tau+1,j}
        \begin{bmatrix}
          1 \\ 0
        \end{bmatrix}
        \begin{bmatrix}
          1 \\ 0
        \end{bmatrix}^\top
      \right)
    \\
    =&
    \sigma^2 \sum_{\tau=0}^{T-1} \eta_\tau^2
      \sum_{j=1}^d \lambda_j^2 \tr\left(
        \begin{bmatrix}
          1 \\ 0
        \end{bmatrix}^\top
        \bT_{\tau+1,j}^\top \bT_{\tau+2,j}^\top \dots \bT_{T-1,j}^\top
        \begin{bmatrix}
          1 \\ 0
        \end{bmatrix}
        \begin{bmatrix}
          1 \\ 0
        \end{bmatrix}^\top
        \bT_{T-1,j} \bT_{T-2,j} \dots \bT_{\tau+1,j}
        \begin{bmatrix}
          1 \\ 0
        \end{bmatrix}
      \right)
    \\
    &\mcomment{Cyclic property of trace}
    \\
    =&
    \sigma^2 \sum_{\tau=0}^{T-1} \eta_\tau^2
      \sum_{j=1}^d \lambda_j^2 \left(
        \begin{bmatrix}
          1 \\ 0
        \end{bmatrix}^\top
        \bT_{\tau+1,j}^\top \bT_{\tau+2,j}^\top \dots \bT_{T-1,j}^\top
        \begin{bmatrix}
          1 \\ 0
        \end{bmatrix}
      \right)
      \left(
        \begin{bmatrix}
          1 \\ 0
        \end{bmatrix}^\top
        \bT_{T-1,j} \bT_{T-2,j} \dots \bT_{\tau+1,j}
        \begin{bmatrix}
          1 \\ 0
        \end{bmatrix}
      \right)
    \\
    &\mcomment{Notice that the term inside the trace is a scalar}
    \\
    =&
    \sigma^2 \sum_{\tau=0}^{T-1} \eta_\tau^2
      \sum_{j=1}^d \lambda_j^2
      \left(
        \begin{bmatrix}
          1 \\ 0
        \end{bmatrix}^\top
        \bT_{T-1,j} \bT_{T-2,j} \dots \bT_{\tau+1,j}
        \begin{bmatrix}
          1 \\ 0
        \end{bmatrix}
      \right)^2
    \\
    &\mcomment{Transpose of a scalar is itself}
    \\
    \le&
    \sigma^2 \sum_{\tau=0}^{T-1} \eta_\tau^2
      \sum_{j=1}^d \lambda_j^2
      \norm{
        \bT_{T-1,j} \bT_{T-2,j} \dots \bT_{\tau+1,j}
      }^2
    \\
    =&
    \sigma^2 \sum_{j=1}^d \lambda_j^2
      \sum_{\tau=0}^{T-1} \eta_\tau^2
        \norm{\bT_{T-1,j} \bT_{T-2,j} \dots \bT_{\tau+1,j}}^2.
  \end{align*}
  Here the last inequality is entailed by the fact that for $\forall \bC \in \RR^{2\times2}$,
  \begin{align*}
    \begin{bmatrix}
      1 \\ 0
    \end{bmatrix}^\top
    \bC
    \begin{bmatrix}
      1 \\ 0
    \end{bmatrix}
    =&
    \left(
      \bC
      \begin{bmatrix}
        1 \\ 0
      \end{bmatrix}
    \right)_1
    \le
    \sqrt{
      \left(
        \bC
        \begin{bmatrix}
          1 \\ 0
        \end{bmatrix}
      \right)_1^2
      + \left(
        \bC
        \begin{bmatrix}
          1 \\ 0
        \end{bmatrix}
      \right)_2^2
    }
    =
    \norm{
      \bC
      \begin{bmatrix}
        1 \\ 0
      \end{bmatrix}
    }
    \le
    \norm{\bC},
  \end{align*}
  with $(\bx)_1, (\bx)_2$ standing for the first and second element of vector $\bx$.
\end{proof}

\begin{lemma}
\label{lem:bound_bias_with_T_norm}
\textbf{(Bound Bias with $\norm{\bT_{t,j}^k}$)}
\begin{equation}
\label{eq:bound_bias_with_T_norm}
\begin{aligned}
  &\EE\left[
    \begin{bmatrix}
      \bw_0 - \bw_* \\
      \bw_{-1} - \bw_*
    \end{bmatrix}^\top
    \bM_0^\top \bM_1^\top \dots \bM_{T-1}^\top
    \begin{bmatrix}
      \bH & \bO \\
      \bO & \bO
    \end{bmatrix}
    \bM_{T-1} \bM_{T-2} \dots \bM_0
    \begin{bmatrix}
      \bw_0 - \bw_* \\
      \bw_{-1} - \bw_*
    \end{bmatrix}
  \right]
  \\
  \le&
  \sum_{j=1}^d \lambda_j
    \norm{
      \bT_{T-1,j} \bT_{T-2,j} \dots \bT_{0,j}
    }^2
    \EE\norm{
      \left(
        \bPi^\top \bV^\top
        \begin{bmatrix}
          \bw_0 - \bw_* \\
          \bw_{-1} - \bw_*
        \end{bmatrix}
      \right)_{2j-1:2j}
    }^2,
\end{aligned}
\end{equation}
where $\bM_t$ is defined in Eqn.~\eqref{eq:def_M_t}, $\bT_{t,j} \in \RR^{2\times2}$ is defined in~\eqref{eq:T_tj_def} and $\bPi, \bV$ are orthogonal matrices defined in~\eqref{eq:Pi_V_def}. Here notation $\bz_{j_1:j_2}$ means
\begin{equation}
\label{eq:notation_subvector}
\begin{aligned}
  \mbox{For } \forall
  \bz =
  \begin{bmatrix}
    z_1 \\ z_2 \\ \vdots \\ z_{d'}
  \end{bmatrix}
  \in \RR^{d'},
  1 \le j \le j' \le d',
  \quad \quad\bz_{j:j'}
  \triangleq
  \begin{bmatrix}
    z_{j} \\ z_{j+1} \\ \vdots \\ z_{j'}
  \end{bmatrix}  
\end{aligned}
\end{equation}
\end{lemma}
\begin{proof}
  The proof is similar to a simplified version of the variance case, so we will reuse some of its notations to shorten the proof.
  \begin{align*}
    &\EE\left[
      \begin{bmatrix}
        \bw_0 - \bw_* \\
        \bw_{-1} - \bw_*
      \end{bmatrix}^\top
      \bM_0^\top \bM_1^\top \dots \bM_{T-1}^\top
      \begin{bmatrix}
        \bH & \bO \\
        \bO & \bO
      \end{bmatrix}
      \bM_{T-1} \bM_{T-2} \dots \bM_0
      \begin{bmatrix}
        \bw_0 - \bw_* \\
        \bw_{-1} - \bw_*
      \end{bmatrix}
    \right]
    \\
    \overset{\eqref{eq:bwp_bnp_bHp_def}}{=}&
    \EE\left[
      \bwp_0^\top
      \bM_1^\top \bM_2^\top \dots \bM_{T-1}^\top
      \bHp
      \bM_{T-1} \bM_{T-2} \dots \bM_1
      \bwp_0
    \right]
    \\
    \overset{\eqref{eq:M_t_decomp}\eqref{eq:bTp_def}}{=}&
    \EE\left[
      \bwp_0^\top
      \bV \bPi
      \bTp_0^\top \bTp_1^\top \dots \bTp_{T-1}^\top
      \bPi^\top \bV^\top
      \bHp
      \bV \bPi
      \bTp_{T-1} \bTp_{T-2} \dots \bTp_0
      \bPi^\top \bV^\top
      \bwp_0
    \right]
    \\
    =&
    \EE\left[
      \left(
        \bwp_0^\top
        \bV \bPi
      \right)
      \bTp_0^\top \bTp_1^\top \dots \bTp_{T-1}^\top
      \left(
        \bPi^\top \bV^\top
        \bHp
        \bV \bPi
      \right)
      \bTp_{T-1} \bTp_{T-2} \dots \bTp_0
      \left(
        \bPi^\top \bV^\top \bwp_0
      \right)
    \right]
    \\
    \overset{\eqref{eq:Pi'_V'_H_V_Pi}}{=}&
    \EE\left[
      \left(
        \bwp_0^\top
        \bV \bPi
      \right)
      \bTp_0^\top \bTp_1^\top \dots \bTp_{T-1}^\top
      \left(
        \bLambda \otimes
        \begin{bmatrix}
          1 & 0 \\
          0 & 0
        \end{bmatrix}
      \right)
      \bTp_{T-1} \bTp_{T-2} \dots \bTp_0
      \left(
        \bPi^\top \bV^\top \bwp_0
      \right)
    \right]
    \\
    \overset{\eqref{eq:bTp_def}\eqref{eq:notation_subvector}}{=}&
    \EE\left[
      \begin{bmatrix}
        \left(\bPi^\top \bV^\top \bwp_0\right)_{1:2}
        \\
        \left(\bPi^\top \bV^\top \bwp_0\right)_{3:4}
        \\
        \vdots
        \\
        \left(\bPi^\top \bV^\top \bwp_0\right)_{2d-1:2d}
      \end{bmatrix}^\top
      \begin{bmatrix}
        \bS_1 & & & \\
        & \bS_2 & & \\
        & & \ddots & \\
        & & & \bS_d
      \end{bmatrix}
      \begin{bmatrix}
        \left(\bPi^\top \bV^\top \bwp_0\right)_{1:2}
        \\
        \left(\bPi^\top \bV^\top \bwp_0\right)_{3:4}
        \\
        \vdots
        \\
        \left(\bPi^\top \bV^\top \bwp_0\right)_{2d-1:2d}
      \end{bmatrix}
    \right],
    \\
    &\mbox{ where }
    \bS_j =
      \bT_{0,j}^\top \bT_{1,j}^\top \dots \bT_{T-1,j}^\top
      \begin{bmatrix}
        \lambda_j & 0 \\
        0 & 0
      \end{bmatrix}
      \bT_{T-1,j} \bT_{T-2,j} \dots \bT_{0,j}
    \\
    =&
    \EE\left[
      \sum_{j=1}^d
        \left(
          \bPi^\top \bV^\top \bwp_0
        \right)_{2j-1:2j}^\top
        \bS_j
        \left(
          \bPi^\top \bV^\top \bwp_0
        \right)_{2j-1:2j}
    \right],
    \\
    &\mbox{ where }
    \bS_j =
      \bT_{0,j}^\top \bT_{1,j}^\top \dots \bT_{T-1,j}^\top
      \begin{bmatrix}
        \lambda_j & 0 \\
        0 & 0
      \end{bmatrix}
      \bT_{T-1,j} \bT_{T-2,j} \dots \bT_{0,j}
    \\
    =&
    \EE\left[
      \sum_{j=1}^d
        \left(
          \bPi^\top \bV^\top \bwp_0
        \right)_{2j-1:2j}^\top
        \left(
          \lambda_j \bs_j \bs_j^\top
        \right)
        \left(
          \bPi^\top \bV^\top \bwp_0
        \right)_{2j-1:2j}
    \right],
    \\
    &\mbox{ where }
    \bs_j =
      \bT_{0,j}^\top \bT_{1,j}^\top \dots \bT_{T-1,j}^\top
      \begin{bmatrix}
        1 \\ 0
      \end{bmatrix}
    \\
    =&
    \EE\left[
      \sum_{j=1}^d \lambda_j
        \left(
          \left(
            \bPi^\top \bV^\top \bwp_0
          \right)_{2j-1:2j}^\top
          \bs_j
        \right)
        \left(
          \bs_j^\top
          \left(
            \bPi^\top \bV^\top \bwp_0
          \right)_{2j-1:2j}
        \right)
    \right],
    \\
    &\mbox{ where }
    \bs_j =
      \bT_{0,j}^\top \bT_{1,j}^\top \dots \bT_{T-1,j}^\top
      \begin{bmatrix}
        1 \\ 0
      \end{bmatrix}
    \\
    =&
    \EE\left[
      \sum_{j=1}^d \lambda_j
        \left(
          \bs_j^\top
          \left(
            \bPi^\top \bV^\top \bwp_0
          \right)_{2j-1:2j}
        \right)^2
    \right],
    \mbox{ where }
    \bs_j =
      \bT_{0,j}^\top \bT_{1,j}^\top \dots \bT_{T-1,j}^\top
      \begin{bmatrix}
        1 \\ 0
      \end{bmatrix}
    \\
    =&
    \EE\left[
      \sum_{j=1}^d \lambda_j
        \left(
          \begin{bmatrix}
            1 \\ 0
          \end{bmatrix}^\top
          \bT_{T-1,j} \bT_{T-2,j} \dots \bT_{0,j}
          \left(
            \bPi^\top \bV^\top \bwp_0
          \right)_{2j-1:2j}
        \right)^2
    \right]
    \\
    \le&
    \EE\left[
      \sum_{j=1}^d \lambda_j
        \norm{
          \bT_{T-1,j} \bT_{T-2,j} \dots \bT_{0,j}
        }^2
        \norm{
          \left(
            \bPi^\top \bV^\top \bwp_0
          \right)_{2j-1:2j}
        }^2
    \right]
    \\
    =&
    \sum_{j=1}^d \lambda_j
      \norm{
        \bT_{T-1,j} \bT_{T-2,j} \dots \bT_{0,j}
      }^2
      \EE\norm{
        \left(
          \bPi^\top \bV^\top \bwp_0
        \right)_{2j-1:2j}
      }^2
    \\
    &\mcomment{Linearity of expectation}
    \\
    \overset{\eqref{eq:bwp_bnp_bHp_def}}{=}&
    \sum_{j=1}^d \lambda_j
      \norm{
        \bT_{T-1,j} \bT_{T-2,j} \dots \bT_{0,j}
      }^2
      \EE\norm{
       \left(
         \bPi^\top \bV^\top
           \begin{bmatrix}
             \bw_0 - \bw_* \\
             \bw_{-1} - \bw_*
           \end{bmatrix}
        \right)_{2j-1:2j}
      }^2.
  \end{align*}
  Here the inequality is entailed by the fact that for $\forall \bC \in \RR^{2\times2}, \bz \in \RR^2$,
  \begin{align*}
    \begin{bmatrix}
      1 \\ 0
    \end{bmatrix}^\top
    \bC \bz
    =&
    \left(
      \bC \bz
    \right)_1
    \le
    \sqrt{
      \left(
        \bC \bz
      \right)_1^2
      + \left(
        \bC \bz
      \right)_2^2
    }
    =
    \norm{
      \bC \bz
    }
    \le
    \norm{\bC} \norm{\bz}
  \end{align*}
\end{proof}

\input{new_proof}

\end{document}

%% file: math_commands.tex
\usepackage{amsmath,amsfonts,bm}

\def\eqref#1{equation~\ref{#1}}
\def\ceil#1{\lceil #1 \rceil}

\def\1{\bm{1}}

\DeclareMathAlphabet{\mathsfit}{\encodingdefault}{\sfdefault}{m}{sl}
\SetMathAlphabet{\mathsfit}{bold}{\encodingdefault}{\sfdefault}{bx}{n}

%% file: new_proof.tex
\newcommand{\Tb}{\mathbf{T}}
\newcommand{\Pb}{\mathbf{P}}
\newcommand{\Jb}{\mathbf{J}}
\newcommand{\wb}{\mathbf{w}}
\newcommand{\Deltab}{\mathbf{\Delta}}
\newcommand{\Norm}[1]{\left\| #1 \right\|}
\newcommand{\Abs}[1]{\left| #1 \right|}
\newcommand{\skuo}[1]{\left({#1}\right)}

\setlength{\parindent}{0em}

% \begin{abstract}
% Your abstract.
% \end{abstract}

% \section{Problem Setting}
% We denote the matrices
% \begin{align*}
% \Tb_{t,j}=\left[\begin{matrix}\alpha&-\beta\\1&0\end{matrix}\right],\qquad &
% \Delta=\left[\begin{matrix}\delta&0\\0&0\end{matrix}\right],
% \end{align*}
% where $\alpha=1+\beta-\eta_t\lambda_j$ and $\delta=\cO{1/(T^2)}$. We pursue a bound for
% \begin{align*}
%     \Norm{\left(\Tb_{t,j}+\Delta \right)^k-\Tb_{t,j}^k}.
% \end{align*}

\subsection{Bounding $\Norm{\Tb_{t+k,j}...\Tb_{t+1,j}}$ with $\rho(\Tb_{t+k,j})$}\label{appendix:matrix_product_to_spectral_radius}

This section upper bounds the matrix product $\Norm{\Tb_{t+k,j}...\Tb_{t+1,j}}$ with the spectral radius $\rho(\Tb_{t+1,j})$. Similar results for bounding $\Norm{\Tb_{t,j}^k}$ have been shown in~\citep{wang2021modular}(Theorem 5), but our result is more general, so we still put our proofs here.
In the following proof, we use $\Norm{\cdot}_F$ to denote the Frobenius norm of matrices.
% \begin{lemma}[Auxiliary fact for bounding stage power, maybe should be moved to appendix \ref{appendix:useful_lemmas}]\label{lem:aux_e}
%     For $x\in[1,+\infty)$,
%     it holds that
%     \begin{align}\label{eq:lem_aux_e}
%         f(x) = \left( 1-\frac{1}{x} \right)^x \leq \frac{1}{e}.
%     \end{align}
% \end{lemma}
% \begin{proof}
%     The lemma is equivalent to that
%     \begin{align*}
%         x\ln\left(1-\frac{1}{x}\right) \leq -1
%     \end{align*}
%     Denote $g(t) = \ln(1-t)+t,\quad t\in (0,1)$, then $g(t)$ is monotonically decreasing as
%     \begin{align*}
%         g'(t) = -\frac{1}{1-t} + 1 < 0.
%     \end{align*}
%     Therefore, $g(t)\leq g(0)=0$ when $t\in(0,1)$. Thus substituting $t=1/x$, it holds that
%     \begin{align*}
%         \ln\left( 1-\frac{1}{x}\right) + \frac{1}{x} \leq 0.
%     \end{align*}
%     After rearrangement, we can obtain the result.
% \end{proof}

\begin{lemma}[Bounding $\Norm{\Tb_{t,j}^k}_F$ with $\rho(\Tb_{t,j})^k$]\label{lem:1}
    Given momentum matrices $\Tb_{t,j}$ that are defined in Eqn.~\eqref{eq:T_tj_def} and $\beta\geq1/4$,
   for all positive integer $k \ge 1$, it holds that 
   \begin{align}\label{eq:lem_1}
       \Norm{\Tb_{t,j}^k}_F \le \min\left(8k,\frac{8}{\sqrt{\Abs{(1+\beta-\eta_t\lambda_j)^2-4\beta}}}\right)\rho(\Tb_{t,j})^k.
   \end{align}
\end{lemma}
\begin{proof}
According to the definition of $\Tb_{t,j}$ in Eqn.~\eqref{eq:T_tj_def},
\begin{align*}
    \Tb_{t,j}=\begin{bmatrix}
        1+\beta-\eta_t\lambda_j & -\beta \\ 1 & 0
    \end{bmatrix}.
\end{align*}
We can directly analyze the product by Jordan decomposition that there exists $\Pb\in\mathbb{C}^{2\times 2}$ such that
\begin{align*}
    \Tb_{t,j}=\Pb\Jb\Pb^{-1}
\end{align*}
where $\Jb$ can have the following two cases 
\begin{align*}
    \Jb=\left[ \begin{matrix}
        \gamma_1 & 0 \\ 0 & \gamma_2
    \end{matrix}\right] \text{ or }
    \Jb=\left[ \begin{matrix}
        \gamma_1 & 1 \\ 0 & \gamma_2
    \end{matrix}\right]
\end{align*}
depending on whether $\Tb_{t,j}$ is diagnolizable. Here $\gamma_1, \gamma_2$ are the eigenvalues of $\Tb_{t,j}$ and we assume without generality that $\gamma_1\geq \gamma_2$ if $\gamma_1,\gamma_2\in\mathbb{R}$. And when $\gamma_1,\gamma_2\notin\mathbb{R}$, it holds that $\gamma_1$ and $\gamma_2$ are conjugate thus $\Abs{\gamma_1}=\Abs{\gamma_2}$. Therefore $|\gamma_1|$ is the spectral radius of $\Tb_{t,j}$, i.e. $\rho(\Tb_{t,j})=|\gamma_1|$. Then we discuss case by case.

\textbf{i) If $\gamma_1\neq \gamma_2$:} 
In this case one can verify that
\begin{align*}
    \Pb
    =
    \begin{bmatrix}
      \gamma_1 & \gamma_2 \\
      1 & 1
    \end{bmatrix},
    \quad
    \Jb
    =
    \begin{bmatrix}
      \gamma_1 & 0 \\
      0 & \gamma_2
    \end{bmatrix},
    \quad
    \Pb^{-1}
    =
    \frac{1}{\gamma_1 - \gamma_2}
    \begin{bmatrix}
      1 & -\gamma_2 \\
      -1 & \gamma_1
      \end{bmatrix}
\end{align*}
where $\gamma_1,\gamma_2$ are eigenvalues of $\Tb_{t,j}$.
And the characteristic polynomial of $\Tb_{t,j}$
\begin{align*}
    \det\left(\Tb_{t,j} - \gamma \mathbf{I}\right)
    {=}
    \begin{vmatrix}
      1 + \beta - \eta_t \lambda_j - \gamma & -\beta \\
      1 & -\gamma
    \end{vmatrix}
    =
    \gamma^2
    - \left(1 + \beta - \eta_t \lambda_j\right) \gamma
    + \beta
    =
    0
\end{align*}
entails $\gamma_1 + \gamma_2 = 1 + \beta - \eta_t \lambda_j, \gamma_1 \gamma_2 = \beta$. Thus in this case, it holds that
\begin{align*}
    \Tb_{t,j}^k &= \Pb\Jb^k\Pb^{-1} = \begin{bmatrix}
      \gamma_1 & \gamma_2 \\
      1 & 1
    \end{bmatrix}
    \cdot
    \begin{bmatrix}
      \gamma_1^k & 0 \\
      0 & \gamma_2^k
    \end{bmatrix}
    \cdot
    \frac{1}{\gamma_1 - \gamma_2}
    \begin{bmatrix}
      1 & -\gamma_2 \\
      -1 & \gamma_1
      \end{bmatrix} \\
      &= \begin{bmatrix}
          \frac{\gamma_1^{k+1}-\gamma_2^{k+1}}{\gamma_1-\gamma_2} & -\beta\frac{\gamma_1^{k}-\gamma_2^{k}}{\gamma_1-\gamma_2} \\ \frac{\gamma_1^{k}-\gamma_2^{k}}{\gamma_1-\gamma_2} &
          -\beta\frac{\gamma_1^{k-1}-\gamma_2^{k-1}}{\gamma_1-\gamma_2}
      \end{bmatrix}. 
\end{align*}
It holds that
\begin{equation*}
\begin{aligned}
    \Norm{\Tb_{t,j}}_F =& \frac{1}{\Abs{\gamma_1-\gamma_2}} \Norm{\begin{bmatrix}
        \gamma_1^{k+1}-\gamma_2^{k+1} & 
        -\beta\left(\gamma_1^{k}-\gamma_2^{k}\right) \\
        \gamma_1^{k}-\gamma_2^{k} &
        -\beta\left(\gamma_1^{k-1}-\gamma_2^{k-1}\right)
    \end{bmatrix}}_F \\
    \le&
    \frac{1}{\Abs{\gamma_1-\gamma_2}}
    \Norm{\begin{bmatrix}
        2\rho(\Tb_{t,j})^{k+1} & 
        -2\beta\rho(\Tb_{t,j})^{k} \\
        2\rho(\Tb_{t,j})^{k} &
        -2\beta\rho(\Tb_{t,j})^{k-1}
    \end{bmatrix}}_F \\
    \le&
    \frac{1}{\Abs{\gamma_1-\gamma_2}}
    \Norm{\begin{bmatrix}
        2\rho(\Tb_{t,j})^{k} & 
        -2\beta\rho(\Tb_{t,j})^{k} \\
        2\rho(\Tb_{t,j})^{k} &
        -4\beta\rho(\Tb_{t,j})^{k}
    \end{bmatrix}}_F 
    \le
    \frac{2\rho(\Tb_{t,j})^{k}}{\Abs{\gamma_1-\gamma_2}}
    \Norm{\begin{bmatrix}
        1 & 
        -\beta \\
        1 &
        -2\beta
    \end{bmatrix}}_F \\
    \le&
    \frac{8\rho(\Tb_{t,j})^{k}}{\Abs{\gamma_1-\gamma_2}}
    =
    \frac{8}{\sqrt{\Abs{(1+\beta-\eta_t\lambda_j)^2-4\beta}}}\rho(\Tb_{t,j})^{k},
\end{aligned}   
\end{equation*}
where the first inequality holds as $\Abs{\gamma_1}=\Abs{\gamma_2}=\rho(\Tb_{t,j})$ and the third inequality holds as $\rho(\Tb_{t,j})\geq\gamma_1\gamma_2=\beta\geq1/4$ according to Lemma~\ref{lem:roots_of_quad_equations} and thus $\rho(\Tb_{t,j})\in(1/2,1)$. For more details about the properties of $\rho(\Tb_{t,j})$ one can refer to Appendix~\ref{appendix:spetral_raidus_to_parameters}. We can also analyze the norm from another point: it holds that for all $k$,
\begin{equation*}
\begin{aligned}
    \Abs{\frac{\gamma_1^{k}-\gamma_2^{k}}{\gamma_1-\gamma_2}}=\Abs{\sum_{j=0}^{k-1}\gamma_1^j\gamma_2^{k-1-j}} \le
    \sum_{j=0}^{k-1} \Abs{\gamma_1^{j}\gamma_2^{k-1-j} }=\sum_{j=0}^{k-1}\rho(\Tb_{t,j})^{k-1}=k\rho(\Tb_{t,j})^{k-1}.
\end{aligned}
\end{equation*}
Substituting we have
\begin{equation*}
\begin{aligned}
    \Norm{\Tb_{t,j}}_F =& 
    \Norm{\begin{bmatrix}
          \frac{\gamma_1^{k+1}-\gamma_2^{k+1}}{\gamma_1-\gamma_2} & -\beta\frac{\gamma_1^{k}-\gamma_2^{k}}{\gamma_1-\gamma_2} \\ \frac{\gamma_1^{k}-\gamma_2^{k}}{\gamma_1-\gamma_2} &
          -\beta\frac{\gamma_1^{k-1}-\gamma_2^{k-1}}{\gamma_1-\gamma_2}
      \end{bmatrix}}_F \\
      \le&
      \Norm{\begin{bmatrix}
          (k+1)\rho(\Tb_{t,j})^{k} & -\beta k\rho(\Tb_{t,j})^{k-1} \\ k\rho(\Tb_{t,j})^{k-1} &
          -\beta(k-1)\rho(\Tb_{t,j})^{k-2}
      \end{bmatrix}}_F \\
      \le&
      \Norm{\begin{bmatrix}
          2k\rho(\Tb_{t,j})^{k} & -2 k\rho(\Tb_{t,j})^{k} \\ 2k\rho(\Tb_{t,j})^{k} &
          -4k\rho(\Tb_{t,j})^{k}
      \end{bmatrix}}_F 
      \le 8k\rho(\Tb_{t,j})^k.
\end{aligned}
\end{equation*}
Thus we prove the lemma when $\gamma_1\neq \gamma_2$.

\textbf{ii) If $\gamma_1=\gamma_2=\gamma$:} In this case, $\Tb_{t,j}$ can not be diagonalized, which means that
\begin{align*}
    \Tb_{t,j} = \begin{bmatrix}
        \gamma & 1 \\ 0 & \gamma
    \end{bmatrix}
\end{align*}
one can verify that
\begin{align*}
    \Pb
    =
    \begin{bmatrix}
      \gamma & 1 \\
      1 & 0
    \end{bmatrix},
    \quad
    \Jb^k
    =
    \begin{bmatrix}
      \gamma^k & k\gamma^{k-1} \\
      0 & \gamma^k
    \end{bmatrix},
    \quad
    \Pb^{-1}
    =
    \begin{bmatrix}
      0 & 1 \\
      1 & -\gamma
    \end{bmatrix}
\end{align*}
where $\gamma_1=\gamma_2=\gamma\in\mathbb{R}$. In this case, it holds that
\begin{align*}
    \Tb_{t,j}^{k}=\Pb\Jb^{k}\Pb^{-1} &=
    \begin{bmatrix}
      \gamma & 1 \\
      1 & 0
    \end{bmatrix}
    \cdot
    \begin{bmatrix}
      \gamma^k & k\gamma^{k-1} \\
      0 & \gamma^k
    \end{bmatrix}
    \cdot
    \begin{bmatrix}
      0 & 1 \\
      1 & -\gamma
    \end{bmatrix} \\
    &=
    \begin{bmatrix}
        (k+1)\gamma^k & -k\gamma^{k+1} \\
        k\gamma^{k-1} & -(k-1)\gamma^{k}
    \end{bmatrix}
\end{align*}
Then it holds that
\begin{align*}
    \Norm{\Tb_{t,j}^k}_F =& \Norm{\begin{bmatrix}
        (k+1)\gamma^k & -k\gamma^{k+1} \\
        k\gamma^{k-1} & -(k-1)\gamma^{k}
    \end{bmatrix}}_F
    \le
    \Norm{\begin{bmatrix}
        2k\gamma^k & -k\gamma^{k} \\
        2k\gamma^{k} & -k\gamma^{k}
    \end{bmatrix}}_F
    \le
    8k\gamma^k=8k\rho(\Tb_{t,j})^k.
\end{align*}
Therefore, combining the two cases, we obtain the conclusion.    
\end{proof}

\begin{lemma}[Bounding $\Norm{(\Tb_{t,j}+\Deltab_1)(\Tb_{t,j}+\Deltab_1)...(\Tb_{t,j}+\Deltab_k)}_F$ with matrix power]\label{lem:2}
    Given matrices $\Tb_{t,j}$ defined in Eqn.~\eqref{eq:T_tj_def} and  $\Deltab_i$, $\Deltab$ defined as 
    \begin{align*}
        \Deltab_i = \begin{bmatrix}
            \delta_i &0 \\ 0 & 0
        \end{bmatrix}, \qquad
        \Deltab = \begin{bmatrix}
            \delta &0 \\ 0 & 0
        \end{bmatrix},
    \end{align*}
    where $\delta_i\geq0$ and $\delta = \max_{1\leq i \leq k}\delta_i$, if 
    $
        (1+\beta-\eta_t\lambda_j)^2-4\beta\geq0,
    $
    it holds that
    \begin{align}\label{eq:lem_2}
        \Norm{(\Tb_{t,j}+\Deltab_1)(\Tb_{t,j}+\Deltab_2)...(\Tb_{t,j}+\Deltab_k)}_F\le
        \Norm{\left(\Tb_{t,j}+\Deltab \right)^k}_F.
    \end{align}
\end{lemma}
\begin{proof}
As we assume $(1+\beta-\eta_t\lambda_j)^2-4\beta\geq0$, the eigenvalues of $\Tb_{t,j}$ $\gamma_1, \gamma_2\in \mathbb{R}$. Following the same method as the proof of Lemma \ref{lem:1} to apply Jordan decomposition to $\Tb_{t,j}$, the power of momentum matrix $\Tb_{t,j}$ can be written as
\begin{align*}
    \Tb_{t,j}^k
    &=\left[\begin{matrix}\frac{\gamma_1^{k+1}-\gamma_2^{k+1}}{\gamma_1-\gamma_2}&\frac{-\beta(\gamma_1^{k}-\gamma_2^{k})}{\gamma_1-\gamma_2}\\\frac{\gamma_1^k-\gamma_2^k}{\gamma_1-\gamma_2}&\frac{-\beta(\gamma_2^{k-1}-\gamma_1^{k-1})}{\gamma_1-\gamma_2}\end{matrix}\right], \quad \text{if } \gamma_1\neq \gamma_2\\
    % &=\left[\begin{matrix} \sum_{j=0}^k\gamma_1^{j}\gamma_2^{k-1-j} & -\beta\sum_{j=0}^{k-1}\gamma_1^{j}\gamma_2^{k-2-j} \\ \sum_{j=0}^{k-1}\gamma_1^{j}\gamma_2^{k-2-j} & -\beta\sum_{j=0}^{k-2}\gamma_1^{j}\gamma_2^{k-3-j}\end{matrix}\right], \quad \text{if } \gamma_1\neq \gamma_2 \\
    \Tb_{t,j}^k
    &= \begin{bmatrix}
        (k+1)\gamma^k & -k\gamma^{k+1} \\
        k\gamma^{k-1} & -(k-1)\gamma^{k}
    \end{bmatrix}, \quad \text{if } \gamma_1=\gamma_2=\gamma
\end{align*}
We can observe that in this case, the first column of $\Tb_{t,j}^k$ is nonnegative and the second column is nonpositive as $\gamma_1,\gamma_2,\gamma\in\mathbb{R}$. 
For simplicity, in the following proof we use $\prod$ to denote a product from $i=1$ to $i=k$ orderly from left to right, namely,  $\prod_{i=1}^k\Tb_{t+i,k}=\Tb_{t+k,j}\Tb_{t+k-1,j}...\Tb_{t+1,j}$. 
We first consider the combination product form of ${\prod_{i=1}^{k}(\Tb_{t,j}+\Deltab_i)}$ that
\begin{align*}
    &\prod_{i=1}^k(\Tb_{t,j}+\Deltab_i)=\sum_{l_1+...+l_t+k_1+...+k_{t+1}=k}\Tb_{t,j}^{k_1}\Deltab_{11}...\Deltab_{1l_1}\Tb_{t,j}^{k_2}\Deltab_{21}...\Deltab_{2l_2}\Tb_{t,j}^{k_3}...\Deltab_{t1}...\Deltab_{tl_t}\Tb_{t,j}^{k_{t+1}},
\end{align*}
which is similar to the binomial expansion but without the commutativity of $\Tb$ and $\Deltab_i$. Now we consider one arbitrary combination term $\Sb$ that
\begin{align*}
    \Sb=\Tb_{t,j}^{k_1}\Deltab_{11}...\Deltab_{1l_1}\Tb_{t,j}^{k_2}\Deltab_{21}...\Deltab_{2l_2}\Tb_{t,j}^{k_3}...\Deltab_{t1}...\Deltab_{tl_t}\Tb_{t,j}^{k_{t+1}}.
\end{align*}
We first prove by induction that $\Sb$ has the following properties: 
\begin{enumerate}
    \item the first column of $\Sb$ is nonnegative and the second column of $\Sb$ is nonpositive;
    \item the absolute value of each entry of $\Sb$ is monotonically increasing with respect to $\delta_{1},...,\delta_{k}$.
\end{enumerate}
We call $\Tb_{t,j}^{k_i}$ or $\Deltab_{ij}$ one multiple component of $\Sb$ in the following proof. And we denote $\Sb_p$ the product of the first $p$ multiple component of $\Sb$ in the following proof. The first multiple component of $\Sb$ can be $\Sb_1=\Tb_{t,j}^{k_1}$ or $\Sb_1=\Deltab_{11}$, which satisfies the two desired properties naturally. Then we assume that the product of the first $p$ multiple component $\Sb_p$ satisfies the two properties. We discuss $\Sb_{p+1}$ in cases that
\begin{enumerate}
    \item if the $p$-th multiple component is $\Tb_{t,j}^{i}$, where $i$ represents an arbitrary integer, then the $(p+1)$-th multiple component should be $\Deltab_{i'}$, where $i'$ also represents an arbitrary integer, or the $p$-th and $(p+1)$-th component can be merged. Then it holds that
    \begin{align*}
        \Sb_{p+1}=\Sb_p\Deltab_{i'}=\left[\begin{matrix}
            \Sb_{p,11} & \Sb_{p,12} \\ \Sb_{p,21} & \Sb_{p,22}
        \end{matrix}\right]
        \left[\begin{matrix}
            \delta_{i'} & 0 \\ 0 & 0
        \end{matrix}\right]
        =
        \left[\begin{matrix}
            \delta_{i'}\Sb_{p,11} & 0 \\ \delta_{i'}\Sb_{p,21} & 0
        \end{matrix}\right].
    \end{align*}
    Thus if the two properties hold for $p$, it also holds for $p+1$ in this case.
    \item if the $p$-th multiple component is $\Deltab_{i}$, where $i$ represents an arbitrary integer, and the $(p+1)$-th multiple component is $\Deltab_{i+1}$, then it holds that
    \begin{align*}
        \Sb_{p+1}=\Sb_p\Deltab_{i+1}=\left[\begin{matrix}
            \Sb_{p,11} & \Sb_{p,12} \\ \Sb_{p,21} & \Sb_{p,22}
        \end{matrix}\right]
        \left[\begin{matrix}
            \delta_{i+1} & 0 \\ 0 & 0
        \end{matrix}\right]
        =
        \left[\begin{matrix}
            \delta_{i+1}\Sb_{p,11} & 0 \\ \delta_{i+1}\Sb_{p,21} & 0
        \end{matrix}\right].
    \end{align*}
    Thus if the two properties hold for $p$, it also holds for $p+1$ in this case.
    \item if the $p$-th multiple component is $\Deltab_{i}$, where $i$ represents an arbitrary integer, and the $(p+1)$-th multiple component is $\Tb_{t,j}^{i'}$, then it holds that
    \begin{align*}
        \Sb_{p}=\Sb_{p-1}\Deltab_{i}=\left[\begin{matrix}
            \Sb_{p-1,11} & \Sb_{p-1,12} \\ \Sb_{p-1,21} & \Sb_{p-1,22}
        \end{matrix}\right]
        \left[\begin{matrix}
            \delta_{i} & 0 \\ 0 & 0
        \end{matrix}\right]
        =
        \left[\begin{matrix}
            \delta_{i}\Sb_{p-1,11} & 0 \\ \delta_{i}\Sb_{p-1,21} & 0
        \end{matrix}\right],
    \end{align*}
    which implies that $\Sb_{p,12}=\Sb_{p,22}=0$, thus we can substitute that
    \begin{align*}
        \Sb_{p+1}=\Sb_p\Tb_{t,j}^{i'}=\left[\begin{matrix}
            \Sb_{p,11} & 0 \\ \Sb_{p,21} & 0
        \end{matrix}\right]
        \left[\begin{matrix}
            t_{11} & t_{12} \\ t_{21} & t_{22}
        \end{matrix}\right]
        =
        \left[\begin{matrix}
            \Sb_{p,11}t_{11} & \Sb_{p,11}t_{12} \\ \Sb_{p,21}t_{11} & \Sb_{p,21}t_{12}
        \end{matrix}\right].
    \end{align*}
    As $t_{11},\Sb_{p,11},\Sb_{p,21}$ are nonnegative, $t_{12}$ is nonpositive, the two properties also hold for $\Sb_{p+1}$.
\end{enumerate}

Therefore, the two properties hold for $\Sb_{p+1}$ and thus for any combination term $\Sb$ by induction. And one can verify that the properties also hold for their summation $\prod_{i=1}^k(\Tb+\Deltab_i)$. Because of the definition of frobenius norm, when the two properties hold, $\Norm{\prod_{i=1}^k(\Tb+\Deltab_i)}_F$ is monotonically increasing with respect to $\delta_1,...\delta_k$ as well. Therefore, it holds that
\begin{align*}
    \Norm{\prod_{i=1}^k(\Tb_{t,j}+\Deltab_i)}_F \le \Norm{\prod_{i=1}^k(\Tb_{t,j}+\Deltab)}_F
    = \Norm{(\Tb_{t,j}+\Deltab)^k}_F,
\end{align*}
which concludes the proof.
\end{proof}

Then combining Lemma \ref{lem:1} and Lemma \ref{lem:2}, we can obtain a conclusion.
\begin{lemma}\label{lem:matrix_main}
Given $\beta\in [1/4,1)$, $\Tb_{t,j}$ defined as Eqn.~\eqref{eq:T_tj_def},
if $\Tb_{t,j}$ only has real eigenvalues, which is equivalent to that the discriminant of $\Tb_{t,j}$ satisfies that $(1+\beta-\eta_t\lambda_j)^2-4\beta\geq0$, it holds that
    \begin{align}\label{eq:lem_main}
        \Norm{\Tb_{t+1,j}\Tb_{t+2,j}...\Tb_{t+k,j}}\le& \min\left(8k, \frac{8}{\sqrt{\Abs{(1+\beta-\eta_{t+k}\lambda_j)^2-4\beta}}} \right)\rho(\Tb_{t+k,j})^{k}
        .
    \end{align}
\end{lemma}
\begin{proof}
    The difference of two momentum matrices $\Tb_{t,j}$ and $\Tb_{t',j}$ that $\eta_{t'}\leq \eta_t$ is
    \begin{align*}
        \Tb_{t',j} - \Tb_{t,j} = \begin{bmatrix}
            (\eta_{t'}-\eta_t)\lambda_j & 0 \\ 0 & 0
        \end{bmatrix},
    \end{align*}
    which has the same structure with $\Deltab_i$ in Lemma~\ref{lem:2}. Thus Lemma~\ref{lem:matrix_main} is a natural combination of Lemma \ref{lem:1} and Lemma \ref{lem:2}. One can verify that
    \begin{align*}
        \Norm{\Tb_{t+1,j}\Tb_{t+2,j}...\Tb_{t+k,j}} \le& \Norm{\Tb_{t+1,j}\Tb_{t+2,j}...\Tb_{t+k,j}}_F 
        \overset{\eqref{eq:lem_2}}{\le} \Norm{\Tb_{t+k,j}^k}_F \\
        \overset{\eqref{eq:lem_1}}{\le}&
        \min\left(8k, \frac{8}{\sqrt{\Abs{(1+\beta-\eta_{t+k}\lambda_j)^2-4\beta}}} \right)\rho(\Tb_{t+k,j})^{k},
    \end{align*}
    which concludes the proof.
\end{proof}

\subsection{Key Properties of $\rho(\Tb)$}\label{appendix:spetral_raidus_to_parameters}

This section offers some useful property of spectral radius $\rho(\bT_{t,j})$ in terms of different $\eta_t$, $\beta$ and $\lambda_j$.
\begin{lemma}[The exact form of $\rho$ and its relationship with $\{\beta,\eta,\lambda\}$]\label{lem:rho_monotonic}
    Given momentum matrix
    \begin{align*}
        \Tb=\left[\begin{matrix}
            1+\beta-\eta\lambda & -\beta \\ 1 & 0
        \end{matrix}\right],
    \end{align*}
    if $\Tb_{t,j}$ only has real eigenvalues, which is equivalent to that the discriminant $(1+\beta-\eta\lambda)^2-4\beta>0$
    the spectral radius of $\Tb$ is
    \begin{align*}
        \rho(\Tb)=\frac{1}{2}\left[1+\beta-\eta\lambda+\sqrt{(1+\beta-\eta\lambda)^2-4\beta}\right].
    \end{align*}
    Else the spectral radius of $\Tb$ is
    \begin{align*}
        \rho(\Tb)=\sqrt{\beta}.
    \end{align*}
    Thus under the assumption that $\eta\lambda\leq 1$, we have $\rho(\Tb)$ is monotonically decreasing with respect to $\eta\lambda$.
\end{lemma}
\begin{proof}
    To find the spectral radius, we first derive the eigenvalues of $\Tb$, which is equivalent to solving the equation
    \begin{align*}
        \text{det}\Abs{\Tb-\gamma\bI}=\text{det}\Abs{\begin{matrix}
            1 + \beta - \eta\lambda - \gamma & -\beta \\ 1 & -\gamma
        \end{matrix}} = 0.
    \end{align*}
    After rearrangement we have
    \begin{align*}
        \gamma^2-(1+\beta-\eta\lambda)\gamma +\beta = 0,
    \end{align*}
    which is a quadratic equation. The discriminant is that
    \begin{align*}
        \Delta = (1+\beta-\eta\lambda)^2-4\beta= \left( \left(1-\sqrt{\beta} \right)^2-\eta\lambda\right) \left( \left(1+\sqrt{\beta} \right)^2-\eta\lambda\right).
    \end{align*}
    Under the assumption that $\eta\lambda\le 1$, the positivity of $\Delta$ depends on the positivity of $\left(1-\sqrt{\beta} \right)^2-\eta\lambda$. If $\Delta\geq0$, then the eigenvalues $\gamma_1,\gamma_2\in\mathbb{R}$. If $\Delta<0$, then the eigenvalues $\gamma_1,\gamma_2\notin\mathbb{R}$. We then discuss these two cases.

    \textbf{Case 1: $\Delta\geq0$} In this case, we have
    \begin{align*}
        \gamma_1 &= \frac{1}{2}\left[1+\beta-\eta\lambda+\sqrt{(1+\beta-\eta\lambda)^2-4\beta}\right], \\ \gamma_2 &= \frac{1}{2}\left[1+\beta-\eta\lambda-\sqrt{(1+\beta-\eta\lambda)^2-4\beta}\right].
    \end{align*}
    Then we have $\gamma_1\geq \gamma_2$, thus the spectral radius
    \begin{align*}
        \rho(\Tb) = \gamma_1 = \frac{1}{2}\left[1+\beta-\eta\lambda+\sqrt{(1+\beta-\eta\lambda)^2-4\beta}\right].
    \end{align*}
    Then we justify the monotonicity in this case. We have
    \begin{align*}
        \frac{\partial \rho(\Tb)}{\partial (\eta\lambda)} = \frac{1}{2}\left[-1+\frac{-(1+\beta-\eta\lambda)}{\sqrt{(1+\beta-\eta\lambda)^2-4\beta}}\right] < 0,
    \end{align*}
    thus $\rho(\Tb)$ is monotonically decreasing with respect to $\eta\lambda$.

    \textbf{Case 2: $\Delta<0$} In this case, we have
    \begin{align*}
        \gamma_1 &= \frac{1}{2}\left[1+\beta-\eta\lambda+\sqrt{4\beta-(1+\beta-\eta\lambda)^2}i\right], \\ \gamma_2 &= \frac{1}{2}\left[1+\beta-\eta\lambda-\sqrt{4\beta-(1+\beta-\eta\lambda)^2}i\right],
    \end{align*}
    where $i$ is the imaginary unit. One can observe that $\gamma_1$ is the complex conjugate of $\gamma_2$. Thus we have the spectral radius is
    \begin{align*}
        \rho(\Tb_{t,j}) = \Abs{\gamma_1}=\Abs{\gamma_2} = \sqrt{\gamma_1\bar \gamma_1} = \sqrt{\gamma_1\gamma_2}\overset{\eqref{eq:roots_of_quad_equations}}{=}\sqrt{\beta}.
    \end{align*}
    In this case, $\rho(\Tb_{t,j})$ does not depend on $\eta\lambda$. Thus we finish the proof.
\end{proof}

\begin{lemma}[$\rho$ bounded by $\eta,\lambda,\beta$ in real-eigenvalue case]\label{lem:rho_eta_beta}
    For momentum matrix
    \begin{align*}
        \Tb=\left[\begin{matrix}
            1+\beta-\eta\lambda & -\beta \\ 1 & 0
        \end{matrix}\right],
    \end{align*}
    if $\Tb_{t,j}$ only has real eigenvalues, which is equivalent to that the discriminant $(1+\beta-\eta\lambda)^2-4\beta>0$,
    it holds that
    \begin{align}\label{eq:lem_rho_eta_beta}
        \rho(\Tb) \leq 1-\frac{\eta\lambda}{2}-\frac{\eta\lambda}{4\left(1-\sqrt{\beta}\right)}.
    \end{align}
\end{lemma}
\begin{proof}
    According to Lemma~\ref{lem:rho_monotonic}, the spectral radius of $\Tb$ is
    \begin{align*}
        \rho(\Tb)=\frac{1+\beta-\eta\lambda+\sqrt{(1+\beta-\eta\lambda)^2-4\beta}}{2}.
    \end{align*} 
    It holds that
\begin{equation*}
\begin{aligned}
    \rho(\Tb)&=\frac{1}{2}\left[1+\beta-\eta\lambda+\sqrt{(1+\beta-\eta\lambda)^2-4\beta}\right] \\
    &= 
    \frac{1}{2}\left[1+\beta-\eta\lambda+\sqrt{\left(1+\sqrt{\beta}\right)^2-\eta\lambda}\cdot\sqrt{\left(1-\sqrt{\beta}\right)^2-\eta\lambda}\right] \\
    &\le
    \frac{1}{2}\left[1+\beta-\eta\lambda+\left(1+\sqrt{\beta}\right)\sqrt{\left(1-\sqrt{\beta}\right)^2-\eta\lambda}\right] \\
    &=
    \frac{1}{2}\left[1+\beta-\eta\lambda+\left(1-\beta\right)\sqrt{1-\frac{\eta\lambda}{\left(1-\sqrt{\beta}\right)^2}}\right] \\
    &\overset{\eqref{eq:sqrt_1-x_le_1-x/2}}{\le}
    \frac{1}{2}\left[1+\beta-\eta\lambda+\left(1-\beta\right)\left(1-\frac{\eta\lambda}{2\left(1-\sqrt{\beta}\right)^2}\right)\right] \\
    &=
     \frac{1}{2}\left[2-\eta\lambda-\frac{(1-\beta)\eta\lambda}{2(1-\sqrt{\beta})^2}\right] 
    \le
    \frac{1}{2}\left[2-\eta\lambda-\frac{\eta\lambda}{2(1-\sqrt{\beta})}\right] \\
    &=
    1-\frac{\eta\lambda}{2}-\frac{\eta\lambda}{4\left(1-\sqrt{\beta}\right)}.
\end{aligned}
\end{equation*}

\end{proof}

\subsection{Proof of Theorem \ref{thm:main}}\label{appendix:speify_to_step_decay}

In this section, we specify the schedule to step decay and prove Theorem~\ref{thm:main}. We denote the step size of the $\ell$-th stage $\eta_\ell'$ and its corresponding momentum matrix $\Tb_{\ell,j}'$ to specify the stagewise case.

We first present a lemma to simplify our proof in the stagewise case.
\begin{lemma}\label{lem:stage_less_1}
    For all stage $\ell>1$, given matrices $\Tb_{\ell,j}'$ defined in Eqn.~\eqref{eq:T_tj_def}, if
    $
        \eta_\ell'\lambda_j > (1-\sqrt{\beta})^2,
    $
    and the length of the stage
        \begin{align*}
            \stagelength \geq \sqrt{\kappa}\ln \left(8T\right),
        \end{align*}
    it holds that
    \begin{align}\label{eq:lem_stage_less_1}
        \Norm{\left(\Tb_{\ell,j}'\right)^{\stagelength}} \le 1.
    \end{align}
\end{lemma}
\begin{proof}
    In this case, the eigenvalues of $\Tb_{\ell,j}'$ are not real and thus the spectral radius $\rho(\Tb_{\ell,j}')=\sqrt{\beta}$ as Lemma~\ref{lem:rho_monotonic} suggests.
    Thus it holds that
    \begin{align*}
     \Norm{\left(\Tb_{\ell,j}'\right)^{\stagelength}} \le& \Norm{\left(\Tb_{\ell,j}'\right)^{\stagelength}}_F 
        \overset{\eqref{eq:lem_1}}{\le}
        8\stagelength\rho\left(\Tb_{\ell,j}'\right)^{\stagelength} = 8\stagelength\left(\sqrt{\beta}\right)^{\stagelength} \\
        \leq&
        8T\left( 1-\frac{1}{\sqrt{\kappa}} \right)^{\sqrt{\kappa}\ln \left(8T\right)} \\
        \overset{\eqref{eq:lem_aux_e}}{\leq}&
        8T\cdot e^{-\ln \left(8T\right)} = 1,
    \end{align*}
    which concludes the proof.
\end{proof}
Then we are ready to prove the convergence of step decay schedule.

\theoremmain*

\begin{proof}

    From \eqref{eq:req_var_T}, the total iteration number $T$ satisfies that 
    \begin{align}\label{eq:main_T}
        T \geq 2C\sqrt{\kappa}\ln\left( 2^{14}T^6\kappa \right)\cdot\ln\left( 2^6T^4\right)\cdot \log_C\left(T\sqrt{\kappa}\right),
    \end{align}
    and we define an auxiliary constant for our proof that
    \begin{align}\label{eq:main_h}
        h\equiv h(T,\kappa)=4 \ln\left( 2^6T^4\right)\cdot \log_C\left(T\sqrt{\kappa}\right)\geq1.
    \end{align}
    From \eqref{eq:req_var_T}, we know that $T\geq 2C\sqrt{\kappa}$. Then with \eqref{eq:main_T} and \eqref{eq:main_h}, we can verify that the following requirements are satisfied. 
    %\textcolor{red}{I am not so sure how to write the requirements here.}
\begin{enumerate}
    \item From Lemma \ref{lem:stage_less_1}:
    \begin{align}\label{eq:req_sat_lem_6}
        \stagelength \geq \sqrt{\kappa}\ln \left(8T\right).
    \end{align}
    \textbf{Verify:} As $T\geq \sqrt{\kappa}$, it holds that
    \begin{align*}
        \stagelength \overset{\eqref{eq:req_var_k_l}}{=}
        \frac{T}{\log_C(T\sqrt{\kappa})}
        \overset{\eqref{eq:req_var_T}}{\ge} 
        2C\sqrt{\kappa}\ln(2^{14}T^8)\ln(2^6T^4) \geq \sqrt{\kappa}\ln(8T).
    \end{align*}
    \item From variance case 1.1, the final stage needs the variance to be small enough that $\eta_{\numstage}'L\leq h/(T\sqrt{\kappa})$:
    \begin{align}\label{eq:req_sat_small_real_T}
        \stagelength \leq \frac{T}{\log_C\left(\frac{T\sqrt{\kappa}}{h}\right)}.
    \end{align}
    \textbf{Verify:} As $T\geq \sqrt{\kappa}$ and $h\geq 1$, it holds that
    \begin{align*}
        \stagelength \overset{\eqref{eq:req_var_k_l}}{=}
        \frac{T}{\log_C(T\sqrt{\kappa})}
         \leq \frac{T}{\log_C\left(\frac{T\sqrt{\kappa}}{h}\right)}.
    \end{align*}
    \item From variance case 1.1:
    \begin{align}\label{eq:req_sat_small_real_kappa}
        \stagelength \leq \frac{T}{\log_C\left(\frac{4\kappa}{3}\right)+1}.
    \end{align}
    \textbf{Verify:} As $T\geq 2C\sqrt{\kappa}$, it holds that
    \begin{align*}
        \stagelength \overset{\eqref{eq:req_var_k_l}}{=}
        \frac{T}{\log_C(T\sqrt{\kappa})}
        \leq \frac{T}{\log_C\left(\frac{4\kappa}{3}\right)+1}.
    \end{align*}
    \item From variance case 1.2: 
    \begin{align}\label{eq:req_sat_large_real}
        \stagelength \geq \frac{4T}{h}\ln\left( 2^6T^4\right).
    \end{align}
    \textbf{Verify:} It holds that
    \begin{align*}
        h \overset{\eqref{eq:main_h}}{=}
        4 \ln\left( 2^6T^4\right)\cdot \log_C\left(T\sqrt{\kappa}\right) 
        \overset{\eqref{eq:req_var_k_l}}{=}
        \frac{4T}{\stagelength}\ln\left( 2^6T^4\right)
        .
    \end{align*}
    \item From variance case 2.1:
    \begin{align}\label{eq:req_sat_small_complex}
        \stagelength \geq {2C\sqrt{\kappa}}\ln\left(2^{12}T^6 \right).
    \end{align}
    \textbf{Verify:} It holds that
    \begin{align*}
        \stagelength \overset{\eqref{eq:req_var_k_l}}{=}
        \frac{T}{\log_C(T\sqrt{\kappa})}
        \overset{\eqref{eq:req_var_T}}{\ge} 
        2C\sqrt{\kappa}\ln(2^{14}T^8)\ln(2^6T^4) \geq {2C\sqrt{\kappa}}\ln\left(2^{12}T^6 \right).
    \end{align*}
    \item From variance case 2.2:
    \begin{align}\label{eq:req_sat_large_complex}
        \stagelength \geq \frac{\sqrt{\kappa}}{2}\ln\left( {2^{14}T^6}{\kappa}\right).
    \end{align}
    \textbf{Verify:} As $T\geq \sqrt{\kappa}$, it holds that
    \begin{align*}
        \stagelength \overset{\eqref{eq:req_var_k_l}}{=}
        \frac{T}{\log_C(T\sqrt{\kappa})}
        \overset{\eqref{eq:req_var_T}}{\ge} 
        2C\sqrt{\kappa}\ln(2^{14}T^8)\ln(2^6T^4) \geq \frac{\sqrt{\kappa}}{2}\ln\left( {2^{14}T^6}{\kappa}\right).
    \end{align*}
\end{enumerate}
    % In the proof, we will use $ks_{\ell_i}$ to denote the length of stage $i$ for better comprehension and completeness of the proof, e.g. $ks_{\ell_1}\equiv \stagelength$ is the length of the first stage. Note that $ks_{\ell_i}\equiv \stagelength$ for all $\ell=1,\cdots,\numstage$.
    
    Now we are ready to start our main analysis. From the former lemmas, it holds that
    \begin{align*}
        &2 \EE\left[f(\bw_T) - f(\bw_*)\right]
  \\
  \overset{\eqref{eq:loss-is-H-norm}}{=}&
  \left(\bw_T - \bw_*\right)^\top \bH \left(\bw_T - \bw_*\right)
  \\
  \overset{\eqref{eq:bias-var-decomp}}{=}&
  \EE\left[
    \begin{bmatrix}
      \bw_0 - \bw_* \\
      \bw_{-1} - \bw_*
    \end{bmatrix}^\top
    \bM_0^\top \bM_1^\top \dots \bM_{T-1}^\top
    \begin{bmatrix}
      \bH & \bO \\
      \bO & \bO
    \end{bmatrix}
    \bM_{T-1} \bM_{T-2} \dots \bM_0
    \begin{bmatrix}
      \bw_0 - \bw_* \\
      \bw_{-1} - \bw_*
    \end{bmatrix}
  \right]
  \\
  &+
  \sum_{\tau=0}^{T-1}
    \EE\left[
      \eta_\tau^2
      \begin{bmatrix}
        \bn_\tau \\
        \bzero
      \end{bmatrix}^\top
      \bM_{\tau+1}^\top \bM_{\tau+2}^\top \dots \bM_{T-1}^\top
      \begin{bmatrix}
        \bH & \bO \\
        \bO & \bO
      \end{bmatrix}
      \bM_{T-1} \bM_{T-2} \dots \bM_{\tau+1}
      \begin{bmatrix}
        \bn_\tau \\
        \bzero
      \end{bmatrix}
    \right]
  \\
  \overset{\eqref{eq:bound_var_with_T_norm}\eqref{eq:bound_bias_with_T_norm}}{\le}&
  \sum_{j=1}^d \lambda_j
    \Norm{
      \Tb_{T-1,j} \Tb_{T-2,j} \dots \Tb_{0,j}
    }^2
    \EE\Norm{
      \left(
        \bPi^\top \bV^\top
        \begin{bmatrix}
          \bw_0 - \bw_* \\
          \bw_{-1} - \bw_*
        \end{bmatrix}
      \right)_{2j-1:2j}
    }^2
  \\
  &+
  \sigma^2 \sum_{j=1}^d \lambda_j^2
    \sum_{\tau=0}^{T-1} \eta_\tau^2
      \Norm{\Tb_{T-1,j} \Tb_{T-2,j} \dots \Tb_{\tau+1,j}}^2.
    \end{align*}
If we denote bias and variance as
\begin{align*}
    B&\overset{\triangle}{=} \sum_{j=1}^d \lambda_j\Norm{\Tb_{T-1,j}\Tb_{T-2,j}...\Tb_{0,j}}^2\mathbb{E}\Norm{\left( \mathbf{\Pi}^\top\mathbf{V}^\top\left[\begin{matrix}
        w_0-w_* \\ w_{-1}-w_*
    \end{matrix}\right] \right)_{2j-1:2j}}^2,\\
    V&\overset{\triangle}{=} \sigma^2 \sum_{j=1}^d \lambda_j^2 \sum_{\tau=0}^{T-1} \eta_{\tau}^2\Norm{\Tb_{T-1,j}\Tb_{T-2,j}...\Tb_{\tau+1,j}}^2,
\end{align*}
we have the following results.

\textbf{(1) Bounding bias term:}

It holds that
    \begin{align*}
        B=& \sum_{j=1}^d \lambda_j\Norm{\Tb_{T-1,j}\Tb_{T-2,j}...\Tb_{0,j}}^2\mathbb{E}\Norm{\left( \mathbf{\Pi}^\top\mathbf{V}^\top\left[\begin{matrix}
        w_0-w_* \\ w_{-1}-w_*
    \end{matrix}\right] \right)_{2j-1:2j}}^2\\
    =& \sum_{j=1}^d \lambda_j\Norm{\Tb_{T-1,j}\Tb_{T-2,j}...\Tb_{0,j}}^2\mathbb{E}\Norm{\left( \mathbf{\Pi}^\top\mathbf{V}^\top\Tilde{\wb}_0 \right)_{2j-1:2j}}^2 \\
    =&
    \sum_{j=1}^d \lambda_j\Norm{\left(\Tb_{\numstage,j}'\right)^{k_{\numstage}}\left(\Tb_{n_{\ell-1},j}'\right)^{k_{n_{\ell-1}}}...\left(\Tb_{1,j}'\right)^{k_{1}}}^2 \mathbb{E}\Norm{\left( \mathbf{\Pi}^\top\mathbf{V}^\top\Tilde{\wb}_0 \right)_{2j-1:2j}}^2 \\
    \overset{\eqref{eq:lem_stage_less_1},\eqref{eq:lem_main}}{\le}&
    \sum_{j=1}^d \lambda_j\Norm{\left(\Tb_{1,j}'\right)^{k_{1}}}^2 \left(\frac{8}{\sqrt{(1+\beta-\eta_{T-1}\lambda_j)^2-4\beta}}\right)^2 \mathbb{E}\Norm{\left( \mathbf{\Pi}^\top\mathbf{V}^\top\Tilde{\wb}_0 \right)_{2j-1:2j}}^2 \\
    \le&
    \frac{256}{\eta_1'\mu}\sum_{j=1}^d \lambda_j\Norm{\left(\Tb_{1,j}'\right)^{k_{1}}}^2\mathbb{E}\Norm{\left( \mathbf{\Pi}^\top\mathbf{V}^\top\Tilde{\wb}_0 \right)_{2j-1:2j}}^2 \\
    %%%%%
    \overset{\eqref{eq:lem_1}}{\le}&
  \frac{256}{\eta_1'\mu}\sum_{j=1}^d \lambda_j
    \left(
      8 k_1 \rho\left(\Tb'_{1,j}\right)^{k_1}
    \right)^2 \cdot
    \EE\Norm{
      \left(
        \bPi^\top \bV^\top \bwp_0
      \right)_{2j-1:2j}
    }^2
  \\
  \overset{\text{Lem.}~\ref{lem:rho_monotonic}}{\le}&
  \frac{256}{\eta_1'\mu}\sum_{j=1}^d \lambda_j
    \left(
      8 k_1 \rho\left(\Tb'_{1,d}\right)^{k_1}
    \right)^2 \cdot
    \EE\Norm{
      \left(
        \bPi^\top \bV^\top \bwp_0
      \right)_{2j-1:2j}
    }^2
  \\
  =&
  \frac{256}{\eta_1'\mu}\left(
    8 k_1 \rho\left(\Tb'_{1,d}\right)^{k_1}
  \right)^2 \cdot
  \sum_{j=1}^d \lambda_j
    \EE\Norm{
      \left(
        \bPi^\top \bV^\top \bwp_0
      \right)_{2j-1:2j}
    }^2
  \\
  =&
  \frac{256}{\eta_1'\mu}\left(
    8 k_1 \rho\left(\Tb'_{1,d}\right)^{k_1}
  \right)^2
  \cdot
  \EE\left[\sum_{j=1}^d \lambda_j
    \left(
        \bPi^\top \bV^\top \bwp_0
    \right)_{2j-1:2j}^\top
    \left(
        \bPi^\top \bV^\top \bwp_0
    \right)_{2j-1:2j}
  \right]
  \\
  =&
  \frac{256}{\eta_1'\mu}\left(
    8 k_1 \rho\left(\Tb'_{1,d}\right)^{k_1}
  \right)^2
  \cdot
  \EE\left[\sum_{j=1}^d
    \left(
        \bPi^\top \bV^\top \bwp_0
    \right)_{2j-1:2j}^\top
    \begin{bmatrix}
      \lambda_j & 0 \\
      0 & \lambda_j
    \end{bmatrix}
    \left(
        \bPi^\top \bV^\top \bwp_0
    \right)_{2j-1:2j}
  \right]
  \\
  =&
  \frac{256}{\eta_1'\mu}\left(
    8 k_1 \rho\left(\Tb'_{1,d}\right)^{k_1}
  \right)^2
  \\
  &\cdot
  \EE\left[
    \begin{bmatrix}
      \left(\bPi^\top \bV^\top \bwp_0\right)_{1:2}
      \\
      \left(\bPi^\top \bV^\top \bwp_0\right)_{3:4}
      \\
      \vdots
      \\
      \left(\bPi^\top \bV^\top \bwp_0\right)_{2d-1:2d}
    \end{bmatrix}^\top
    \begin{bmatrix}
      \bS_1 & & & \\
      & \bS_2 & & \\
      & & \ddots & \\
      & & & \bS_d
    \end{bmatrix}
    \begin{bmatrix}
      \left(\bPi^\top \bV^\top \bwp_0\right)_{1:2}
      \\
      \left(\bPi^\top \bV^\top \bwp_0\right)_{3:4}
      \\
      \vdots
      \\
      \left(\bPi^\top \bV^\top \bwp_0\right)_{2d-1:2d}
    \end{bmatrix}
  \right],
  \\
  &\mbox{ with } \bS_j =
    \begin{bmatrix}
      \lambda_j & 0 \\
      0 & \lambda_j
    \end{bmatrix}
  \\
  =&
  \frac{256}{\eta_1'\mu}\left(
    8 k_1 \rho\left(\Tb'_{1,d}\right)^{k_1}
  \right)^2
  \cdot
  \EE\left[
    \left(\bPi^\top \bV^\top \bwp_0\right)^\top
    \left(
      \bLambda \otimes
      \begin{bmatrix}
        1 & 0 \\ 0 & 1
      \end{bmatrix}
    \right)
    \left(\bPi^\top \bV^\top \bwp_0\right)
  \right]
  \\
  =&
  \frac{256}{\eta_1'\mu}\left(
    8 k_1 \rho\left(\Tb'_{1,d}\right)^{k_1}
  \right)^2
  \cdot
  \EE\left[
    \bwp_0^\top \left(
      \bV \bPi
      \left(
        \bLambda \otimes
        \begin{bmatrix}
          1 & 0 \\ 0 & 1
        \end{bmatrix}
      \right)
    \bPi^\top \bV^\top \right)
    \bwp_0
  \right] ,
    %%%%%
    \end{align*}
    where the first inequality is because that Eqn.~\ref{eq:lem_stage_less_1} can be applied for the stages $\ell$ where $\Tb_{\ell,j}'$ has only complex eigenvalues and Eqn.~\ref{eq:lem_main} can be applied to the matrix product for all stages $\ell$ that $\Tb_{\ell,j}$ has only real eigenvalues.
    And the second inequality holds as
    \begin{align*}
    \sqrt{{(1+\beta-\eta_{\numstage}'\lambda_j)^2-4\beta}} &=\sqrt{(1-\sqrt{\beta})^2-\eta_{\numstage}'\lambda_j}\cdot\sqrt{(1+\sqrt{\beta})^2-\eta_{\numstage}'\lambda_j} \\
    &\ge
    \sqrt{(1-\sqrt{\beta})^2-\eta_{\numstage}'\lambda_j}
    = \sqrt{(1-\sqrt{\beta})^2-C^{-n_{\ell}+1}\cdot\eta_{1}'\lambda_j}
    \\
    &\overset{(\ref{eq:req_sat_small_real_kappa})}{\geq} 
    \sqrt{(1-\sqrt{\beta})^2-\frac{3}{4}(1-\sqrt{\beta})^2}
    =
    \frac{1}{2}(1-\sqrt{\beta }) 
    \overset{}{=} \frac{1}{2}\sqrt{\eta_1'\mu}.
    \end{align*}
    
    Notice that
\begin{align*}
  \bV \bPi
    \left(
      \bLambda \otimes
      \begin{bmatrix}
        1 & 0 \\ 0 & 1
      \end{bmatrix}
    \right)
  \bPi^\top \bV^\top
  =
  \begin{bmatrix}
    \bH & \bO \\
    \bO & \bH
  \end{bmatrix}
\end{align*}
since
\begin{align*}
  &\bPi^\top \bV^\top
  \begin{bmatrix}
    \bH & \bO \\
    \bO & \bH
  \end{bmatrix}
  \bV \bPi
  \\
  \overset{\eqref{eq:Pi_V_def}}{=}&
    \begin{bmatrix}
      \be_1^\top & \bzero^\top \\
      \bzero^\top & \be_1^\top \\
      \vdots & \\
      \be_d^\top & \bzero^\top \\
      \bzero^\top & \be_d^\top
    \end{bmatrix}
    \begin{bmatrix}
      \bU^\top & \bO \\
      \bO & \bU^\top
    \end{bmatrix}
    \begin{bmatrix}
      \bH & \bO \\
      \bO & \bH
    \end{bmatrix}
    \begin{bmatrix}
      \bU & \bO \\
      \bO & \bU
    \end{bmatrix}
    \begin{bmatrix}
      \be_1 & \bzero & \be_2 & \bzero & \dots & \be_d & \bzero \\
      \bzero & \be_1 & \bzero & \be_2 & \dots & \bzero & \be_d
    \end{bmatrix}
    \\
    \overset{\eqref{eq:H_eigendecomp}}{=}&
    \begin{bmatrix}
      \be_1^\top & \bzero^\top \\
      \bzero^\top & \be_1^\top \\
      \vdots & \\
      \be_d^\top & \bzero^\top \\
      \bzero^\top & \be_d^\top
    \end{bmatrix}
    \begin{bmatrix}
      \bLambda & \bO \\
      \bO & \bLambda
    \end{bmatrix}
    \begin{bmatrix}
      \be_1 & \bzero & \be_2 & \bzero & \dots & \be_d & \bzero \\
      \bzero & \be_1 & \bzero & \be_2 & \dots & \bzero & \be_d
    \end{bmatrix}
    \\
    =&
    \begin{bmatrix}
      \lambda_1 & & & & & & \\
      & \lambda_1 & & & & & \\
      & & \lambda_2 & & & & \\
      & & & \lambda_2 & & & \\
      & & & & \ddots & & \\
      & & & & & \lambda_d & \\
      & & & & & & \lambda_d
    \end{bmatrix}
    =
    \bLambda \otimes
    \begin{bmatrix}
      1 & 0 \\
      0 & 1
    \end{bmatrix}
\end{align*}
and $\bPi, \bV$ are orthogonal matrices (Eqn.~\eqref{eq:Pi_V_are_orthogonal}). We then further have
    
    \begin{align*}
    B\le&
    %%%%
    \frac{256}{\eta_1'\mu}\left(
    8 k_1 \rho\left(\bT'_{1,d}\right)^{k_1}
  \right)^2
  \cdot
  \EE\left[
    \bwp_0^\top
    \begin{bmatrix}
      \bH & \bO \\
      \bO & \bH
    \end{bmatrix}
    \bwp_0
  \right]
 \\
 \overset{\eqref{eq:bwp_bnp_bHp_def}}{=}&
  \frac{256}{\eta_1'\mu}\left(
    8 k_1 \rho\left(\bT'_{1,d}\right)^{k_1}
  \right)^2
  \cdot
  \EE\left[
    \begin{bmatrix}
      \bw_0 - \bw_* \\
      \bw_{-1} - \bw_*
    \end{bmatrix}^\top
    \begin{bmatrix}
      \bH & \bO \\
      \bO & \bH
    \end{bmatrix}
    \begin{bmatrix}
      \bw_0 - \bw_* \\
      \bw_{-1} - \bw_*
    \end{bmatrix}
  \right]
  \\
  =&
  \frac{256}{\eta_1'\mu}\left(
    8 k_1 \rho\left(\bT'_{1,d}\right)^{k_1}
  \right)^2
  \cdot
  \EE\left[
    \left(\bw_0 - \bw_*\right)^\top \bH  \left(\bw_0 - \bw_*\right)
    +
    \left(\bw_{-1} - \bw_*\right)^\top \bH  \left(\bw_{-1} - \bw_*\right)
  \right]
  \\
  \overset{\eqref{eq:loss-is-H-norm}}{=}&
  \frac{256}{\eta_1'\mu}\left(
    8 k_1 \rho\left(\bT'_{1,d}\right)^{k_1}
  \right)^2
  \cdot
  \EE\left[
    f(\bw_{-1}) + f(\bw_0) - 2f(\bw_*)
  \right]
  \\
    %%%%
    % \le&
    % \frac{256}{\eta_1'\mu}\Norm{\left(\Tb_{1,d}'\right)^{k_{1}}}^2\cdot\mathbb{E}\left[ f(w_0)+f(w_1)-2f(w_*) \right] \\
    % \le&
    % \frac{256}{\eta_1'\mu}\left(8k_1\rho\left(\Tb_{1,d}'\right)^{k_1} \right)^2\cdot\mathbb{E}\left[ f(w_0)+f(w_1)-2f(w_*) \right] \\
    =&
    \frac{256}{\eta_1'\mu}\left(8k_1\left(\sqrt{\beta}\right)^{k_1} \right)^2\cdot\mathbb{E}\left[ f(\bw_{-1})+f(\bw_0)-2f(\bw_*) \right] \\
    =&
    \frac{256}{\eta_1'\mu}\left(8k_1\left(1-\sqrt{\eta_1'\mu}\right)^{k_1} \right)^2\cdot\mathbb{E}\left[ f(\bw_{-1})+f(\bw_0)-2f(\bw_*) \right] \\
    \le&
    \exp\left( 14\ln 2+2\ln k_1-2\ln \left(\eta_1'\mu\right)-2k_1\sqrt{\eta_1'\mu} \right)\cdot\mathbb{E}\left[ f(\bw_{-1})+f(\bw_0)-2f(\bw_*) \right] \\
    =& \mathbb{E}\left[ f(\bw_{-1})+f(\bw_0)-2f(\bw_*) \right]\cdot  \exp\left( 14\ln 2+2\ln k_1-2\ln \left(\frac{1}{\kappa}\right) -\frac{2T}{\sqrt{\kappa}\log_C\left(T\sqrt{\kappa}\right)} \right) \\
    \le&
    \mathbb{E}\left[ f(\bw_{-1})+f(\bw_0)-2f(\bw_*) \right]\cdot  \exp\left( 14\ln 2+2\ln T+2\ln \kappa -\frac{2T}{\sqrt{\kappa}\log_C\left(T\sqrt{\kappa}\right)} \right) \\
    =&
    \mathbb{E}\left[ f(\bw_0)-f(\bw_*) \right]\cdot  \exp\left( 15\ln 2+2\ln T+2\ln \kappa -\frac{2T}{\sqrt{\kappa}\log_C\left(T\sqrt{\kappa}\right)} \right),
    \end{align*}
    
    where the second last inequality is because of the fact that $1-x\leq \exp(-x)$ for $x\geq0$ and the last equality is because of the setting of Algorithm~\ref{alg:multistage_sgdm} that $\bv_0=\bzero$ and thus $\bw_{-1}=\bw_{0}$.

\textbf{(2) Bounding variance term:}

We denote
\begin{align*}
    V_{t,j}&\overset{\triangle}{=} \eta_{t}^2\lambda_j^2 \Norm{\Tb_{T-1,j}\Tb_{T-2,j}...\Tb_{t+1,j}}^2
\end{align*}
in the following analysis. We first assume that the batch size $M=1$ in the main analysis and we will transfer the result to the general $M\geq1$ case.
% we desire for all $t,j$, $V_{t,j}$ can be bounded by
% \begin{align*}
%     V_{t,j} = \Tilde{O}\left(\frac{1}{T^2}\right).
% \end{align*}

We make use of Lemma~\ref{lem:rho_monotonic} and \ref{lem:rho_eta_beta} and divide the analysis of $V_{t,j}$ into 4 cases with respect to $\eta_t\lambda_j$ and equivalently the corresponding eigenvalues. The division of the 4 cases is due to two major boarders: $b_1=(1-\sqrt{\beta})^2$ and $b_2=h/(T\sqrt{\kappa})$. 
\begin{itemize}
    \item For momentum matrix $\Tb_{t,j}$ with $\eta_t\lambda_j>b_1$, $\Tb$ has complex eigenvalues and a large spectral radius that supports sufficient geometric decay of $V_{t,j}$, the variance generated at iteration $t$ on $\lambda_j$. Case 2.1 and Case 2.2 discuss this case.
    \item For momentum matrix $\Tb_{t,j}$ with $b_2<\eta_t\lambda_j\le b_1$, $\Tb_{t,j}$ has real eigenvalues but its spectral radius is still large enough to support sufficient geometric decay of $V_{t,j}$. Case 1.2 discusses this case.
    \item For momentum matrix $\Tb_{t,j}$ with $\eta_t\lambda_j\le b_2$, the corresponding step size $\eta_t\lambda_j$ is small enough to ensure the generated variance $V_{t,j}$ is small. Case 1.1 discusses this case.
\end{itemize}
 
Among this cases, we use requirement \eqref{eq:req_sat_small_real_kappa} to ensure that Case 1.1 exists and thus the variance can be small enough.
Let's discuss $V_{t,j}$ case by case then. 

\textbf{Case 1.1: Real eigenvalues with small step size.} We first consider the case that $\Tb_{t,j}$ only has real eigenvalues. This case is equivalent to that
\begin{align*}
    (1+\beta-\eta_{t}\lambda_j)^2- 4\beta \geq 0.
\end{align*}
It is also equivalent to that
\begin{align}\label{eq:var_eig_real}
    \eta_{t}\lambda_j \leq (1-\sqrt{\beta})^2=\eta_1'\mu.
\end{align}
And in this case we further assume that $\eta_t\lambda_j$ is small enough that
\begin{align}\label{eq:var_case_real_small_cond}
    \eta_t\lambda_j \le \frac{Ch}{T\sqrt{\kappa}},
\end{align}
where $h$ is defined in \eqref{eq:main_h}. We also discuss  From requirement~\eqref{eq:req_sat_small_real_T}, this case exists.  
Then from the fact that Frobenius norm of a matrix is always larger than $\ell_2$ norm and using Lemma~\ref{lem:matrix_main} we can obtain that
\begin{equation}\label{eq:var_case_real_small_tprod}
\begin{aligned}
    \Norm{\Tb_{T-1,j}\Tb_{T-2,j}...\Tb_{\tau+1,j}} \le& \Norm{\Tb_{T-1,j}\Tb_{T-2,j}...\Tb_{\tau+1,j}}_F \\
    \overset{\eqref{eq:lem_main}}{\le}&
    \min\left(8(T-1-\tau), \frac{8}{\sqrt{{(1+\beta-\eta_{T-1}\lambda_j)^2-4\beta}}} \right)\rho(\Tb_{T-1,j})^{T-1-\tau} \\
    \le& 
    \frac{8}{\sqrt{{(1+\beta-\eta_{T-1}\lambda_j)^2-4\beta}}}\rho(\Tb_{T-1,j})^{T-1-\tau} \\
    =&
    \frac{8}{\sqrt{{(1+\beta-\eta_{\numstage}'\lambda_j)^2-4\beta}}}\rho(\Tb_{T-1,j})^{T-1-\tau}.
\end{aligned}
\end{equation}
Then we analyze the right hand side term by term. First it holds that
\begin{equation}\label{eq:var_product_real}
\begin{aligned}
    \sqrt{{(1+\beta-\eta_{\numstage}'\lambda_j)^2-4\beta}} &=\sqrt{(1-\sqrt{\beta})^2-\eta_{\numstage}'\lambda_j}\cdot\sqrt{(1+\sqrt{\beta})^2-\eta_{\numstage}'\lambda_j} \\
    &\ge
    \sqrt{(1-\sqrt{\beta})^2-\eta_{\numstage}'\lambda_j}
    = \sqrt{(1-\sqrt{\beta})^2-C^{-n_{\ell}+1}\cdot\eta_{1}'\lambda_j}
    \\
    &\overset{(\ref{eq:req_sat_small_real_kappa})}{\geq} 
    \sqrt{(1-\sqrt{\beta})^2-\frac{3}{4}(1-\sqrt{\beta})^2}
    =
    \frac{1}{2}(1-\sqrt{\beta }) 
    \overset{}{=} \frac{1}{2}\sqrt{\eta_1'\mu}. 
\end{aligned}
\end{equation}
From Lemma~\ref{lem:rho_monotonic}, $\rho(\Tb_{t,j})<1$ holds for all $t,j$. Combining all above we can obtain that for $t,j$ satisfying $\eta_t\lambda_j\leq h/\left(T\sqrt{\kappa}\right)$, it holds that
\begin{equation*}
\begin{aligned}
    V_{t,j}&=\eta_{t}^2\lambda_j^2 \Norm{\Tb_{T-1,j}\Tb_{T-2,j}...\Tb_{t+1,j}}^2 \\
    &\overset{\eqref{eq:var_case_real_small_cond},\eqref{eq:var_case_real_small_tprod}}{\le}
    \left(\frac{Ch}{T\sqrt{\kappa}}\right)^2\left( \frac{8}{\sqrt{{(1+\beta-\eta_{\numstage}'\lambda_j)^2-4\beta}}}\right)^2 \\
    &\overset{\eqref{eq:var_product_real}}{\le}
    \frac{C^2h^2}{T^2\kappa}\frac{64}{\left(\frac{1}{2}\sqrt{\eta_1'\mu}\right)^2} = \frac{256C^2h^2}{T^2\kappa\eta_1'\mu}=\frac{256C^2h^2}{T^2},
\end{aligned}
\end{equation*}
where the last equality holds as $\eta_1'=1/L$.
% Moreover, if $t$ is an iteration in the last stage that satisfies $\eta_t\lambda_j\geq h/(T\sqrt{\kappa})$, it holds that
% \begin{align}\label{eq:var_case_real_small_cond_2}
%     \frac{h}{T\sqrt{\kappa}} \le \eta_t\lambda_j \leq \frac{Ch}{T\sqrt{\kappa}}.
% \end{align}
% Then we have 
% \begin{align*}
%     V_{t,j}&=\eta_{t}^2\lambda_j^2 \Norm{\Tb_{T-1,j}\Tb_{T-2,j}...\Tb_{t+1,j}}^2 \\
%     &\overset{\eqref{eq:var_case_real_small_cond_2},\eqref{eq:var_case_real_small_tprod}}{\le}
%     \left(\frac{Ch}{T\sqrt{\kappa}}\right)^2\left( \frac{8}{\sqrt{{(1+\beta-\eta_{\numstage}'\lambda_j)^2-4\beta}}}\right)^2 \\
%     &\overset{\eqref{eq:var_product_real}}{\le}
%     \frac{C^2h^2}{T^2\kappa}\frac{64}{\left(\frac{1}{2}\sqrt{\eta_1'\mu}\right)^2} = \frac{256C^2h^2}{T^2}.
% \end{align*}

\textbf{Case 1.2: Real eigenvalues with large step size.} Then we continue to consider the case that $\Tb_{t,j}$ only has real eigenvalues. Similar to case 1.1, this case is equivalent to that
\begin{align*}
    (1+\beta-\eta_{t}\lambda_j)^2- 4\beta \geq 0.
\end{align*}
It is also equivalent to that
\begin{align}\label{eq:var_eig_real_2}
    \eta_{t}\lambda_j \leq (1-\sqrt{\beta})^2=\eta_1'\mu.
\end{align}
But in this case we further assume that $\eta_t\lambda_j$ is large enough that
\begin{align*}
    \eta_t\lambda_j \ge \frac{Ch}{T\sqrt{\kappa}}
\end{align*}
as the case that $\eta_t\lambda_j\le Ch/(T\sqrt{\kappa})$ has been discussed in case 1.1. From requirement~\eqref{eq:req_var_T} we know that such stage exists. Then according to Lemma \ref{lem:rho_eta_beta}, with $\eta_t\lambda_j\geq h/(T\sqrt{\kappa})$, it holds that
\begin{align*}
    \rho(\Tb_{t,j})&=\frac{1}{2}\left[1+\beta-\eta_t\lambda_j+\sqrt{(1+\beta-\eta_t\lambda_j)^2-4\beta} \right] \\
    &\overset{\eqref{eq:lem_rho_eta_beta}}{\le}
    1-\frac{\eta_t\lambda_j}{4\left(1-\sqrt{\beta}\right)} = 1 - \frac{\eta_t\lambda_j}{4\sqrt{\eta_1'\mu}} \\
    &\leq
    1 - \frac{h}{4T}.
\end{align*}
Thus if we denote $t_*$ to be the first iteration that $\eta_{t_*}\lambda_j\leq h/\left(T\sqrt{\kappa}\right)$, it holds that
\begin{equation*}
\begin{aligned}
    V_{t,j}&=\eta_{t}^2\lambda_j^2 \Norm{\Tb_{T-1,j}\Tb_{T-2,j}...\Tb_{t+1,j}}^2 \\
    &\le
    \eta_{t}^2\lambda_j^2\Norm{\Tb_{T-1,j}\Tb_{T-2,j}...\Tb_{t_*,j}}^2\Norm{\Tb_{t_*-1,j}\Tb_{t_*-2,j}...\Tb_{t+1,j}}^2 \\
    &\overset{\eqref{eq:lem_2}}{\le}
    \eta_{t}^2\lambda_j^2\Norm{\Tb_{T-1,j}\Tb_{T-2,j}...\Tb_{t_*,j}}^2\Norm{\Tb_{t_*,j}^{t_*-t-1}}^2 \\
    &\overset{\eqref{eq:var_product_real},\eqref{eq:lem_1}}{\le}
   \eta_{t}^2\lambda_j^2\left(\frac{8}{\sqrt{{(1+\beta-\eta_{\numstage}'\lambda_j)^2-4\beta}}}\right)^2\cdot\left(8(t_*-t-1)\right)^2\left( 1-\frac{h}{4T}\right)^{2(t_*-t-1)} \\
    &\overset{\eqref{eq:var_product_real}}{\le}
    \eta_{t}^2\lambda_j^2\frac{2^{12}T^2}{\left(\frac{1}{2}\sqrt{\eta_1'\mu}\right)^2}\left( 1-\frac{h}{4T}\right)^{2(t_*-t-1)} 
    \le
    \eta_{t}^2\lambda_j^2\frac{2^{12}T^2}{\left(\frac{1}{2}\sqrt{\eta_1'\mu}\right)^2}\left( 1-\frac{h}{4T}\right)^{2\stagelength} \\
    &\overset{\eqref{eq:req_sat_large_real},\eqref{eq:var_eig_real_2}}{\le}
    \left(\eta_1'\mu\right)^2\frac{2^{12}T^2}{\left(\frac{1}{2}\sqrt{\eta_1'\mu}\right)^2}\left( 1-\frac{h}{4T}\right)^{\frac{4T}{h}\cdot 2\ln(2^6T^4)}
    \overset{\eqref{eq:lem_aux_e}}{\le}
    \frac{256}{T^2\kappa},
\end{aligned}
\end{equation*}
where the third last inequality holds as $\eta_t\lambda_j\geq Ch/(T\sqrt{\kappa})$, which suggests that there is at least one stage between $t$ and $t_*$.

\textbf{Case 2.1: Complex eigenvalues with small step size.} Then we consider the case that $\Tb_{\ell,j}'$ has complex eigenvalues but $\Tb_{\ell+1,j}'$ has real eigenvalues, which is equivalent to that
\begin{equation}\label{eq:var_eig_complex}
\begin{aligned}
    C(1-\sqrt{\beta})^2 \geq \eta_t\lambda_j > (1-\sqrt{\beta})^2 &= \eta_1'\mu.
\end{aligned}
\end{equation}
 Thus if we denote $t_*$ to be the first iteration in the $\ell+1$ stage, consider the spectral radius of the $\ell+1$ stage, it holds that
\begin{equation}\label{eq:var_small_complex_rho}
\begin{aligned}
    \rho(\Tb_{t_*,j})=\rho(\Tb_{\ell+1,j}')=&\frac{1}{2}\left[ 1+\beta-\eta_{\ell+1}'\lambda_j + \sqrt{(1+\beta-\eta_{\ell+1}'\lambda_j)^2-4\beta} \right] \\
    \overset{\eqref{eq:lem_rho_eta_beta}}{\le}&
    1-\frac{\eta_{\ell+1}'\lambda_j}{4(1-\sqrt{\beta})} = 1 - \frac{\eta_{\ell+1}'\lambda_j}{4\sqrt{\eta_1'\mu}} \\
    \overset{\eqref{eq:var_eig_complex}}{\le}&
    1 - \frac{\eta_1'\mu}{4C\sqrt{\eta_1'\mu}} = 1 - \frac{\sqrt{\eta_1'\mu}}{4C}.
\end{aligned}
\end{equation}

Thus we have
\begin{align*}
    V_{t,j}&=\eta_{t}^2\lambda_j^2 \Norm{\Tb_{T-1,j}\Tb_{T-2,j}...\Tb_{t+1,j}}^2 \\
    &\le
    (C\eta_{\ell+1}'\lambda_j)^2\Norm{\Tb_{T-1,j}\Tb_{T-2,j}...\Tb_{t+1,j}}^2 
    \overset{(\ref{eq:var_eig_real})}{\le}
    C^2\left(\eta_1'\mu\right)^2\Norm{\Tb_{T-1,j}\Tb_{T-2,j}...\Tb_{t+1,j}}^2 \\
    &\le
    C^2(\eta_1'\mu)^2\Norm{\Tb_{T-1,j}\Tb_{T-2,j}...\Tb_{t_*+\stagelength,j}}^2 \Norm{\Tb_{t_*+\stagelength-1,j}\Tb_{T-2,j}...\Tb_{t_*,j}}^2 \Norm{\Tb_{t_*-1,j}\Tb_{t_*-2,j}...\Tb_{t+1,j}}^2 \\
    &\overset{(\ref{eq:var_product_real})}{\le}
    C^2(\eta_1'\mu)^2\cdot \frac{256}{\eta_1'\mu} \Norm{\Tb_{t_*+\stagelength-1,j}\Tb_{T-2,j}...\Tb_{t_*,j}}^2 \Norm{\Tb_{t_*-1,j}\Tb_{t_*-2,j}...\Tb_{t+1,j}}^2 \\
    &=
    256C^2\eta_1'\mu\cdot\Norm{\left(\Tb_{\ell+1,j}'\right)}^2\Norm{\Tb_{t_*-1,j}\Tb_{t_*-2,j}...\Tb_{t+1,j}}^2 \\
    &\overset{(\ref{eq:var_small_complex_rho})}{\le}
    256C^2\eta_1'\mu\cdot\left(8T\left(1-\frac{\sqrt{\eta_1'\mu}}{4C}\right)^{\stagelength+1}\right)^2\Norm{\Tb_{t_*-1,j}\Tb_{t_*-2,j}...\Tb_{t+1,j}}^2 \\
    &\le 256C^2\eta_1'\mu\cdot\left(8T\left(1-\frac{\sqrt{\eta_1'\mu}}{4C}\right)^{\stagelength}\right)^2\cdot \max_{0\le i \le \stagelength} \Norm{\left(\Tb_{\ell,j}'\right)^{\stagelength-i}}^2 \\
    &\overset{\eqref{eq:lem_1}}{\le}
    256C^2\eta_1'\mu\cdot\left(8T\left(1-\frac{\sqrt{\eta_1'\mu}}{4C}\right)^{\stagelength}\right)^2\cdot(8\stagelength)^2 
    \overset{}{\le}
    {2^{20}C^2T^4\eta_1'\mu} \left(1-\frac{\sqrt{\eta_1'\mu}}{4C}\right)^{2\stagelength} \\
    &\overset{\eqref{eq:req_sat_small_complex},\eqref{eq:var_eig_complex}}{\le}
    {2^{20}C^2T^4\eta_1'\mu} \left(1-\frac{\sqrt{\eta_1'\mu}}{4C}\right)^{4C\sqrt{\kappa}\cdot\ln(2^{12}T^6)}
    \overset{\eqref{eq:lem_aux_e}}{\le}
    \frac{256C^2}{T^2},
\end{align*}
where in the fifth inequality we introduce $\max_i$ because the leading stage $\ell$ may be incomplete.

\textbf{Case 2.2: Complex eigenvalues with large step size.} Finally we consider the case that $\Tb_{t,j}$ has complex eigenvalues which is equivalent to Equation (\ref{eq:var_eig_complex}) and also $\Tb_{t,j}$ is far away from the boundary, namely, the step size of the next stage can still satisfy Equation~(\ref{eq:var_eig_complex}). In this stage, it holds that
\begin{align}\label{eq:var_eig_complex_big}
    \eta_t\lambda_j > C\eta_1'\mu.
\end{align}
Thus if we denote $t_*$ the first iteration of the $\ell_*+1$ stage where $\ell_*$ is the last stage that $\Tb_{\ell,j}'$ has complex eigenvalues, namely, $\eta_{\ell_*}'\lambda_j\in (\eta_1'\mu,C\eta_1'\mu]$.
We denote $\ell_t$ the stage where iteration $t$ is in. Thus it holds that $\ell_t\le \ell_*-1$. Then we consider that
    \begin{align*}
    V_{t,j}=&\eta_{t}^2\lambda_j^2 \Norm{\Tb_{T-1,j}\Tb_{T-2,j}...\Tb_{t+1,j}}^2 \\
    \overset{(\ref{eq:req_var_eta_1})}{\le}&
    \Norm{\Tb_{T-1,j}...\Tb_{t_*,j}}^2\Norm{\left(\Tb_{\ell_*,j}'\right)^{\stagelength}}^2\Norm{\left(\Tb_{\ell_*-1,j}'\right)^{\stagelength}}^2...\Norm{\left( \Tb_{\ell_{t}+1,j}'\right)^{\stagelength}}^2 \\
    &\cdot\max_{0\le i\le \stagelength}\Norm{\left( \Tb_{\ell_{t},j}'\right)^{\stagelength-i}}^2 \\
    \overset{\eqref{eq:var_case_real_small_tprod}}{\le}&
    \left(\frac{8}{\sqrt{(1+\beta-\eta_{\numstage}'\lambda_j)^2-4\beta}}\right)^2\Norm{\left(\Tb_{\ell_*,j}'\right)^{\stagelength}}^2\Norm{\left(\Tb_{\ell_*-1,j}'\right)^{\stagelength}}^2...\Norm{\left( \Tb_{\ell_{t}+1,j}'\right)^{\stagelength}}^2 \\
    &\cdot\max_{0\le i\le \stagelength}\Norm{\left( \Tb_{\ell_{t},j}'\right)^{\stagelength-i}}^2 \\
    \overset{(\ref{eq:var_product_real})}{\le}&
    \frac{256}{\eta_1'\mu}\Norm{\left(\Tb_{\ell_*,j}'\right)^{\stagelength}}^2\Norm{\left(\Tb_{\ell_*-1,j}'\right)^{\stagelength}}^2...\Norm{\left( \Tb_{\ell_{t}+1,j}'\right)^{\stagelength}}^2
    \cdot\max_{0\le i\le \stagelength}\Norm{\left( \Tb_{\ell_{t},j}'\right)^{\stagelength-i}}^2\\
    \overset{\eqref{eq:lem_stage_less_1}}{\le}&
    \frac{256}{\eta_1'\mu}\Norm{\left(\Tb_{\ell_*,j}'\right)^{\stagelength}}^2\cdot\max_{0\le i\le \stagelength}\Norm{\left( \Tb_{\ell_{t},j}'\right)^{\stagelength-i}}^2 \\
    \overset{\eqref{eq:lem_1}}{\le}&
    \frac{256}{\eta_1'\mu}\left( 8\stagelength\left(\sqrt{\beta}\right)^{\stagelength} \right)^2 \left( 8\stagelength \right)^2 
    \overset{}{\le}
    \frac{2^{20}\stagelength^4}{\eta_1'\mu}\left(1-\sqrt{\eta_1'\mu}\right)^{2\stagelength} \\
    \overset{\eqref{eq:req_sat_large_complex},\eqref{eq:var_eig_complex_big}}{\le}&
    \frac{2^{20}T^4}{\eta_1'\mu}\left(1-\sqrt{\eta_1'\mu}\right)^{\sqrt{\kappa}\ln(2^{14}T^6\kappa)}
    \overset{\eqref{eq:lem_aux_e}}{\le}
    \frac{256}{T^2},
    \end{align*}
where in the first inequality we introduce $\max_i$ because the leading stage $\ell_t$ may be incomplete.

Therefore, combining the four cases, we have the result
\begin{align*}
    V = \sigma^2\sum_{j=1}^{d}\sum_{t=0}^{T-1}V_{t,j} \le& \sigma^2\sum_{j=1}^{d}\sum_{t=0}^{T-1} \frac{256C^2h^2}{T^2}   \overset{\eqref{eq:main_h}}{\le}\frac{4096C^2d\sigma^2}{T}\ln^2\left( 2^6T^4\right)\cdot \log_C^2\left(T\sqrt{\kappa}\right),
\end{align*}
which concludes the proof in the case that batch size $M=1$. 

For general batch size $M \ge 1$, the gradient noise
\begin{align*}
  \bn'_t
  =&
  \nabla_\bw f(\bw_t) - \frac{1}{|\mathcal{B}_t|} \sum_{\xi \in \mathcal{B}_t} \nabla_\bw f(\bw_t, \xi)
  \\
  =&
  \frac{1}{M} \sum_{\xi \in \mathcal{B}_t} \left(\nabla_\bw f(\bw_t) - \nabla_\bw f(\bw_t, \xi)\right)
  \\
  =&
  \frac{1}{M} \sum_{i=0}^{M-1} \bn_{tM + i}
\end{align*}
satisfies
\begin{align*}
  &\EE\left[\bn'_t \left(\bn'_t\right)^\top\right]
  \\
  =&
  \EE\left[
    \left(
      \frac{1}{M} \sum_{i=0}^{M-1} \bn_{tM + i}
    \right)
    \left(
      \frac{1}{M} \sum_{i'=0}^{M-1} \bn_{tM + i'}^\top
    \right)
  \right]
  \\
  =&
  \frac{1}{M^2} \cdot
  \EE\left[
    \sum_{i=0}^{M-1} \bn_{tM + i} \bn_{tM + i}^\top
    +
    \sum_{i \neq i'} \bn_{tM + i} \bn_{tM + i'}^\top
  \right]
  \\
  =&
  \frac{1}{M^2} \cdot \left(
    \sum_{i=0}^{M-1} \EE\left[
      \bn_{tM + i} \bn_{tM + i}^\top
    \right] 
    +
    \sum_{i \neq i'} \EE\left[
      \bn_{tM + i} \bn_{tM + i'}^\top
    \right]
  \right)
  \\
  =&
  \frac{1}{M^2} \cdot \left(
    \sum_{i=0}^{M-1} \EE\left[
      \bn_{tM + i} \bn_{tM + i}^\top
    \right]
    +
    \sum_{i \neq i'}
      \EE\left[\bn_{tM + i}\right]
      \EE\left[ \bn_{tM + i'}^\top\right]
  \right)
  \mcomment{Assumption~\ref{ass:independent_noise}}
  \\
  =&
  \frac{1}{M^2} \cdot
    \sum_{i=0}^{M-1} \EE\left[
      \bn_{tM + i} \bn_{tM + i}^\top
    \right]
  \mcomment{Assumption~\ref{ass:unbiased_noise}}
  \\
  \preceq&
  \frac{1}{M^2} \cdot
    M \cdot \sigma^2 \bH
  \mcomment{Assumption~\ref{ass:anisotropic_noise}}
  \\
  =&
  \frac{\sigma^2 \bH}{M}.
\end{align*}
So for general $M \ge 1$, it is equivalent to replacing the noise term $\sigma^2$ with $\sigma^2 / M$. Thus it holds that
\begin{align*}
    V \leq \frac{4096C^2d\sigma^2}{MT}\ln^2\left( 2^6T^4\right)\cdot \log_C^2\left(T\sqrt{\kappa}\right).
\end{align*}
Combining the bias and variance term, we can verify that Theorem~\ref{thm:main} holds.

\end{proof}

\section{Preliminary: Useful Lemmas}
\label{appendix:useful_lemmas}

We provide some basic mathematical tools relevant to our proof. It serves as a manual section and can be skipped if one is already familiar with those tools.

\subsection{Random Variables}

\begin{lemma}
\label{lem:independent_matrix_prod}
If each pair of entries of matrix $\bX$ and matrix $\bY$ are independent from each other, then
\begin{equation}
\label{eq:independent_matrix_prod}
\begin{aligned}
  \EE\left[\bX \bY\right] =
  \EE\left[\bX\right] \EE\left[\bY\right]
\end{aligned}
\end{equation}
\begin{proof}
  For $\forall i,j$,
  \begin{align*}
    \EE\left[\bX\bY\right]_{i,j}
    =& \EE\left[\left(\bX\bY\right)_{i,j}\right]
    = \EE\left[\sum_k \bX_{i,k} \bY_{k,j}\right]
    = \sum_k \EE\left[ \bX_{i,k} \bY_{k,j} \right]
    = \sum_k \EE\left[ \bX_{i,k}\right] \EE\left[ \bY_{k,j} \right]
    \\
    =& \sum_k \EE\left[ \bX\right]_{i,k} \EE\left[ \bY \right]_{k,j}
    = \left(\EE\left[ \bX\right] \EE\left[ \bY\right]\right)_{i,j}
  \end{align*}
  Thus $\EE[\bX\bY] = \EE[\bX]\EE[\bY]$.
\end{proof}
\end{lemma}

\subsection{Loewner Order}

In Loewner order, $\bX \preceq \bY$ if and only if $\forall \bz, \bz^\top (\bY - \bX) \bz \ge 0$.

\begin{lemma}
\label{lem:loewner_replace_partial_in_2x2}
Given two $2\times2$ block matrices $\bX$ and $\bY$ where
\begin{align*}
  \bX_{11} \preceq \bY_{11}, \quad
  \bX_{12} = \bY_{12}, \quad
  \bX_{21} = \bY_{21}, \quad
  \bX_{22} = \bY_{22},
\end{align*}
then $\bX \preceq \bY$.
\end{lemma}
\begin{proof}
  \begin{align*}
    \text{For }
    \forall \bz =
    \begin{bmatrix} \bz_1 \\ \bz_2 \end{bmatrix},
    \quad
    \bz^\top (\bY - \bX) \bz
    =&
    \begin{bmatrix}
      \bz_1^\top & \bz_2^\top
    \end{bmatrix}
    \begin{bmatrix}
      \bY_{11} - \bX_{11} & \bO \\
      \bO & \bO
    \end{bmatrix}
    \begin{bmatrix}
      \bz_1 \\ \bz_2
    \end{bmatrix}
    =
    \bz_1^\top (\bY_{11} - \bX_{11}) \bz_1
    \ge
    0,
  \end{align*}
  where the inequality comes from $\bX_{11} \preceq \bY_{11}$. Therefore $\bX \preceq \bY$ according to the definition of Loewner order.
\end{proof}

\begin{lemma}
\label{lem:loewner_outside_product}
If $\bX \preceq \bY \in \RR^{n\times n}$, then for $\forall \bC \in \RR^{n \times m}$,
\begin{align*}
  \bC^\top \bX \bC \preceq \bC^\top \bY \bC
\end{align*}
\end{lemma}
\begin{proof}
For $\forall \bz \in \RR^m$,
\begin{align*}
  \bz^\top \left(\bC^\top \bY \bC - \bC^\top \bX \bC\right) \bz
  =
  (\bC \bz)^\top (\bY - \bX) (\bC \bz)
  \ge 0
  \quad \Rightarrow \quad
  \bC^\top \bX \bC \preceq \bC^\top \bY \bC
\end{align*}
\end{proof}

\begin{lemma}
\label{lem:loewner_trace}
If $\bX \preceq \bY$, then $\tr(\bX) \le \tr(\bY)$.
\end{lemma}
\begin{proof}
  Denote
  \begin{align*}
    \be_i =
    \begin{bmatrix}\strut\smash{
      0 \dots 0 \underbrace{1}_{i-th} 0 \dots 0
    }\end{bmatrix}^\top \in \RR^d,
  \end{align*}
  \begin{align*}
     \bY_{ii} - \bX_{ii}
     = \be_i^\top (\bY - \bX) \be_i
     \ge 0
     \quad \text{ for } \forall i
     \quad \Rightarrow \quad
     \tr(\bY) - \tr(\bX)
     = \sum_{i=1}^d \left(\bY_{ii} - \bX_{ii}\right)
     \ge 0
  \end{align*}
\end{proof}

\subsection{Quadratic Equations}

\begin{lemma}
\label{lem:roots_of_quad_equations}
\textbf{(Roots of quadratic equations)}
If $x_1, x_2 \in \CC$ are roots of equation $x^2 + Bx + C = 0$, where $B, C \in \RR$, then
\begin{equation}
\label{eq:roots_of_quad_equations}
\begin{aligned}
  (1) \quad &
  x_1 + x_2 = -B,
  \\
  (2) \quad &
  x_1 x_2 = C
  \\
  (3) \quad &
  x_1 \conjugate{x_2} + \conjugate{x_1} x_2
  =
  \begin{cases}
    2C, \quad &\mbox{ if $x_1, x_2$ are real }
    \\
    B^2 - 2C, \quad &\mbox{ if $x_1, x_2$ are imaginary }
  \end{cases}
  \\
  (4) \quad &
  \modulus{x_1}^2 + \modulus{x_2}^2
  =
  \begin{cases}
    B^2 - 2C, \quad &\mbox{ if $x_1, x_2$ are real }
    \\
    2C, \quad &\mbox{ if $x_1, x_2$ are imaginary }
  \end{cases}
\end{aligned}
\end{equation}
\end{lemma}
\begin{proof}
  (1) and (2) are a special case of the famous Vieta's formulas, which can be directly obtained from
  \begin{align*}
    x^2 + Bx + C
    =
    (x - x_1)(x - x_2)
    =
    x^2 - (x_1 + x_2) x + (x_1 x_2).
  \end{align*}
  (3) can be derived from the form of $x_1$ and $x_2$. The roots of $x^2 + Bx + C = 0$ is
  \begin{align*}
    x
    = \frac{1}{2} \left(
      -B \pm \sqrt{B^2 - 4C}
    \right),
  \end{align*}
  If $x_1$ and $x_2$ are both real, then
  \begin{align*}
    x_1 \conjugate{x_2}
    + \conjugate{x_1} x_2
    =
    x_1 x_2 + x_1 x_2
    =
    2 x_1 x_2
    =
    2C.
  \end{align*}
  Otherwise $x_1$ and $x_2$ are both imaginary, then  $x_1 = \conjugate{x_2}$, $x_2 = \conjugate{x_1}$, which follows
  \begin{align*}
    x_1 \conjugate{x_2}
    + \conjugate{x_1} x_2
    =
    x_1^2 + x_2^2
    =
    (x_1 + x_2)^2 - 2 x_1 x_2
    =
    B^2 - 2C.
  \end{align*}
  
  (4) can be obtained from (1) and (3) by
  \begin{align*}
    \modulus{x_1}^2 + \modulus{x_2}^2
    =&
    x_1 \conjugate{x_1}
    +
    x_2 \conjugate{x_2}
    =
    (x_1 + x_2)\left(\conjugate{x_1} + \conjugate{x_2}\right)
    - \left(x_1 \conjugate{x_2} + \conjugate{x_1} x_2\right)
    =
    B^2 - \left(x_1 \conjugate{x_2} + \conjugate{x_1} x_2\right)
    \\
    =&
    \begin{cases}
      B^2 - 2C, \quad &\mbox{ if $x_1, x_2$ are real }
      \\
      2C, \quad &\mbox{ if $x_1, x_2$ are imaginary }
    \end{cases}
  \end{align*}
\end{proof}

\subsection{Bounding Special Functions}

\begin{lemma}
\begin{equation}
\label{eq:sqrt_1-x_le_1-x/2}
\begin{aligned}
  \sqrt{1-x} \le 1 - \frac{x}{2}
  \quad \mbox{ holds for }
  \forall
  x \in [0, 1].
\end{aligned}
\end{equation}
\end{lemma}
\begin{proof}
  Only in this lemma, denote
  \begin{align*}
    f(x) \triangleq \sqrt{1-x} - \left(1 - \frac{x}{2}\right),
  \end{align*}
  then for $\forall x \in [0, 1)$,
  \begin{align*}
    &f(0) = 0,
    \quad \frac{d}{dx} f(x)
    =
    -\frac{1}{2} \cdot \frac{1}{\sqrt{1-x}} + \frac{1}{2}
    =
    \frac{1}{2}\left(1 - \frac{1}{\sqrt{1-x}}\right)
    \le 0
    \\
    \Rightarrow \quad&
    f(x) \le f(0) = 0
    \\
    \Rightarrow \quad&
    \sqrt{1 - x} \le 1 - \frac{x}{2}
  \end{align*}
  For $x=1$, we also have $\sqrt{1 - x} = 0 \le 1/2 = 1 - x/2$.
\end{proof}

\begin{lemma}
% [Auxiliary fact for bounding stage power, maybe should be moved to appendix \ref{appendix:useful_lemmas}]
\label{lem:aux_e}
    For $x\in[1,+\infty)$,
    it holds that
    \begin{align}\label{eq:lem_aux_e}
        f(x) = \left( 1-\frac{1}{x} \right)^x \leq \frac{1}{e}.
    \end{align}
\end{lemma}
\begin{proof}
    The lemma is equivalent to that
    \begin{align*}
        x\ln\left(1-\frac{1}{x}\right) \leq -1
    \end{align*}
    Denote $g(t) = \ln(1-t)+t,\quad t\in (0,1)$, then $g(t)$ is monotonically decreasing as
    \begin{align*}
        g'(t) = -\frac{1}{1-t} + 1 < 0.
    \end{align*}
    Therefore, $g(t)\leq g(0)=0$ when $t\in(0,1)$. Thus substituting $t=1/x$, it holds that
    \begin{align*}
        \ln\left( 1-\frac{1}{x}\right) + \frac{1}{x} \leq 0.
    \end{align*}
    After rearrangement, we can obtain the result.
\end{proof}

%% file: main.bbl
\begin{thebibliography}{83}
\providecommand{\natexlab}[1]{#1}
\providecommand{\url}[1]{\texttt{#1}}
\expandafter\ifx\csname urlstyle\endcsname\relax
  \providecommand{\doi}[1]{doi: #1}\else
  \providecommand{\doi}{doi: \begingroup \urlstyle{rm}\Url}\fi

\bibitem[Arjevani and Field(2020)]{arjevani2020analytic}
Yossi Arjevani and Michael Field.
\newblock Analytic characterization of the hessian in shallow relu models: A tale of symmetry, 2020.

\bibitem[Arora et~al.(2019)Arora, Du, Hu, Li, Salakhutdinov, and Wang]{arora2019exact}
Sanjeev Arora, Simon~S Du, Wei Hu, Zhiyuan Li, Russ~R Salakhutdinov, and Ruosong Wang.
\newblock On exact computation with an infinitely wide neural net.
\newblock \emph{Advances in neural information processing systems}, 32, 2019.

\bibitem[Aybat et~al.(2019)Aybat, Fallah, Gurbuzbalaban, and Ozdaglar]{aybat2019universally}
Necdet~Serhat Aybat, Alireza Fallah, Mert Gurbuzbalaban, and Asuman Ozdaglar.
\newblock A universally optimal multistage accelerated stochastic gradient method.
\newblock \emph{Advances in neural information processing systems}, 32, 2019.

\bibitem[Bach and Moulines(2013)]{bach2013non}
Francis Bach and Eric Moulines.
\newblock Non-strongly-convex smooth stochastic approximation with convergence rate o (1/n).
\newblock \emph{Advances in neural information processing systems}, 26, 2013.

\bibitem[Bollapragada et~al.(2022)Bollapragada, Chen, and Ward]{bollapragada2022fast}
Raghu Bollapragada, Tyler Chen, and Rachel Ward.
\newblock On the fast convergence of minibatch heavy ball momentum.
\newblock \emph{arXiv preprint arXiv:2206.07553}, 2022.

\bibitem[Brown et~al.(2020)Brown, Mann, Ryder, Subbiah, Kaplan, Dhariwal, Neelakantan, Shyam, Sastry, Askell, Agarwal, Herbert-Voss, Krueger, Henighan, Child, Ramesh, Ziegler, Wu, Winter, Hesse, Chen, Sigler, Litwin, Gray, Chess, Clark, Berner, McCandlish, Radford, Sutskever, and Amodei]{brown2020gpt3}
Tom Brown, Benjamin Mann, Nick Ryder, Melanie Subbiah, Jared~D Kaplan, Prafulla Dhariwal, Arvind Neelakantan, Pranav Shyam, Girish Sastry, Amanda Askell, Sandhini Agarwal, Ariel Herbert-Voss, Gretchen Krueger, Tom Henighan, Rewon Child, Aditya Ramesh, Daniel Ziegler, Jeffrey Wu, Clemens Winter, Chris Hesse, Mark Chen, Eric Sigler, Mateusz Litwin, Scott Gray, Benjamin Chess, Jack Clark, Christopher Berner, Sam McCandlish, Alec Radford, Ilya Sutskever, and Dario Amodei.
\newblock Language models are few-shot learners.
\newblock In H.~Larochelle, M.~Ranzato, R.~Hadsell, M.F. Balcan, and H.~Lin, editors, \emph{Advances in Neural Information Processing Systems}, volume~33, pages 1877--1901. Curran Associates, Inc., 2020.
\newblock URL \url{https://proceedings.neurips.cc/paper/2020/file/1457c0d6bfcb4967418bfb8ac142f64a-Paper.pdf}.

\bibitem[Can et~al.(2019)Can, Gurbuzbalaban, and Zhu]{can2019accelerated}
Bugra Can, Mert Gurbuzbalaban, and Lingjiong Zhu.
\newblock Accelerated linear convergence of stochastic momentum methods in wasserstein distances.
\newblock In \emph{International Conference on Machine Learning}, pages 891--901. PMLR, 2019.

\bibitem[Chang and Lin(2011)]{chang2011libsvm}
Chih-Chung Chang and Chih-Jen Lin.
\newblock Libsvm: a library for support vector machines.
\newblock \emph{ACM transactions on intelligent systems and technology (TIST)}, 2\penalty0 (3):\penalty0 1--27, 2011.

\bibitem[Chaudhari and Soatto(2018)]{chaudhari2018stochastic}
Pratik Chaudhari and Stefano Soatto.
\newblock Stochastic gradient descent performs variational inference, converges to limit cycles for deep networks.
\newblock In \emph{2018 Information Theory and Applications Workshop (ITA)}, pages 1--10. IEEE, 2018.

\bibitem[Davis et~al.(2023)Davis, Drusvyatskiy, and Charisopoulos]{davis2023stochastic}
Damek Davis, Dmitriy Drusvyatskiy, and Vasileios Charisopoulos.
\newblock Stochastic algorithms with geometric step decay converge linearly on sharp functions.
\newblock \emph{Mathematical Programming}, pages 1--46, 2023.

\bibitem[Devlin et~al.(2019)Devlin, Chang, Lee, and Toutanova]{devlin2018bert}
Jacob Devlin, Ming-Wei Chang, Kenton Lee, and Kristina Toutanova.
\newblock Bert: Pre-training of deep bidirectional transformers for language understanding.
\newblock In \emph{Proceedings of the 2019 Conference of the North American Chapter of the Association for Computational Linguistics: Human Language Technologies, Volume 1 (Long and Short Papers)}, pages 4171--4186, 2019.

\bibitem[Devolder et~al.(2013)Devolder, Glineur, Nesterov, et~al.]{devolder2013first}
Olivier Devolder, Fran{\c{c}}ois Glineur, Yurii Nesterov, et~al.
\newblock First-order methods with inexact oracle: the strongly convex case.
\newblock \emph{CORE Discussion Papers}, 2013016:\penalty0 47, 2013.

\bibitem[Dieuleveut et~al.(2017)Dieuleveut, Flammarion, and Bach]{dieuleveut2017harder}
Aymeric Dieuleveut, Nicolas Flammarion, and Francis Bach.
\newblock Harder, better, faster, stronger convergence rates for least-squares regression.
\newblock \emph{The Journal of Machine Learning Research}, 18\penalty0 (1):\penalty0 3520--3570, 2017.

\bibitem[Dua and Graff(2017)]{Dua2017UCI}
Dheeru Dua and Casey Graff.
\newblock {UCI} machine learning repository, 2017.
\newblock URL \url{http://archive.ics.uci.edu/ml}.

\bibitem[Ge et~al.(2019)Ge, Kakade, Kidambi, and Netrapalli]{ge2019step}
Rong Ge, Sham~M Kakade, Rahul Kidambi, and Praneeth Netrapalli.
\newblock The step decay schedule: A near optimal, geometrically decaying learning rate procedure for least squares.
\newblock In \emph{Advances in Neural Information Processing Systems}, pages 14977--14988, 2019.

\bibitem[Ghadimi and Lan(2013)]{ghadimi2013optimal}
Saeed Ghadimi and Guanghui Lan.
\newblock Optimal stochastic approximation algorithms for strongly convex stochastic composite optimization, ii: shrinking procedures and optimal algorithms.
\newblock \emph{SIAM Journal on Optimization}, 23\penalty0 (4):\penalty0 2061--2089, 2013.

\bibitem[Gitman et~al.(2019)Gitman, Lang, Zhang, and Xiao]{gitman2019understanding}
Igor Gitman, Hunter Lang, Pengchuan Zhang, and Lin Xiao.
\newblock Understanding the role of momentum in stochastic gradient methods.
\newblock \emph{Advances in Neural Information Processing Systems}, 32, 2019.

\bibitem[Harvey et~al.(2019)Harvey, Liaw, Plan, and Randhawa]{harvey2019tight}
Nicholas~JA Harvey, Christopher Liaw, Yaniv Plan, and Sikander Randhawa.
\newblock Tight analyses for non-smooth stochastic gradient descent.
\newblock In \emph{Conference on Learning Theory}, pages 1579--1613. PMLR, 2019.

\bibitem[Hazan and Kale(2014)]{hazan2014beyond}
Elad Hazan and Satyen Kale.
\newblock Beyond the regret minimization barrier: optimal algorithms for stochastic strongly-convex optimization.
\newblock \emph{The Journal of Machine Learning Research}, 15\penalty0 (1):\penalty0 2489--2512, 2014.

\bibitem[He et~al.(2015)He, Zhang, Ren, and Sun]{he2015resnet}
Kaiming He, Xiangyu Zhang, Shaoqing Ren, and Jian Sun.
\newblock Deep residual learning for image recognition, 2015.

\bibitem[Howard and Ruder(2018)]{howard2018lineardecay}
Jeremy Howard and Sebastian Ruder.
\newblock Universal language model fine-tuning for text classification.
\newblock In \emph{Proceedings of the 56th Annual Meeting of the Association for Computational Linguistics (Volume 1: Long Papers)}, pages 328--339, 2018.

\bibitem[Huang et~al.(2017)Huang, Liu, Van Der~Maaten, and Weinberger]{huang2017densenet}
Gao Huang, Zhuang Liu, Laurens Van Der~Maaten, and Kilian~Q Weinberger.
\newblock Densely connected convolutional networks.
\newblock In \emph{Proceedings of the IEEE conference on computer vision and pattern recognition}, pages 4700--4708, 2017.

\bibitem[Izsak et~al.(2021)Izsak, Berchansky, and Levy]{24hbert2021}
Peter Izsak, Moshe Berchansky, and Omer Levy.
\newblock How to train bert with an academic budget, 2021.
\newblock URL \url{https://arxiv.org/abs/2104.07705}.

\bibitem[Jacot et~al.(2018)Jacot, Gabriel, and Hongler]{jacot2018neural}
Arthur Jacot, Franck Gabriel, and Cl{\'e}ment Hongler.
\newblock Neural tangent kernel: Convergence and generalization in neural networks.
\newblock \emph{Advances in neural information processing systems}, 31, 2018.

\bibitem[Jain et~al.(2018{\natexlab{a}})Jain, Kakade, Kidambi, Netrapalli, and Sidford]{jain2018accelerating}
Prateek Jain, Sham~M Kakade, Rahul Kidambi, Praneeth Netrapalli, and Aaron Sidford.
\newblock Accelerating stochastic gradient descent for least squares regression.
\newblock In \emph{Conference On Learning Theory}, pages 545--604. PMLR, 2018{\natexlab{a}}.

\bibitem[Jain et~al.(2018{\natexlab{b}})Jain, Kakade, Kidambi, Netrapalli, and Sidford]{jain2018parallelizing}
Prateek Jain, Sham~M Kakade, Rahul Kidambi, Praneeth Netrapalli, and Aaron Sidford.
\newblock Parallelizing stochastic gradient descent for least squares regression: mini-batching, averaging, and model misspecification.
\newblock \emph{Journal of machine learning research}, 18\penalty0 (223):\penalty0 1--42, 2018{\natexlab{b}}.

\bibitem[Jain et~al.(2019)Jain, Nagaraj, and Netrapalli]{jain2019making}
Prateek Jain, Dheeraj Nagaraj, and Praneeth Netrapalli.
\newblock Making the last iterate of sgd information theoretically optimal.
\newblock In \emph{Conference on Learning Theory}, pages 1752--1755. PMLR, 2019.

\bibitem[Jastrz{\k{e}}bski et~al.(2017)Jastrz{\k{e}}bski, Kenton, Arpit, Ballas, Fischer, Bengio, and Storkey]{jastrzkebski2017three}
Stanis{\l}aw Jastrz{\k{e}}bski, Zachary Kenton, Devansh Arpit, Nicolas Ballas, Asja Fischer, Yoshua Bengio, and Amos Storkey.
\newblock Three factors influencing minima in sgd.
\newblock \emph{arXiv preprint arXiv:1711.04623}, 2017.

\bibitem[Jin et~al.(2022)Jin, Xing, and He]{jin2022msgd-smooth}
Ruinan Jin, Yu~Xing, and Xingkang He.
\newblock On the convergence of msgd and adagrad for stochastic optimization, 2022.
\newblock URL \url{https://arxiv.org/abs/2201.11204}.

\bibitem[Kairouz et~al.(2021)Kairouz, McMahan, Avent, Bellet, Bennis, Bhagoji, Bonawitz, Charles, Cormode, Cummings, D’Oliveira, Eichner, Rouayheb, Evans, Gardner, Garrett, Gascón, Ghazi, Gibbons, Gruteser, Harchaoui, He, He, Huo, Hutchinson, Hsu, Jaggi, Javidi, Joshi, Khodak, Konecný, Korolova, Koushanfar, Koyejo, Lepoint, Liu, Mittal, Mohri, Nock, Özgür, Pagh, Qi, Ramage, Raskar, Raykova, Song, Song, Stich, Sun, Suresh, Tramèr, Vepakomma, Wang, Xiong, Xu, Yang, Yu, Yu, and Zhao]{kairouz2021federated-learning}
Peter Kairouz, H.~Brendan McMahan, Brendan Avent, Aurélien Bellet, Mehdi Bennis, Arjun~Nitin Bhagoji, Kallista Bonawitz, Zachary Charles, Graham Cormode, Rachel Cummings, Rafael G.~L. D’Oliveira, Hubert Eichner, Salim~El Rouayheb, David Evans, Josh Gardner, Zachary Garrett, Adrià Gascón, Badih Ghazi, Phillip~B. Gibbons, Marco Gruteser, Zaid Harchaoui, Chaoyang He, Lie He, Zhouyuan Huo, Ben Hutchinson, Justin Hsu, Martin Jaggi, Tara Javidi, Gauri Joshi, Mikhail Khodak, Jakub Konecný, Aleksandra Korolova, Farinaz Koushanfar, Sanmi Koyejo, Tancrède Lepoint, Yang Liu, Prateek Mittal, Mehryar Mohri, Richard Nock, Ayfer Özgür, Rasmus Pagh, Hang Qi, Daniel Ramage, Ramesh Raskar, Mariana Raykova, Dawn Song, Weikang Song, Sebastian~U. Stich, Ziteng Sun, Ananda~Theertha Suresh, Florian Tramèr, Praneeth Vepakomma, Jianyu Wang, Li~Xiong, Zheng Xu, Qiang Yang, Felix~X. Yu, Han Yu, and Sen Zhao.
\newblock Advances and open problems in federated learning.
\newblock \emph{Foundations and Trends® in Machine Learning}, 14\penalty0 (1–2):\penalty0 1--210, 2021.
\newblock ISSN 1935-8237.
\newblock \doi{10.1561/2200000083}.
\newblock URL \url{http://dx.doi.org/10.1561/2200000083}.

\bibitem[Kidambi et~al.(2018)Kidambi, Netrapalli, Jain, and Kakade]{kidambi2018moment-insufficiency}
Rahul Kidambi, Praneeth Netrapalli, Prateek Jain, and Sham~M. Kakade.
\newblock On the insufficiency of existing momentum schemes for stochastic optimization, 2018.
\newblock URL \url{https://arxiv.org/abs/1803.05591}.

\bibitem[Krizhevsky et~al.(2009)Krizhevsky, Hinton, et~al.]{krizhevsky2009cifar10}
Alex Krizhevsky, Geoffrey Hinton, et~al.
\newblock Learning multiple layers of features from tiny images.
\newblock 2009.

\bibitem[Kulunchakov and Mairal(2019)]{kulunchakov2019generic}
Andrei Kulunchakov and Julien Mairal.
\newblock A generic acceleration framework for stochastic composite optimization.
\newblock \emph{Advances in Neural Information Processing Systems}, 32, 2019.

\bibitem[Kushner and Clark(2012)]{kushner2012stochastic}
Harold~Joseph Kushner and Dean~S Clark.
\newblock \emph{Stochastic approximation methods for constrained and unconstrained systems}, volume~26.
\newblock Springer Science \& Business Media, 2012.

\bibitem[Li et~al.(2022)Li, Liu, and Orabona]{li2022moment-last-iterate}
Xiaoyu Li, Mingrui Liu, and Francesco Orabona.
\newblock On the last iterate convergence of momentum methods.
\newblock In Sanjoy Dasgupta and Nika Haghtalab, editors, \emph{Proceedings of The 33rd International Conference on Algorithmic Learning Theory}, volume 167 of \emph{Proceedings of Machine Learning Research}, pages 699--717. PMLR, 29 Mar--01 Apr 2022.
\newblock URL \url{https://proceedings.mlr.press/v167/li22a.html}.

\bibitem[Liu and Belkin(2019)]{liu2018accelerating}
Chaoyue Liu and Mikhail Belkin.
\newblock Accelerating sgd with momentum for over-parameterized learning.
\newblock In \emph{International Conference on Learning Representations}, 2019.

\bibitem[Liu et~al.(2020)Liu, Gao, and Yin]{liu2020improved}
Yanli Liu, Yuan Gao, and Wotao Yin.
\newblock An improved analysis of stochastic gradient descent with momentum.
\newblock \emph{Advances in Neural Information Processing Systems}, 33:\penalty0 18261--18271, 2020.

\bibitem[Loizou and Richt{\'a}rik(2017)]{loizou2017linearly}
Nicolas Loizou and Peter Richt{\'a}rik.
\newblock Linearly convergent stochastic heavy ball method for minimizing generalization error.
\newblock \emph{arXiv preprint arXiv:1710.10737}, 2017.

\bibitem[Loizou and Richt{\'a}rik(2020)]{loizou2020momentum}
Nicolas Loizou and Peter Richt{\'a}rik.
\newblock Momentum and stochastic momentum for stochastic gradient, newton, proximal point and subspace descent methods.
\newblock \emph{Computational Optimization and Applications}, 77\penalty0 (3):\penalty0 653--710, 2020.

\bibitem[Loizou et~al.(2021)Loizou, Vaswani, Hadj~Laradji, and Lacoste-Julien]{loizou2021polyak-stepsize}
Nicolas Loizou, Sharan Vaswani, Issam Hadj~Laradji, and Simon Lacoste-Julien.
\newblock Stochastic polyak step-size for sgd: An adaptive learning rate for fast convergence.
\newblock In Arindam Banerjee and Kenji Fukumizu, editors, \emph{Proceedings of The 24th International Conference on Artificial Intelligence and Statistics}, volume 130 of \emph{Proceedings of Machine Learning Research}, pages 1306--1314. PMLR, 13--15 Apr 2021.
\newblock URL \url{https://proceedings.mlr.press/v130/loizou21a.html}.

\bibitem[Loshchilov and Hutter(2017)]{LoshchilovH17cosinedecay}
Ilya Loshchilov and Frank Hutter.
\newblock {SGDR:} stochastic gradient descent with warm restarts.
\newblock In \emph{5th International Conference on Learning Representations, {ICLR} 2017, Toulon, France, April 24-26, 2017, Conference Track Proceedings}, 2017.

\bibitem[Mai and Johansson(2020)]{mai2020convergence}
Vien Mai and Mikael Johansson.
\newblock Convergence of a stochastic gradient method with momentum for non-smooth non-convex optimization.
\newblock In \emph{International Conference on Machine Learning}, pages 6630--6639. PMLR, 2020.

\bibitem[Nemirovski(1995)]{nemirovski1995information}
Arkadi Nemirovski.
\newblock Information-based complexity of convex programming.
\newblock \emph{Lecture notes}, 834, 1995.

\bibitem[Nesterov(1983)]{nesterov1983method}
Yurii Nesterov.
\newblock A method for unconstrained convex minimization problem with the rate of convergence o (1/k\^{} 2).
\newblock In \emph{Doklady an ussr}, volume 269, pages 543--547, 1983.

\bibitem[Nesterov(2003)]{nesterov2003introductory}
Yurii Nesterov.
\newblock \emph{Introductory lectures on convex optimization: A basic course}, volume~87.
\newblock Springer Science \& Business Media, 2003.

\bibitem[Ouyang et~al.(2022)Ouyang, Wu, Jiang, Almeida, Wainwright, Mishkin, Zhang, Agarwal, Slama, Ray, et~al.]{ouyang2022instructgpt}
Long Ouyang, Jeffrey Wu, Xu~Jiang, Diogo Almeida, Carroll Wainwright, Pamela Mishkin, Chong Zhang, Sandhini Agarwal, Katarina Slama, Alex Ray, et~al.
\newblock Training language models to follow instructions with human feedback.
\newblock \emph{Advances in Neural Information Processing Systems}, 35:\penalty0 27730--27744, 2022.

\bibitem[Pan et~al.(2021)Pan, Ye, and Zhang]{pan2021eigencurve}
Rui Pan, Haishan Ye, and Tong Zhang.
\newblock Eigencurve: Optimal learning rate schedule for sgd on quadratic objectives with skewed hessian spectrums.
\newblock In \emph{International Conference on Learning Representations}, 2021.

\bibitem[Pan et~al.(2022)Pan, Diao, Chen, and Zhang]{pan2022extremebert}
Rui Pan, Shizhe Diao, Jianlin Chen, and Tong Zhang.
\newblock Extremebert: A toolkit for accelerating pretraining of customized bert.
\newblock \emph{arXiv preprint arXiv:2211.17201}, 2022.

\bibitem[Polyak(1964)]{polyak19641heavyball}
B.T. Polyak.
\newblock Some methods of speeding up the convergence of iteration methods.
\newblock \emph{USSR Computational Mathematics and Mathematical Physics}, 4\penalty0 (5):\penalty0 1--17, 1964.
\newblock ISSN 0041-5553.
\newblock \doi{https://doi.org/10.1016/0041-5553(64)90137-5}.
\newblock URL \url{https://www.sciencedirect.com/science/article/pii/0041555364901375}.

\bibitem[Polyak(1987)]{polyak1987introduction}
BT~Polyak.
\newblock Introduction to optimization, optimization software, inc. publ. division, 1987.

\bibitem[Qian et~al.(2021)Qian, Richt{\'a}rik, and Zhang]{qian2021error}
Xun Qian, Peter Richt{\'a}rik, and Tong Zhang.
\newblock Error compensated distributed sgd can be accelerated.
\newblock \emph{Advances in Neural Information Processing Systems}, 34, 2021.

\bibitem[Robbins and Monro(1951)]{robbins1951sgd}
Herbert Robbins and Sutton Monro.
\newblock A stochastic approximation method.
\newblock \emph{The annals of mathematical statistics}, pages 400--407, 1951.

\bibitem[Sagun et~al.(2017)Sagun, Bottou, and LeCun]{sagun2017eigenvalues}
Levent Sagun, Leon Bottou, and Yann LeCun.
\newblock Eigenvalues of the hessian in deep learning: Singularity and beyond, 2017.

\bibitem[Sagun et~al.(2018)Sagun, Evci, Guney, Dauphin, and Bottou]{sagun2017empirical}
Levent Sagun, Utku Evci, V~Ugur Guney, Yann Dauphin, and Leon Bottou.
\newblock Empirical analysis of the hessian of over-parametrized neural networks.
\newblock \emph{CoRR, abs/1706.04454}, 2018.

\bibitem[Sandler et~al.(2018)Sandler, Howard, Zhu, Zhmoginov, and Chen]{sandler2018mobilenetv2}
Mark Sandler, Andrew Howard, Menglong Zhu, Andrey Zhmoginov, and Liang-Chieh Chen.
\newblock Mobilenetv2: Inverted residuals and linear bottlenecks.
\newblock In \emph{Proceedings of the IEEE conference on computer vision and pattern recognition}, pages 4510--4520, 2018.

\bibitem[Sebbouh et~al.(2021)Sebbouh, Gower, and Defazio]{sebbouh2021almost}
Othmane Sebbouh, Robert~M Gower, and Aaron Defazio.
\newblock Almost sure convergence rates for stochastic gradient descent and stochastic heavy ball.
\newblock In \emph{Conference on Learning Theory}, pages 3935--3971. PMLR, 2021.

\bibitem[Shamir and Zhang(2013)]{shamir2013stochastic}
Ohad Shamir and Tong Zhang.
\newblock Stochastic gradient descent for non-smooth optimization: Convergence results and optimal averaging schemes.
\newblock In \emph{International conference on machine learning}, pages 71--79. PMLR, 2013.

\bibitem[Simonyan and Zisserman(2015)]{simonyan2015vgg16}
Karen Simonyan and Andrew Zisserman.
\newblock Very deep convolutional networks for large-scale image recognition, 2015.

\bibitem[Szegedy et~al.(2015)Szegedy, Liu, Jia, Sermanet, Reed, Anguelov, Erhan, Vanhoucke, and Rabinovich]{szegedy2014googlenet}
Christian Szegedy, Wei Liu, Yangqing Jia, Pierre Sermanet, Scott Reed, Dragomir Anguelov, Dumitru Erhan, Vincent Vanhoucke, and Andrew Rabinovich.
\newblock Going deeper with convolutions.
\newblock In \emph{Proceedings of the IEEE conference on computer vision and pattern recognition}, pages 1--9, 2015.

\bibitem[Tang et~al.(2023)Tang, Liu, and Zhang]{tang2023acceleration}
Kejie Tang, Weidong Liu, and Yichen Zhang.
\newblock Acceleration of stochastic gradient descent with momentum by averaging: finite-sample rates and asymptotic normality.
\newblock \emph{arXiv preprint arXiv:2305.17665}, 2023.

\bibitem[Touvron et~al.(2023{\natexlab{a}})Touvron, Lavril, Izacard, Martinet, Lachaux, Lacroix, Rozi{\`e}re, Goyal, Hambro, Azhar, et~al.]{touvron2023llama}
Hugo Touvron, Thibaut Lavril, Gautier Izacard, Xavier Martinet, Marie-Anne Lachaux, Timoth{\'e}e Lacroix, Baptiste Rozi{\`e}re, Naman Goyal, Eric Hambro, Faisal Azhar, et~al.
\newblock Llama: Open and efficient foundation language models.
\newblock \emph{arXiv preprint arXiv:2302.13971}, 2023{\natexlab{a}}.

\bibitem[Touvron et~al.(2023{\natexlab{b}})Touvron, Martin, Stone, Albert, Almahairi, Babaei, Bashlykov, Batra, Bhargava, Bhosale, et~al.]{touvron2023llama2}
Hugo Touvron, Louis Martin, Kevin Stone, Peter Albert, Amjad Almahairi, Yasmine Babaei, Nikolay Bashlykov, Soumya Batra, Prajjwal Bhargava, Shruti Bhosale, et~al.
\newblock Llama 2: Open foundation and fine-tuned chat models.
\newblock \emph{arXiv preprint arXiv:2307.09288}, 2023{\natexlab{b}}.

\bibitem[Vaswani et~al.(2019)Vaswani, Mishkin, Laradji, Schmidt, Gidel, and Lacoste-Julien]{vaswani2019painless}
Sharan Vaswani, Aaron Mishkin, Issam Laradji, Mark Schmidt, Gauthier Gidel, and Simon Lacoste-Julien.
\newblock Painless stochastic gradient: Interpolation, line-search, and convergence rates.
\newblock \emph{Advances in neural information processing systems}, 32, 2019.

\bibitem[Verbraeken et~al.(2020)Verbraeken, Wolting, Katzy, Kloppenburg, Verbelen, and Rellermeyer]{verbraeken2019distributed-ml}
Joost Verbraeken, Matthijs Wolting, Jonathan Katzy, Jeroen Kloppenburg, Tim Verbelen, and Jan~S Rellermeyer.
\newblock A survey on distributed machine learning.
\newblock \emph{Acm computing surveys (csur)}, 53\penalty0 (2):\penalty0 1--33, 2020.

\bibitem[Wang et~al.(2021)Wang, Lin, and Abernethy]{wang2021modular}
Jun-Kun Wang, Chi-Heng Lin, and Jacob~D Abernethy.
\newblock A modular analysis of provable acceleration via polyak’s momentum: Training a wide relu network and a deep linear network.
\newblock In \emph{International Conference on Machine Learning}, pages 10816--10827. PMLR, 2021.

\bibitem[Wang et~al.(2023)Wang, Malladi, Wang, Lyu, and Li]{wang2023marginal}
Runzhe Wang, Sadhika Malladi, Tianhao Wang, Kaifeng Lyu, and Zhiyuan Li.
\newblock The marginal value of momentum for small learning rate sgd.
\newblock \emph{arXiv preprint arXiv:2307.15196}, 2023.

\bibitem[Wang and Johansson(2021)]{wang2021nonconvex-sgdm-step}
Xiaoyu Wang and Mikael Johansson.
\newblock Bandwidth-based step-sizes for non-convex stochastic optimization, 2021.
\newblock URL \url{https://arxiv.org/abs/2106.02888}.

\bibitem[Wettig et~al.(2022)Wettig, Gao, Zhong, and Chen]{wettig2022mask15}
Alexander Wettig, Tianyu Gao, Zexuan Zhong, and Danqi Chen.
\newblock Should you mask 15% in masked language modeling?, 2022.
\newblock URL \url{https://arxiv.org/abs/2202.08005}.

\bibitem[Wolf(2021)]{wolf2021stochastic}
Florian Wolf.
\newblock Stochastic gradient descent and its application for parametrized boundary value problems under uncertainties.
\newblock 2021.

\bibitem[Wu et~al.(2022{\natexlab{a}})Wu, Zou, Braverman, Gu, and Kakade]{wu2022last}
Jingfeng Wu, Difan Zou, Vladimir Braverman, Quanquan Gu, and Sham Kakade.
\newblock Last iterate risk bounds of sgd with decaying stepsize for overparameterized linear regression.
\newblock In \emph{International Conference on Machine Learning}, pages 24280--24314. PMLR, 2022{\natexlab{a}}.

\bibitem[Wu et~al.(2022{\natexlab{b}})Wu, Wang, and Su]{wu2022alignment}
Lei Wu, Mingze Wang, and Weijie Su.
\newblock The alignment property of sgd noise and how it helps select flat minima: A stability analysis.
\newblock \emph{Advances in Neural Information Processing Systems}, 35:\penalty0 4680--4693, 2022{\natexlab{b}}.

\bibitem[Xu et~al.(2016)Xu, Lin, and Yang]{xu2016accelerate}
Yi~Xu, Qihang Lin, and Tianbao Yang.
\newblock Accelerate stochastic subgradient method by leveraging local error bound.
\newblock \emph{CoRR, abs/1607.01027}, 2016.

\bibitem[Yao et~al.(2020)Yao, Gholami, Keutzer, and Mahoney]{yao2020pyhessian}
Zhewei Yao, Amir Gholami, Kurt Keutzer, and Michael~W Mahoney.
\newblock Pyhessian: Neural networks through the lens of the hessian.
\newblock In \emph{2020 IEEE international conference on big data (Big data)}, pages 581--590. IEEE, 2020.

\bibitem[You et~al.(2017)You, Gitman, and Ginsburg]{you2017large}
Yang You, Igor Gitman, and Boris Ginsburg.
\newblock Large batch training of convolutional networks.
\newblock \emph{arXiv preprint arXiv:1708.03888}, 2017.

\bibitem[You et~al.(2018)You, Zhang, Hsieh, Demmel, and Keutzer]{you2018imagenet}
Yang You, Zhao Zhang, Cho-Jui Hsieh, James Demmel, and Kurt Keutzer.
\newblock Imagenet training in minutes.
\newblock In \emph{Proceedings of the 47th International Conference on Parallel Processing}, pages 1--10, 2018.

\bibitem[You et~al.(2019)You, Li, Reddi, Hseu, Kumar, Bhojanapalli, Song, Demmel, Keutzer, and Hsieh]{you2019large}
Yang You, Jing Li, Sashank Reddi, Jonathan Hseu, Sanjiv Kumar, Srinadh Bhojanapalli, Xiaodan Song, James Demmel, Kurt Keutzer, and Cho-Jui Hsieh.
\newblock Large batch optimization for deep learning: Training bert in 76 minutes.
\newblock In \emph{International Conference on Learning Representations}, 2019.

\bibitem[Yuan et~al.(2016)Yuan, Ying, and Sayed]{yuan2016influence}
Kun Yuan, Bicheng Ying, and Ali~H Sayed.
\newblock On the influence of momentum acceleration on online learning.
\newblock \emph{The Journal of Machine Learning Research}, 17\penalty0 (1):\penalty0 6602--6667, 2016.

\bibitem[Yuan et~al.(2021)Yuan, Chen, Huang, Zhang, Pan, Xu, and Yin]{yuan2021decentlam}
Kun Yuan, Yiming Chen, Xinmeng Huang, Yingya Zhang, Pan Pan, Yinghui Xu, and Wotao Yin.
\newblock Decentlam: Decentralized momentum sgd for large-batch deep training.
\newblock In \emph{Proceedings of the IEEE/CVF International Conference on Computer Vision}, pages 3029--3039, 2021.

\bibitem[Yuan et~al.(2019)Yuan, Yan, Jin, and Yang]{yuan2019stagewise}
Zhuoning Yuan, Yan Yan, Rong Jin, and Tianbao Yang.
\newblock Stagewise training accelerates convergence of testing error over sgd.
\newblock \emph{Advances in Neural Information Processing Systems}, 32, 2019.

\bibitem[Zeng et~al.(2023)Zeng, Han, Su, and Xie]{zeng2023adaptive}
Yun Zeng, Deren Han, Yansheng Su, and Jiaxin Xie.
\newblock On adaptive stochastic heavy ball momentum for solving linear systems.
\newblock \emph{arXiv preprint arXiv:2305.05482}, 2023.

\bibitem[Zhang et~al.(2018)Zhang, Saxe, Advani, and Lee]{zhang2018energy}
Yao Zhang, Andrew~M Saxe, Madhu~S Advani, and Alpha~A Lee.
\newblock Energy--entropy competition and the effectiveness of stochastic gradient descent in machine learning.
\newblock \emph{Molecular Physics}, 116\penalty0 (21-22):\penalty0 3214--3223, 2018.

\bibitem[Zheng et~al.(2019)Zheng, Huang, and Kwok]{zheng2019communication}
Shuai Zheng, Ziyue Huang, and James Kwok.
\newblock Communication-efficient distributed blockwise momentum sgd with error-feedback.
\newblock \emph{Advances in Neural Information Processing Systems}, 32, 2019.

\bibitem[Zhu et~al.(2019)Zhu, Wu, Yu, Wu, and Ma]{zhu2018anisotropic}
Zhanxing Zhu, Jingfeng Wu, Bing Yu, Lei Wu, and Jinwen Ma.
\newblock The anisotropic noise in stochastic gradient descent: Its behavior of escaping from sharp minima and regularization effects.
\newblock In \emph{International Conference on Machine Learning}, pages 7654--7663. PMLR, 2019.

\end{thebibliography}
